\documentclass[acmtog]{acmart}
\acmSubmissionID{331}

\usepackage{hyperref}
\usepackage{url}
\usepackage{verbatim}
\usepackage{color,soul}
\usepackage{graphicx}
\def\code#1{\texttt{#1}}
\usepackage{booktabs}
\usepackage{caption}
\usepackage{paralist}

  \pltopsep=\medskipamount
\usepackage[T1]{fontenc}
\usepackage{dblfloatfix}

\newcommand{\FGym}{\textit{Fusion 360 Gym}}
\newcommand{\FGal}{\textit{Fusion 360 Gallery}}
\newcommand{\FRec}{\textit{Fusion 360 Gallery} reconstruction dataset}
\usepackage{xcolor}

\setcopyright{acmlicensed}
\acmJournal{TOG}
\acmYear{2021}
\acmVolume{40}
\acmNumber{4}
\acmArticle{54}
\acmMonth{8}
\acmDOI{10.1145/3450626.3459818}

\title{Fusion 360 Gallery: A Dataset and Environment for Programmatic CAD Construction from Human Design Sequences}

\author{Karl D.D. Willis}
\affiliation{%
 \institution{Autodesk Research}
 \city{San Francisco}
 \state{California}
 \country{USA}
 }
\email{karl.willis@autodesk.com}
\author{Yewen Pu}
\affiliation{%
 \institution{Autodesk Research}
 \city{San Francisco}
 \state{California}
 \country{USA}
 }
\author{Jieliang Luo}
\affiliation{%
 \institution{Autodesk Research}
 \city{San Francisco}
 \state{California}
 \country{USA}
 }
\author{Hang Chu}
\affiliation{%
 \institution{Autodesk Research}
 \city{Toronto}
 \state{Ontario}
 \country{Canada}
 }
\author{Tao Du}
\affiliation{%
 \institution{Massachusetts Institute of Technology}
 \city{Cambridge}
 \state{Massachusetts}
 \country{USA}
 }
\author{Joseph G. Lambourne}
\affiliation{%
 \institution{Autodesk Research}
 \city{Soho}
 \state{London}
 \country{United Kingdom}
 }
\author{Armando Solar-Lezama}
\affiliation{%
 \institution{Massachusetts Institute of Technology}
 \city{Cambridge}
 \state{Massachusetts}
 \country{USA}
 }
\author{Wojciech Matusik}
\affiliation{%
 \institution{Massachusetts Institute of Technology}
 \city{Cambridge}
 \state{Massachusetts}
 \country{USA}
 }


\begin{document}

\begin{abstract}
Parametric computer-aided design (CAD) is a standard paradigm used to design manufactured objects, where a 3D shape is represented as a program supported by the CAD software.
Despite the pervasiveness of parametric CAD and a growing interest from the research community, currently there does not exist a dataset of realistic CAD models in a concise programmatic form.
In this paper we present the \textit{Fusion 360 Gallery}, consisting of a simple language with just the \textit{sketch} and \textit{extrude} modeling operations, and a dataset of 8,625 human design sequences expressed in this language.
We also present an interactive environment called the \textit{Fusion 360 Gym}, which exposes the sequential construction of a CAD program as a Markov decision process, making it amendable to machine learning approaches.
As a use case for our dataset and environment, we define the CAD reconstruction task of recovering a CAD program from a target geometry. We report results of applying state-of-the-art methods of program synthesis with neurally guided search on this task.
\end{abstract}

\begin{teaserfigure}
    \includegraphics[width=\textwidth]{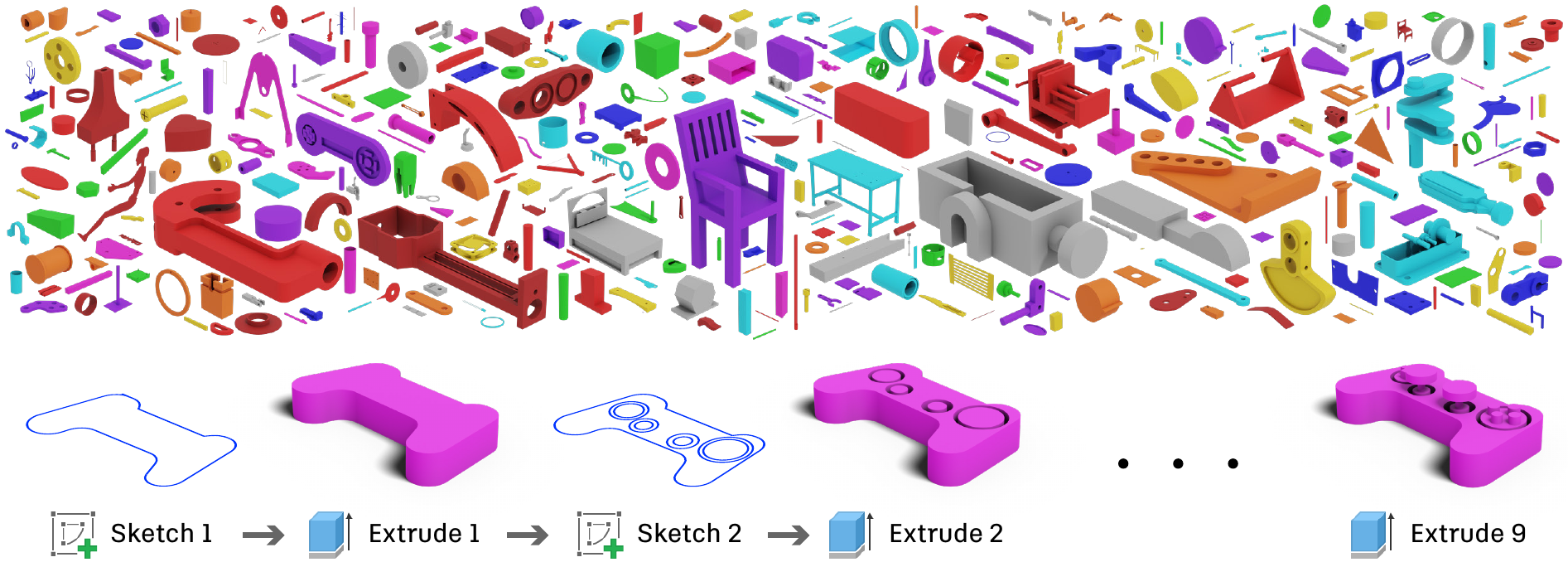}
    \caption{Top: A subset of designs containing ground-truth CAD programs represented as construction sequences from the \FRec{}. Bottom: An example construction sequence using the \textit{sketch} and \textit{extrude} modeling operations with built-in Boolean operations.}
    \label{figure:teaser}
\end{teaserfigure}

\maketitle

\section{Introduction}

The manufactured objects that surround us in everyday life are represented programmatically in computer-aided design (CAD) software as a sequence of 2D and 3D modeling operations.
Parametric CAD files contain programmatic information that is critical for documenting design intent, maintaining editablity, and compatibility with downstream simulation and manufacturing. Embedded within these designs is the knowledge of domain experts who precisely define a sequence of modeling operations to form 3D shapes. We believe having access to a real-world collection of human design sequences, and the ability to execute them, is critical for future advances in CAD that leverage learning-based approaches.

Learning-based approaches show great potential, both for solving  existing problems such as reverse engineering~\cite{Buonamici2017}, and for providing entirely new kinds of functionality which would be unimaginable using traditional techniques.
Recent advances in neural networks have spurred new interest in data driven approaches to generating CAD programs, tackling both the forward problem of 3D shape generation~\cite{mo2019structurenet, jones2020shapeAssembly, li2020sketch2CAD} and the inverse problem of recovering CAD programs from a target geometry~\cite{sharma2017csgnet, tian2018learning, ellis2019write, kania2020ucsg}. 
However, progress has been inhibited by the lack of a human designed dataset of ground-truth CAD programs, written in a simple yet expressive Domain Specific Language (DSL) and an environment to execute them.   

We take a step towards this goal by introducing the first dataset of human designed CAD geometries, paired with their ground-truth CAD programs represented as construction sequences, along with a supporting execution environment to make learning-based approaches amendable to real CAD construction tasks.
Our dataset contains 8,625 CAD programs represented entirely in a simple language allowing sketches to be created and then extruded.
With just the \emph{sketch} and \emph{extrude} modeling operations, that also incorporate Boolean operations, a highly expressive range of 3D designs can be created (Figure~\ref{figure:teaser}).
We provide an interactive environment called the \textit{Fusion 360 Gym}, which can interpret the language of sketch and extrude, providing a geometric data structure as feedback after each operation, simulating the iterative construction process of a human designer.

As a use case for our dataset and environment, we standardize the problem of programmatic CAD reconstruction from a target geometry using a learning-based approach. We provide a benchmark, consisting of a training set of 6,900 designs and a test set of 1,725 designs, and a set of evaluation criteria. 
We then develop neurally guided search approaches for the CAD reconstruction task on this benchmark. 
Our algorithm consists of first training a policy, a message passing network (MPN) with a novel encoding of state and action, using imitation learning on ground truth construction sequences. At inference time the algorithm employs search, leveraging the learned neural policy to repeatedly interact with the \FGym{} environment until a correct CAD program is discovered. This approach is able to recover a correct CAD program for 67.5\% of designs in the test set with a budget of 100 interactions between the agent and the \FGym{}, averaging < 20 sec solve time per design. This paper makes the following contributions:
\begin{itemize}
    \item We present the \FRec{}, containing 8,625 human designed CAD programs, expressed in a simple yet expressive language of \textit{sketch} and \textit{extrude}.
    \item We introduce an environment called the \FGym{}, capable of executing the language of \textit{sketch} and \textit{extrude} and providing a geometric data structure as feedback after each operation.
    \item We standardize the task of CAD reconstruction from input geometry and use a learning-based approach with neurally guided search to produce results on real world data for the first time.
\end{itemize}

\section{Related Work}

\paragraph{CAD Datasets} Existing 3D CAD datasets have largely focused on providing mesh geometry \citep{chang2015shapenet, wu20153d, Thingi10K, mo2019partnet, sangpil2020large}.
However, the de facto standard for parametric CAD is the boundary representation (B-Rep) format, containing valuable analytic representations of surfaces and curves suitable for high level control of 3D shapes. B-Reps are collections of trimmed parametric surfaces along with topological information which describes adjacency relationships between them  \citep{weiler1986}. B-Rep datasets have recently been made available with both human designed \citep{koch2019abc} and synthetic data \citep{zhang2018featurenet, jayaraman2020uvnet, fabwave}. 
Missing from these datasets is programmatic construction sequence information containing the knowledge of how each shape is defined and created.
Although the ABC dataset includes some additional construction information in a proprietary format provided by the Onshape CAD software, missing information can only be retrieved by querying the OnShape API. Combined with sparse documentation, this makes it difficult to interpret the construction information. We are unaware of any method that can be used to rebuild designs in the ABC dataset from the provided construction information, a key requirement for tasks related to CAD construction.
We believe it is critical to understand not only \textit{what} is designed, but \textit{how} that design came about. 

Parametric CAD programs contain valuable information on the construction history of a design. \citet{schulz2014FabByExample} provide a standard collection of human designs with full parametric history, albeit a limited set of 67 designs in a proprietary format. SketchGraphs \citep{SketchGraphs} narrows the broad area of parametric CAD by focusing on the underlying 2D engineering sketches, including sketch construction sequences. Freehand 2D sketch datasets also tackle the challenge of understanding design by looking at the sequence of user actions 
\citep{gryaditskaya2019opensketch, sangkloy2016sketchy, eitz2012humans}. In the absence of human designed sequential 3D data, learning-based approaches have instead leveraged synthetic CAD construction sequences \citep{li2020sketch2CAD, sharma2017csgnet, tian2018learning, ellis2019write}. The dataset presented in this paper is the first to provide human designed 3D CAD construction sequence information suitable for use with machine learning. Table~\ref{tab:dataset_comparison} provides a feature comparison of related CAD datasets.

\begin{figure*}
    \includegraphics[width=\textwidth]{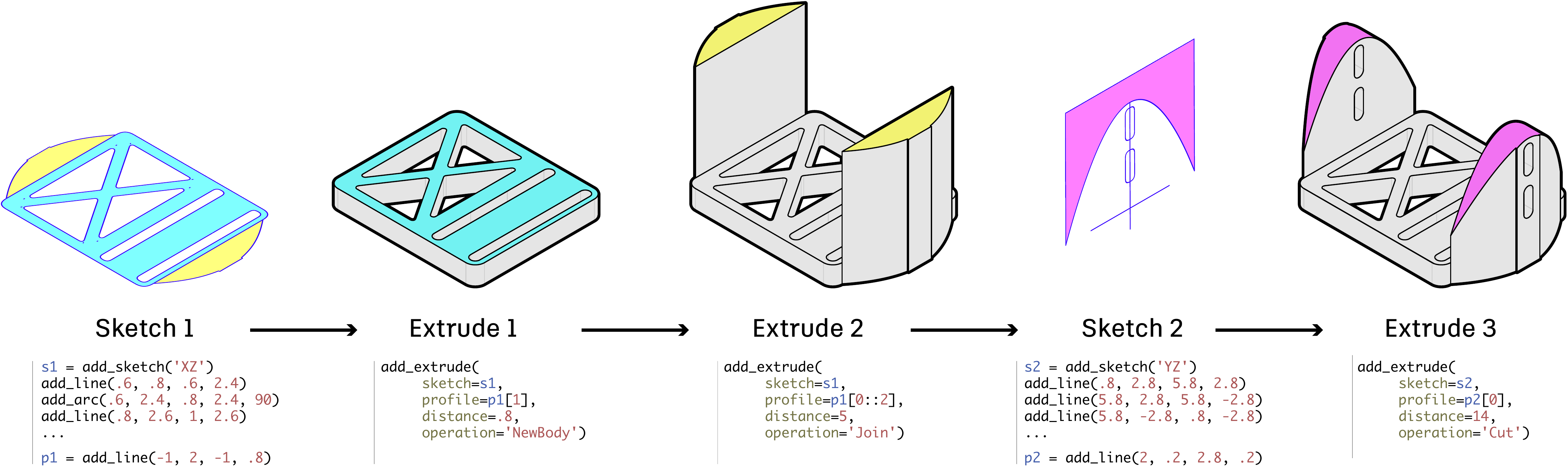}
    \caption{An example design sequence from the dataset with associated CAD program. Sketch elements form profiles that are sequentially extruded to \textit{join} (Extrude 1, Extrude 2) or \textit{cut} (Extrude 3) geometry using Boolean operations. The colored areas show the sketch profiles that partake in each extrusion.}
    \label{figure:construction_seq}
\end{figure*}

\begin{table}
    \small
    \caption{Comparison of related CAD datasets. For each dataset, we report the number of designs (\textbf{$\#$}), the design representation (\textbf{B-Rep}, \textbf{Mesh}, or \textbf{Sketch}), whether it includes a construction sequence capable of rebuilding the final design (\textbf{Seq.}), and whether it contains human annotated labels for tasks such as shape classification (\textbf{Label}). The \textit{F360 Gallery} row indicates our dataset.}
    \label{tab:dataset_comparison}
    \centering
    \begin{tabular}{l|cccccc}
        \toprule
         \textbf{Dataset} & \textbf{\#} & \textbf{B-Rep} & \textbf{Mesh} & \textbf{Sketch} &\textbf{Seq.} & \textbf{Label} \\ 
        \midrule
        ShapeNet & 3M+ & {} & \checkmark & {} & {} & \checkmark \\ 
        ABC & 1M+ & \checkmark & \checkmark & {} & {} & {} \\
        Thingi10k & 10,000 & {} & \checkmark & {} & {} & \checkmark \\
        SketchGraphs & 15M+ & {} & {} & \checkmark & \checkmark & {} \\
        \textit{F360 Gallery} & 8,625 & \checkmark & \checkmark & \checkmark & \checkmark & {} \\
        \bottomrule
    \end{tabular}
\end{table}

\paragraph{3D Shape Generation}
The forward problem of 3D shape generation has been explored extensively in recent years using learning-based approaches. Neural network based generative models are often used to enable previously challenging functionality such as shape interpolation and synthesis. Notable approaches to this problem include leveraging knowledge of object structure~\cite{mo2019structurenet, li2020learning, gao2019sdm, schor2019componet} or learning from a sequence of events to generate 3D shapes~\cite{zou20173d,sung2017complementme,wu2020pq,jones2020shapeAssembly,li2020sketch2CAD,nash2020polygen}. Unique to our work is the challenge of learning from real sequential human design data, requiring a state and action representation suitable for the language of \textit{sketch} and \textit{extrude}.

\paragraph{CAD Reconstruction}
The inverse task of CAD reconstruction involves recovering a CAD program, represented as a sequence of modeling operations, from input such as B-Reps, triangle meshes, or point clouds. Despite extensive prior work \citep{shah2001discourse}, CAD reconstruction remains a challenging problem as it requires deductions on both continuous parameters (e.g., extracting the dimensions of primitives) and discrete operations (e.g., choosing a proper operation for the next step), leading to a mixed combinatorial search space. To recover the sequence of operations, traditional methods typically run global search methods (e.g., evolutionary algorithms as in \citet{hamza2004optimization}, \citet{weiss2009geometry}, \citet{friedrich2019optimizing}, and \citet{fayolle2016evolutionary}) with heuristic rules to prune the search space~\citep{shapiro1993separation,buchele2000three,buchele2001binary,buchele2003three}. Heuristic approaches are also available in a number of commercial software tools, often as a user-guided semi-automatic system \citep{inventorFR, featureWorks} to aid with file conversion between CAD systems. These traditional algorithms operate by removing faces from the B-rep body and reapplying them as parametric modeling operations. This strategy can recover the later modeling operations, but fail to completely rebuild the construction sequence from the first step. We instead tackle the task of recovering the entire construction sequence from the first extrusion.
Another approach is using program synthesis \citep{du2018inversecsg,nandi2017programming,nandi2018functional,nandi2020synthesizing} to infer CAD programs written in DSLs from given shapes. CAD reconstruction is also related to the inverse procedural modeling problem~\citep{talton2011metropolis,stava2014trees,vanegas2012urban}, which attempts to reverse-engineer procedures that can faithfully match a given target. 

Compared to the rule-based or grammar-based methods above, learning-based approaches can potentially learn the rules that are typically hard-coded, automate scenarios that require user-input, and generalize when confronted with unfamiliar geometry. 
One early work is CSGNet \citep{sharma2017csgnet}, which trains a neural network to infer the sequence of Constructive Solid Geometry (CSG) operations based on visual input. More recent works along this line of research include \cite{ellis2019write,tian2018learning,kania2020ucsg,chen2020bsp}. Typically associated with these methods are a customized DSL, such as CSG, that parameterizes the space of geometry, some heuristic rules that limit the search space, and a neural network generative model. \citet{lin2020modeling} reconstruct input shapes with a dual action representation that first positions cuboids and then edits edge-loops for refinement. Although editing edge-loops of cuboids may be a suitable modeling operation in artistic design, it is not as expressive or precise as the sketch and extrude operations used in real mechanical components. Additionally, \citet{lin2020modeling} choose to train and evaluate their network on synthetic data due to the lack of a benchmark dataset of CAD construction sequences, a space that our work aims to fill. Our approach is the first to apply a learning-based method to reconstruction using common \textit{sketch} and \textit{extrude} CAD modeling operations from real human designs.

\section{Fusion 360 Gallery DSL and Reconstruction Dataset}

\begin{figure}
    \begin{center}
        \includegraphics[width=1\columnwidth]{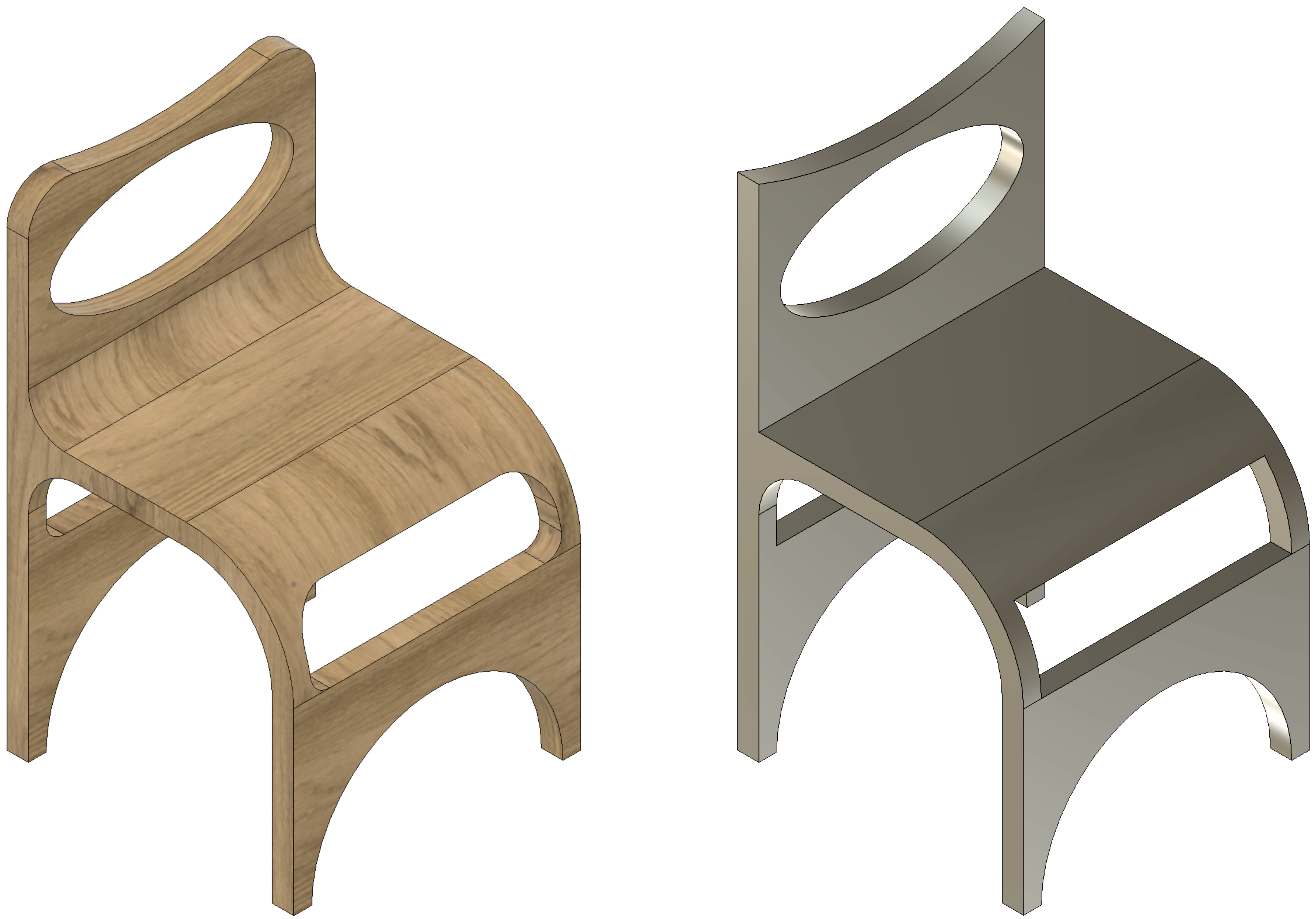}
        \caption{Modeling operations other than \textit{sketch} and \textit{extrude} are suppressed to expand the data quantity. An example design before (left) and after (right) the fillet modeling operation is suppressed. }
        \label{figure:suppression}
    \end{center}
\end{figure}

\begin{figure}
    \begin{center}
        \includegraphics[width=1\columnwidth]{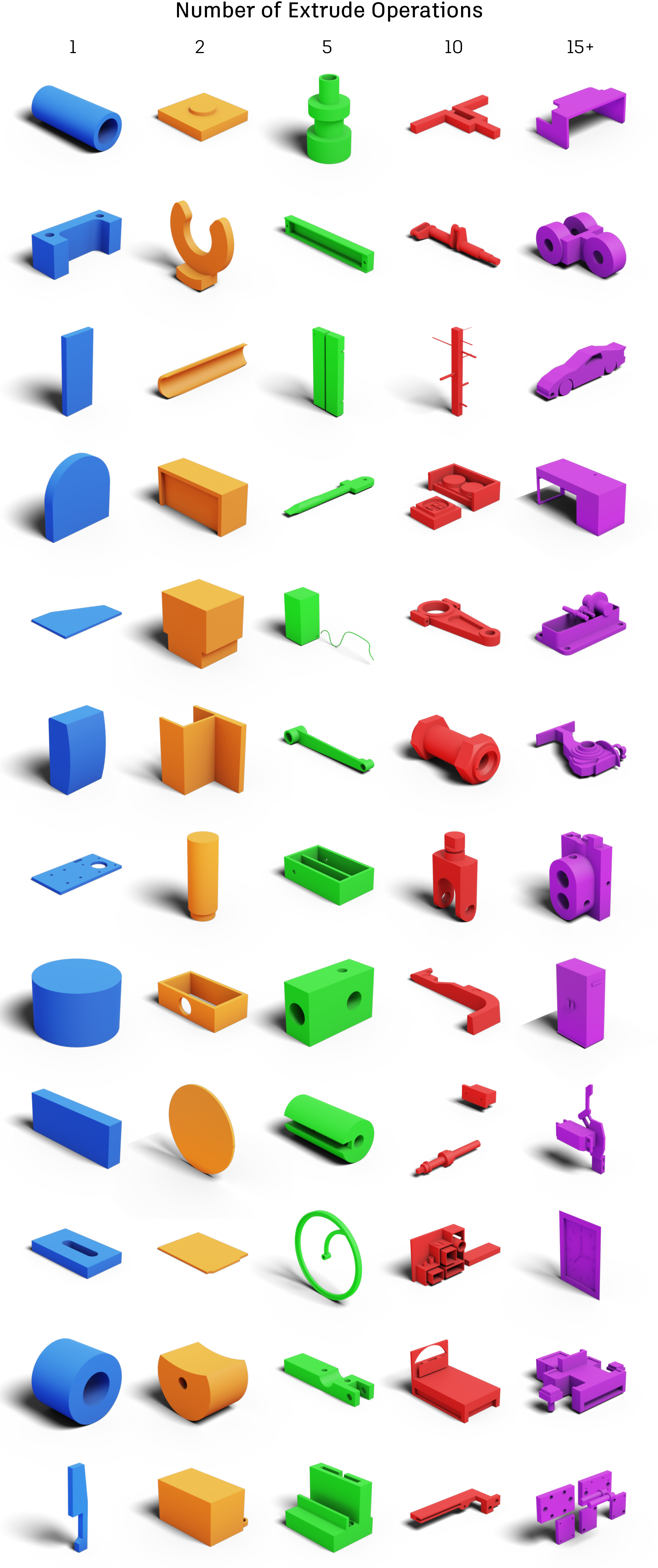}
        \caption{A random sampling of designs from the Fusion 360 Gallery reconstruction dataset, grouped by the number of extrude operations.}
        \label{figure:designs_grouped_by_extrude_count}
    \end{center}
\end{figure}

The \FRec{} consists of 8,625 designs produced by users of the CAD software Autodesk Fusion 360 and submitted to the publicly available Autodesk Online Gallery~\cite{autodeskOnlineGallery}. The data and supporting code is publicly available via GitHub\footnote{Dataset website: \url{https://github.com/AutodeskAILab/Fusion360GalleryDataset}} with a license allowing non-commercial research similar to the ImageNet~\cite{deng2009imagenet} license.
We created the dataset from approximately 20,000 designs in the native Fusion 360 CAD file format. We focus on the \textit{sketch} and \textit{extrude} modeling operations for two main reasons: 1) \textit{sketch} and \textit{extrude} are the two most common CAD modeling operations used in 84\% and 79\% of designs in the original dataset respectively; \textgreater3x more common than operations such as fillet and chamfer, and 2) we seek to balance design expressivity with a tractable problem for learning-based approaches; restricting the modeling operations to \textit{sketch} and \textit{extrude} greatly simplifies the descriptive complexity compared to the full range of CAD modeling operations.
We generate the as-designed sequence of \textit{sketch} and \textit{extrude} modeling operations by parsing the parametric history of the Fusion 360 CAD files.
Multi-component assemblies are divided into separate designs representing the constituent parts, e.g. the blade of a pocket knife. Modeling operations other than \textit{sketch} and \textit{extrude} are suppressed to expand the data quantity. Figure~\ref{figure:suppression} shows an example of suppressing a fillet operation, allowing the resulting design to be included in the dataset. 
Figure~\ref{figure:designs_grouped_by_extrude_count} shows a random sampling of the designs in the dataset grouped by the number of extrude operations used.

\begin{table}
    \caption{The grammar for the \FGal{} domain-specific language. A program consists of a sequence of \textit{sketch} and \textit{extrude} operations that iteratively modify the current geometry.}
    \label{tab:grammar}
    \[
    \begin{array}{rcl}
    P &:=& G;[X] \\
    X &:=& S ~\vert~ E \\
    S &:=& \textrm{add\_sketch}(I); [D] \\
    D &:=& L ~\vert~ A ~\vert~ C \\
    L &:=& \textrm{add\_line} (N,N,N,N) \\
    A &:=& \textrm{add\_arc} (N,N,N,N,N) \\
    C &:=& \textrm{add\_circle} (N,N,N) \\
    E &:=& \textrm{add\_extrude}([I],N,O) \\
    I &:=& \textrm{identifier}\\
    N &:=& \textrm{number}\\
    O &:=& \textrm{new body} ~\vert~ \textrm{join} ~\vert~ \textrm{cut} ~\vert~ \textrm{intersect} \\
    \end{array}
    \]
\end{table}

Each design is represented as a program expressed in a DSL, forming a simplified wrapper around the underlying Fusion 360 Python API~\citep{fusion360API}. Each design consists of a sequence of \textit{sketch} and \textit{extrude} operations that iteratively modifies the current geometry (Figure~\ref{figure:construction_seq}). We specify the core language here, and provide information on additional constructs, such as sketching of splines and double-sided extrudes, in Section~\ref{section:appendix_dataset} of the appendix. 
The \FGal{} DSL is a \emph{stateful} language consisting of a single global variable $G$, representing the current geometry under construction, and a sequence of commands $[X]$ that iteratively modifies the current geometry $G$.  Each command can be either a \textit{sketch} $S$ or an \textit{extrude} $E$ operation.
A grammar describing the core DSL is shown in Table~\ref{tab:grammar}.

\subsection{Current Geometry}
\label{section:dataset_geometry}

The current geometry $G$ is the single \emph{global state} that is updated with the sequence of commands $[X]$. It is a data structure representing all geometric information that would be available to a designer in the construction process using Fusion 360: such as inspecting different aspects of the geometry, and referencing its components for further modifications.

\paragraph{Boundary Representation}
B-Rep is the primary geometry format provided in the dataset and the native format in which designs were created, making it a natural representation for the current geometry. $G$ represents a collection of sketch or B-Rep entities, which can be referenced from the construction sequence through identifier $I$. B-Rep bodies can be expressed as a face adjacency graph, as later described in Section~\ref{subsection:state_representation}.

\paragraph{Execution} Crucially, the current geometry $G$ is iteratively updated through the sequence of commands $[X]$. After \emph{each} command $X$, the interpreter uses the underlying Fusion 360 Python API to generate an updated geometry. After all the commands $[X]$ are executed, we obtain the final geometry, $G_t$.

\paragraph{Storage}
In addition to the program $P$, \FGym{} stores the final geometry $G_t$ as a .smt file, the native B-Rep format used by Fusion 360, and neutral .step files that can be used with other CAD systems. B-Rep entities, such as bodies and faces, can be referenced from the construction sequence back to entities in the .smt file.
A mesh representation of $G_t$ is stored in .obj format representing a triangulated version of the B-Rep. Each B-Rep face is labeled as a group of triangles in the .obj file with the B-Rep face identifier as the group name. This allows the triangles to be traced back to the B-Rep face and associated extrude operation. Any intermediate geometry $G$ can also be exported in these file formats with the API.

\begin{figure}
    \begin{center}
        \includegraphics[width=1\columnwidth]{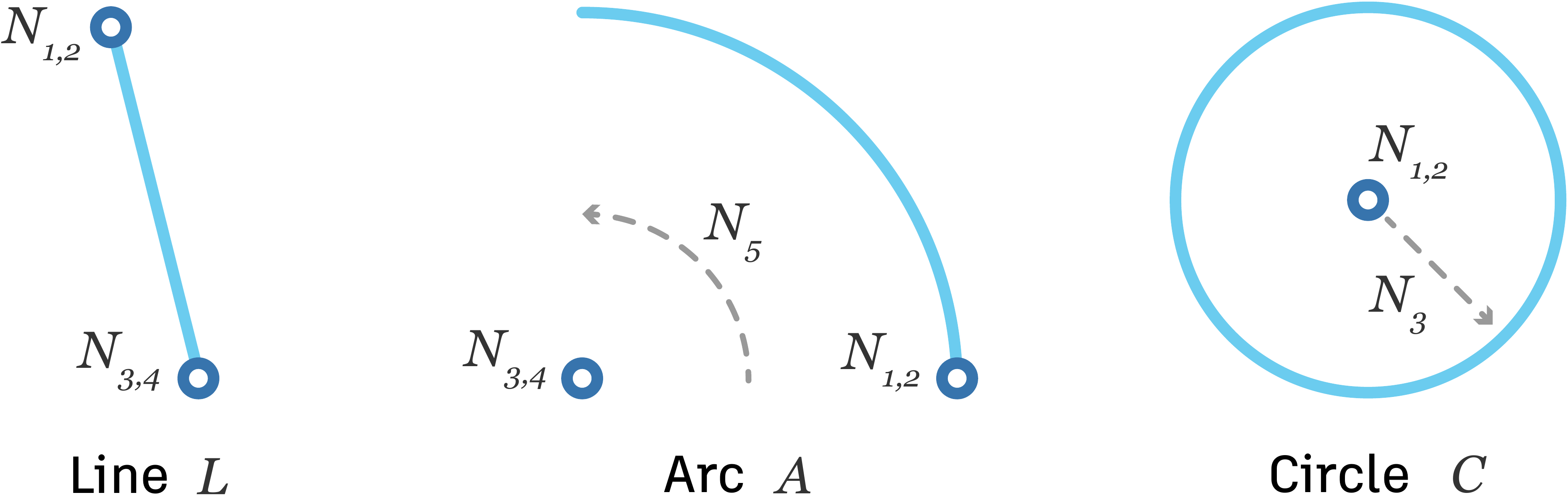}
        \caption{Sketch commands used to create a Line $L$, Arc $A$, and Circle $C$.}
        \label{figure:dataset_dsl_entities}
    \end{center}
\end{figure}

\subsection{Sketch}
\label{section:sketch}
A sketch operation, $S$, is stated by specifying the plane on which the sketch will be created using the $\text{add\_sketch}(I)$ command.  $I$ is a plane identifier, which allows for identification of the three canonical planes $XY,YZ,XZ$ along with other planar faces present in the current geometry $G$. Following the identification of a sketch plane, one can add a sequence of sketch commands $[D]$, where each command is either a line $L$, arc $A$, or circle $C$ (Figure~\ref{figure:dataset_dsl_entities}). Line, arc, and circle represent 95\% of curves in the dataset. A line command $L$ is specified by four numbers, representing the coordinates for the start and end points. A circle command $C$ is specified by three numbers, two representing the circle's center and one representing its radius.  An arc command $A$ is specified by five numbers, representing the start point, the arc's center point, and the angle which the arc subtends. The coordinates for the line $L$, arc $A$, and circle $C$ are specified with respect to the coordinate system of the chosen sketch plane $I$ in $G$. Executing a sketch $S$ command creates a list of new profiles in the current geometry $G$, consisting of enclosed regions resulting from the sketch.

\begin{figure}
    \begin{center}
        \includegraphics[width=0.85\columnwidth]{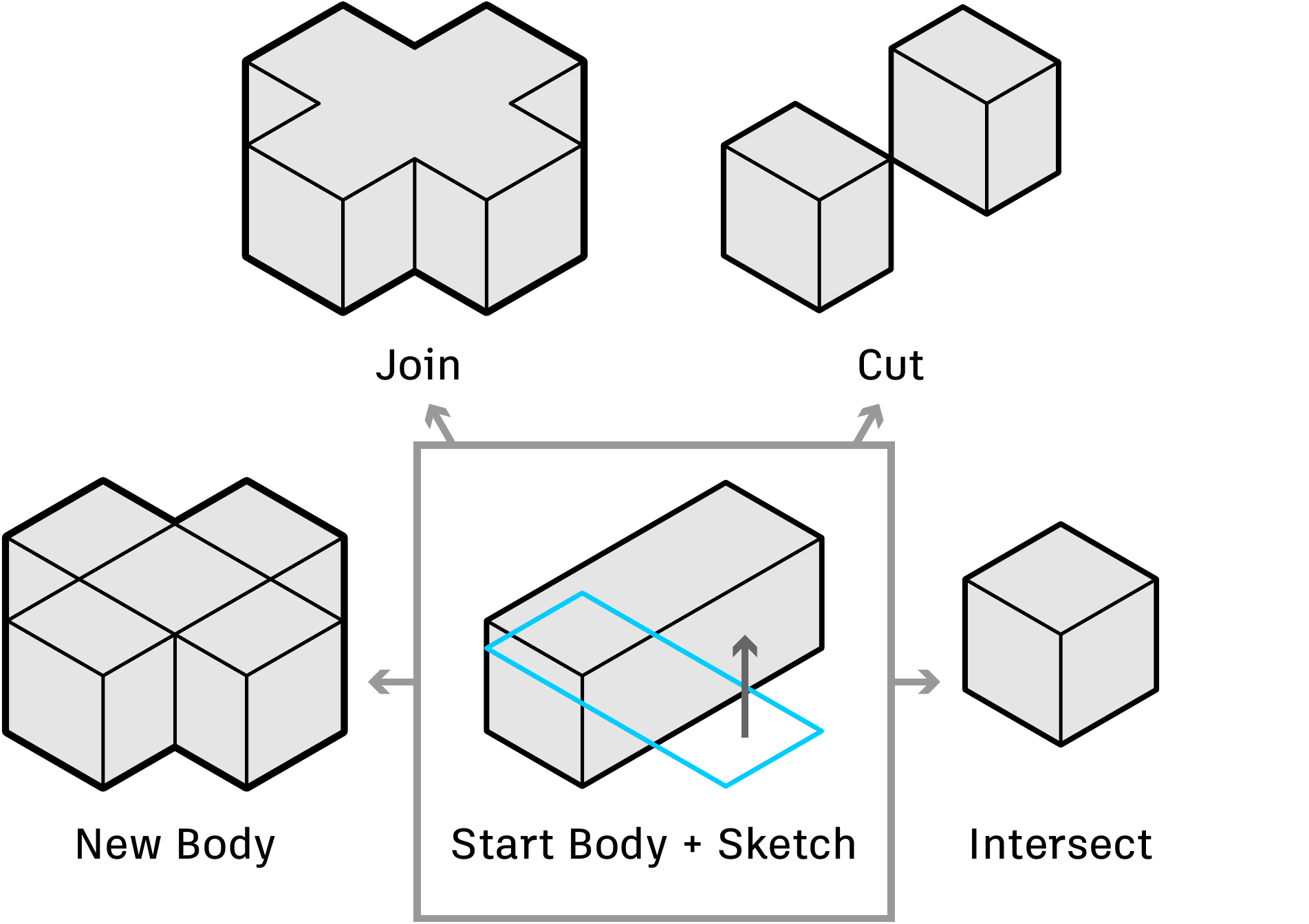}
        \caption{Extrude operations include the ability to Boolean with other geometry. From the start body shown in the center, a sketch is extruded to form a new body overlapping the start body, join with the start body, cut out of the start body, or intersect with the start body.}
        \label{figure:dataset_extrude_operations}
    \end{center}
\end{figure}

\subsection{Extrude}
\label{section:extrude}
An extrude operation $E$ takes a list of identifiers, $[I]$, referencing a list of profiles in the current geometry $G$, and extrudes them from 2D into 3D. A signed distance parameter $N$ defines how far the profile is extruded along the normal direction. The Boolean operation $O$ specifies whether the extruded 3D volume is added to, subtracted from, or intersected with  other 3D bodies in the design. 
Figure~\ref{figure:dataset_extrude_operations} shows a start body and sketch (center) that is extruded to form two separate overlapping bodies, joined to form a single body, cut through the start body to split it in two, or intersected with the start body. Additional extrude options are available such as two-sided extrude, symmetrical extrude, and tapered extrude (See Section~\ref{section:appendix_dataset_extrude} of the appendix). Executing an extrude operation $E$ results in an updated list of bodies in the current geometry $G$. The combination of expressive sketches and extrude operations with built in Boolean capability enables a wide variety of designs to be constructed from only two modeling operations (Figure~\ref{figure:teaser}).

\section{Fusion 360 Gym}
Together with the dataset we provide an open source environment, called the \FGym{}, for standardizing the CAD reconstruction task for learning-based approaches. The \FGym{} further simplifies the \FGal{} DSL and serves as the environment that interacts with an intelligent agent for the task of CAD reconstruction (Figure~\ref{figure:fusion_gym}). Just as a designer can iteratively interact with a CAD software system in a step-by-step fashion, comparing at each step the target geometry to be recovered and the current geometry they have created so-far, the \FGym{} provides the intelligent agent with the same kind of interaction. Specifically, the \FGym{} formalizes the following Markov Decision Process:

\begin{figure}
    \begin{center}
    \includegraphics[width=\columnwidth]{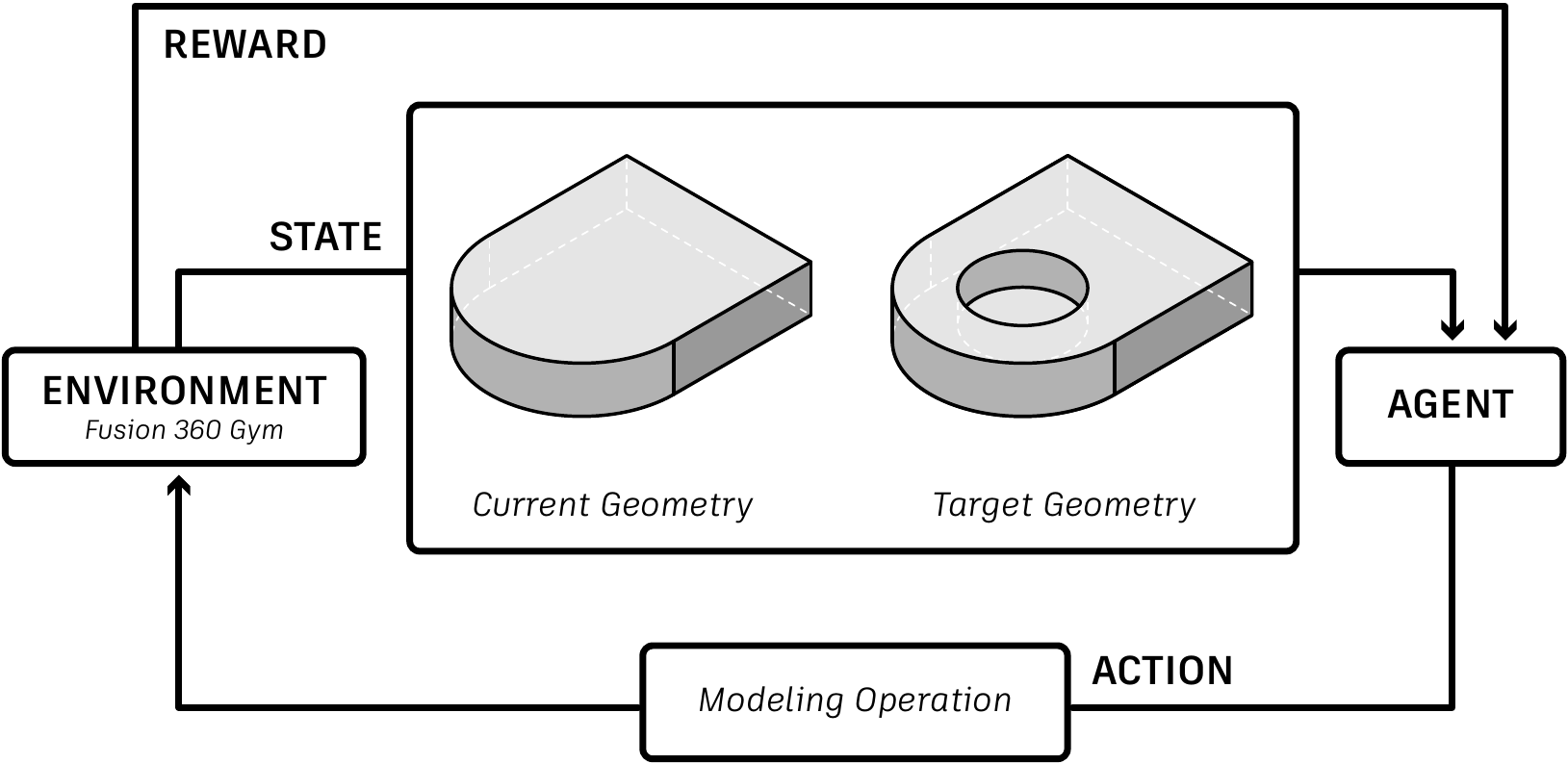}
    \caption{The \FGym{} environment interacts with an \textit{agent} in a sequential decision making scenario. The \textit{state} contains the current and target geometries. The agent outputs an action, in the form of a modeling operation, that advances the current geometry towards the target. }
    \label{figure:fusion_gym}    
    \end{center}
\end{figure}

\setlength{\leftmargini}{0.75cm}
\begin{itemize}
    \item \textbf{state}: Contains the current geometry, and optionally, the target geometry to be reconstructed. We use a B-Rep face-adjacency graph as our state representation.
    \item \textbf{action}: A modeling operation that allows the agent to modify the current geometry. We consider two action representations: sketch extrusion and face extrusion.
    \item \textbf{transition}: \FGym{} implements the transition function that applies the modeling operation to update the current geometry.
    \item \textbf{reward}: The user can define custom reward functions depending on the task. For instance, the agent might receive a reward of $1$ if the current geometry exactly matches the target geometry.
\end{itemize}

\subsection{State Representation}
\label{subsection:state_representation}
\begin{figure}
    \begin{center}
    \includegraphics[width=0.65\columnwidth]{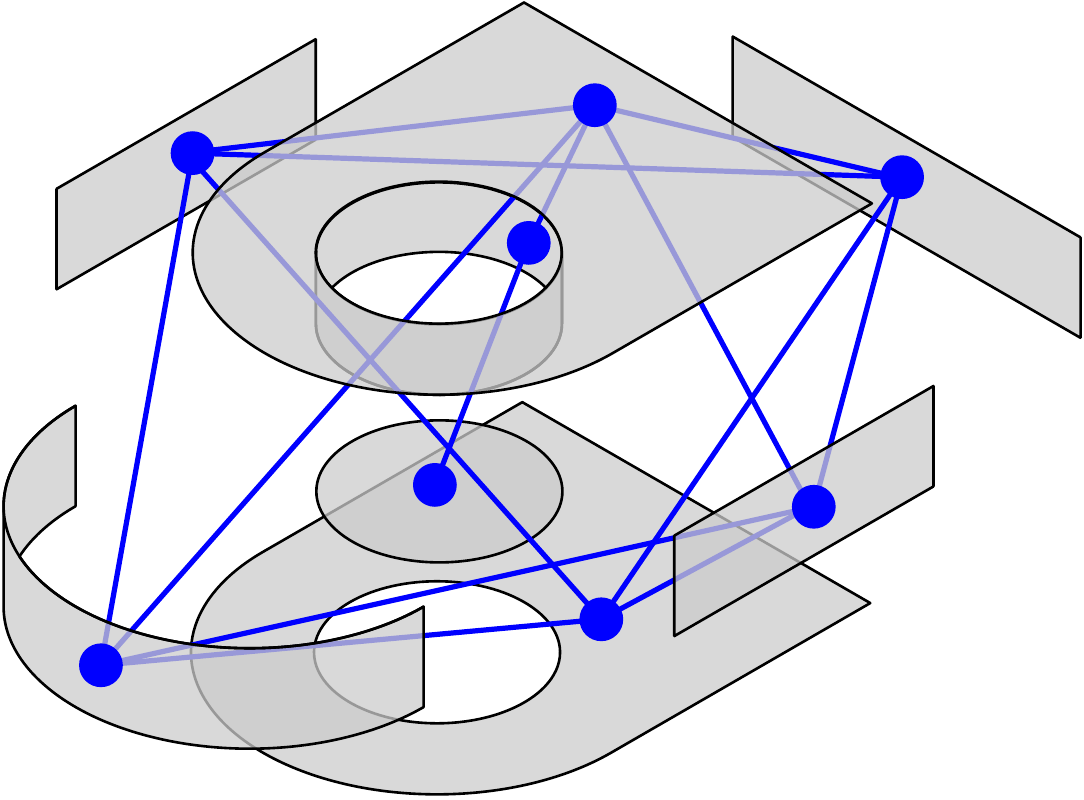}
    \caption{For state representation we use a face adjacency graph with B-Rep faces as graph vertices and B-Rep edges as graph edges.}
    \label{figure:face_adjacency_graph}    
    \end{center}
\end{figure}

In order for an agent to successfully reconstruct the target geometry, it is important that we have a suitable state representation. In the \FGym{}, we use a similar encoding scheme to \citet{jayaraman2020uvnet} and represent the current and target geometry with a B-Rep face-adjacency graph \citep{ansaldi1985geometric}, which contains additional information amenable to a learning agent not present in the language of the \FGal{} DSL (Figure~\ref{figure:face_adjacency_graph}). Crucial to this encoding are the \emph{geometric} features of the elements, such as point-locations, and \emph{topological} features specifying how these elements are connected to each other. Specifically, the vertices of the face-adjacency graph represent B-Rep faces (trimmed parametric surfaces) in the design, with graph vertex features representing the size, orientation, and curvature of the faces. The edges of the face-adjacency graph represent B-Rep edges in the design, that connect the adjacent B-Rep faces to each other. Additional details are provided in Section~\ref{section:appendix_agent} of the appendix. 

\begin{figure*}
    \includegraphics[width=\textwidth]{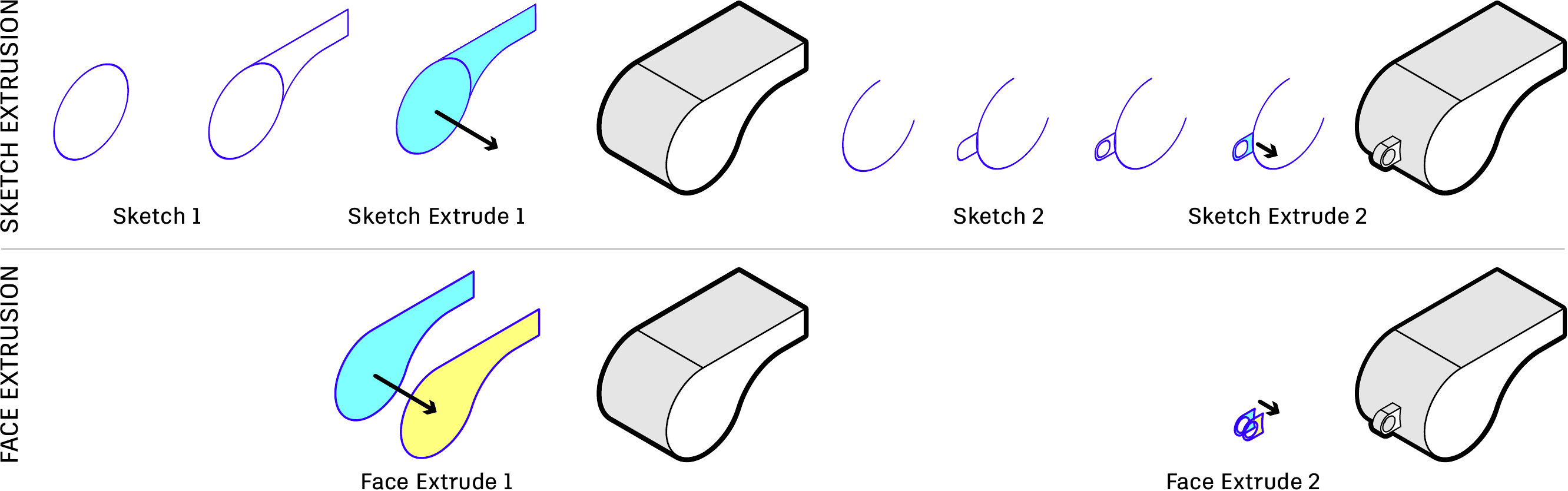}
    \caption{Action representations supported by the \textit{Fusion 360 Gym} include low-level sketch extrusion (top) and simplified face extrusion (bottom).}
    \label{figure:face_extrusion_graph}
\end{figure*}

\subsection{Action Representation}
\label{subsection:action_representation}

In the \FGym{} we support two action representations encompassing different modeling operations: \textit{sketch extrusion} and \textit{face extrusion}.

\subsubsection{Sketch Extrusion}
Sketch extrusion mirrors the \FGal{} DSL closely. In this scheme, the agent must first select a sketch plane, draw on this plane using a sequence of curve primitives, such as lines and arcs, to form closed loop profiles. The agent then selects a profile to extrude a given distance and direction (Figure~\ref{figure:face_extrusion_graph}, top). Using this representation it is possible to construct novel geometries by generating the underlying sketch primitives and extruding them by an arbitrary amount.
Although all designs in the \FRec{} can be constructed using sketch extrusion, in practice this is challenging.  \citet{Benko02constrainedfitting} show that to generate sketches suitable for mechanical engineering parts, the curve primitives often need to be constructed alongside a set of constraints which enforce regularities and symmetries in the design.  Although the construction of constraint graphs is feasible using techniques like the one shown by \citet{liao2019efficient}, enforcing the constraints requires a complex interaction between the machine learning algorithm and a suitable geometric constraint solver, greatly increasing the algorithm complexity. We alleviate this problem by introducing a simplified action representation, called \textit{face extrusion}, that is well suited to learning-based approaches.

\subsubsection{Face Extrusion}
 In face extrusion, a face from the target design is used as the extrusion profile rather than a sketch profile  (Figure~\ref{figure:face_extrusion_graph}, bottom). This is possible because the target design is known in advance during reconstruction. An action $a$ in this scheme is a triple $\{\mathrm{face}_{start}, \mathrm{face}_{end}, \mathrm{op}\}$ where the start and end faces are parallel faces referenced from the target geometry, and the operation type is one of the following: \textit{new body, join, cut, intersect}. The start face defines the extrusion profile and the end face defines the distance to be extruded and does not need to match the shape of the start face.
Target constrained reconstruction using face extrusion has the benefit of narrowly scoping the prediction problem with shorter action sequences and simpler actions. Conversely, not all geometries can be reconstructed with this simplified strategy due to insufficient information in the target, e.g., Extrude 3 in Figure~\ref{figure:construction_seq} cuts across the entire design without leaving a start or end face.

\subsection{Synthetic Data Generation}
\label{section:synthetic_data}
The \FGym{} supports generation of synthetic designs for data augmentation. In addition to procedurally generated synthetic data, semi-synthetic data can be generated by taking existing designs and modifying or recombining them. For instance, we can randomly perturb the sketches and the extrusion distances, and even `graft' sketches from one design onto another. We also support distribution matching of parameters, such as the number of faces, to ensure that synthetic designs match a human designed dataset distribution. Learning-based systems can leverage semi-synthetic data to expand the number of samples in the \FRec{}. In Section~\ref{section:augmentation_comparison} we evaluate the performance of synthetic and semi-synthetic data for the CAD reconstruction task.
We provide examples of synthetic data in Figure~\ref{figure:data_augmentation}  and commands for the \FGym{} in Section~\ref{section:gym} of the appendix.

\section{CAD Reconstruction Task}
\label{section:cad_reconstruction}

\begin{figure*}
    \includegraphics[width=\textwidth]{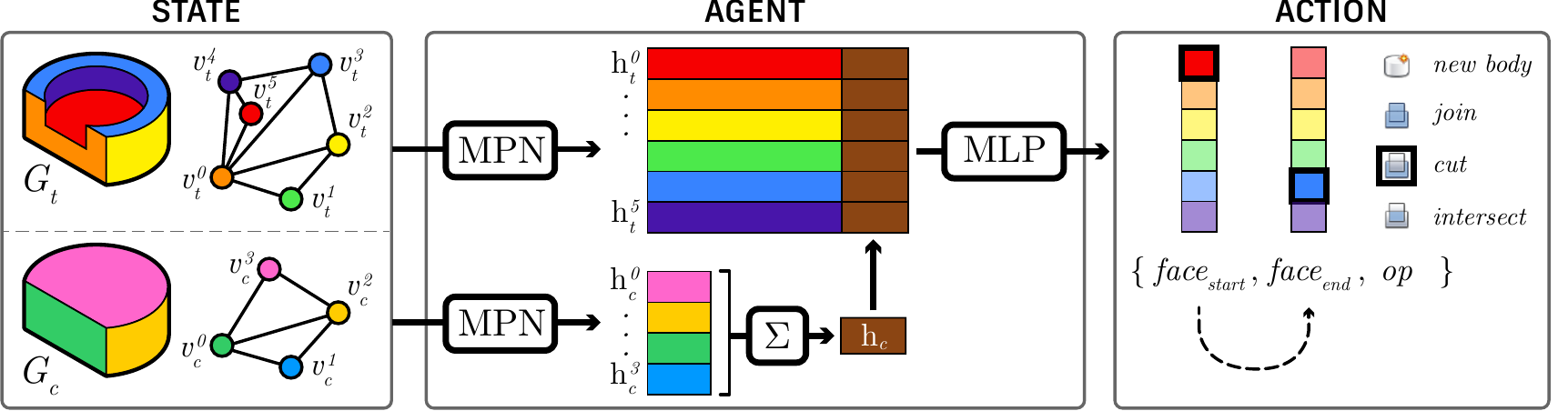}
    \caption{Given a \textit{state} comprising the target geometry $G_t$ and current geometry $G_c$, both of which are represented as a graph, the \textit{agent} uses message passing networks (MPNs) to predict an \textit{action} as a face extrusion modeling operation. The first MPN in the bottom branch produces a set of node embedding vectors $\mathbf{h}_c^0 \cdots \mathbf{h}_c^3$, which are summed over to produce the hidden vector for the current geometry $\mathbf{h}_c$. Another MPN in the top branch produces a set of node embedding vectors $\mathbf{h}_t^0 \cdots \mathbf{h}_t^5$, which are concatenated with $\mathbf{h}_c$ to predict the \textit{action}. We condition the end face prediction on the predicted start face. Colors in the figure correspond to different graph nodes.}
    \label{figure:agent_arch}
\end{figure*}

\subsection{Task Definition}
The goal of CAD reconstruction is to recover a program, represented as a sequence of modeling operations used to construct a CAD model with only the geometry as input. This task can be specified using different input geometry representations, including B-Rep, mesh, or point cloud, with progressively lower fidelity. Each representation presents a realistic scenario where parametric CAD information is absent and needs to be recovered. Given a target geometry $G_t$, we wish to find a sequence of CAD modeling operations (actions)
$\mathcal{A} = \{a_0, a_1, \cdots\}$
such that, once executed in a CAD software system, results in a geometry $G$
where every point in space is in its interior, if and only if, it is also in the interior of $G_t$.

\subsection{Evaluation Metrics} We prescribe three evaluation metrics, IoU, exact reconstruction, and conciseness. IoU measures the intersection over union of $G$ and $G_t$: 
$\mathtt{iou}(G,G_t) = |G \cap G_t|/|G \cup G_t|$.
Exact reconstruction measures whether $\mathtt{iou}(G,G_t) = 1$. 
As multiple correct sequences of CAD modeling operations exist, a proposed reconstruction sequence $\mathcal{A}$ need not match the ground truth sequence $\hat{\mathcal{A}}_t$ provided an exact reconstruction is found. 
To measure the quality of exact reconstructions we consider the conciseness of the construction sequence.
Let $\mathtt{conciseness}(\mathcal{A}, \hat{\mathcal{A}}_t) = |\mathcal{A}| / |\hat{\mathcal{A}}_t|$, where a score $\leq 1$ indicates the agent found an exact reconstruction with equal or fewer steps than the ground truth, and a score $> 1$ indicates more inefficient exact reconstructions.

\subsection{Neurally Guided Search}
\label{section:baseline}
We now present a method for CAD reconstruction using neurally-guided search \citep{ellis2019write,kalyan2018neural,tang2019reinforcement,devlin2017robustfill} from \textit{B-Rep input} using \textit{face extrusion} modeling operations. The training phase consists of imitation learning, where a policy is trained to imitate a known construction sequence from a given geometry. The testing / inference phase leverages search, where the search algorithm repeatedly samples the trained policy for actions and applies these actions in the environment to generate a set of candidate reconstruction sequences.

\subsubsection{Imitation Learning} To perform imitation learning, we leverage the fact that we have the ground truth sequence of modeling operations (actions) $\hat{\mathcal{A}}_t = \{\hat{a}_{t,0} \cdots \hat{a}_{t,n-1}\}$ for each design $G_t$ in the dataset. We feed the ground truth action sequence $\hat{\mathcal{A}}_t$ into the \FGym{}, starting from the empty geometry $G_0$, and output a sequence of partial constructions $G_{t,1} \cdots G_{t,n}$ where $G_{t,n}=G_t$. We then collect the supervised dataset $\mathcal{D} = \{(G_0,G_t) \rightarrow \hat{a}_{t,0}, (G_{t,1},G_t) \rightarrow \hat{a}_{t,1} \cdots \}$ and train a supervised agent $\pi_\theta$ that takes the pair of current-target constructions $(G_c,G_t)$ to a modeling operation action $a_c$, which would transform the current geometry closer to the target. Formally, we optimize the expected log-likelihood of correct actions under the data distribution:
\begin{align}
E_{(G_c,G_t) \sim \mathcal{D}} \bigg[\log \pi_\theta \Big(\hat{a}_c|\big(G_c,G_t\big)\Big)\bigg] 
\end{align}

\subsubsection{Agent} The agent (Figure~\ref{figure:agent_arch}) takes a pair of geometries $(G_c,G_t)$ as state, and outputs the corresponding face-extrusion action $a = \{\mathrm{face}_{start}, \mathrm{face}_{end}, \mathrm{op}\}$.
The two geometries $G_c,G_t$ are given using a face-adjacency graph similar to \citet{jayaraman2020uvnet}, where the graph vertexes represent the faces of the geometry, with vertex features calculated from each face: 10$\times$10 grid of 3D points, normals, and trimming mask, in addition to the face surface type. The 3D points are global xyz values sampled in UV parameter space of the face. The edges define connectivity of adjacent faces. Inputs are encoded using two \emph{separate} message passing networks~\cite{kipf2016semi, gilmer2017neural, kipf2018neural} aggregating messages along the edges of the graph. The encoded vectors representing the \emph{current} geometry are summed together ($\mathbf{h}_c$ in Figure~\ref{figure:agent_arch}), and concatenated with the encoded vertexes of the target geometry ($\mathbf{h}_t^0 \cdots \mathbf{h}_t^5$ in Figure~\ref{figure:agent_arch}). The concatenated vectors are used to output the action using a multi-layer perceptron (MLP), with the end face conditioned on the vertex embedding of the predicted start face. 

We denote the learned vertex embedding vectors produced by the two MPN branches as $\{ \mathbf{h}_c^i \}$ and $\{ \mathbf{h}_t^j \}$ for the current output and target graphs, respectively. We estimate the probability of the $k$-th operation type, and the $j$-th face being the start face or end face as:
\begin{gather}
P^k_{op} = F_{op} \big( \mathbf{h}_c \big), \:\:  \mathbf{h}_c = \sum_i\mathbf{h}_c^i\\
P^j_{start} = \underset{j}{\mathtt{softmax}} \Big( F_{start} \big( \mathbf{h}_t^j, \mathbf{h}_c \big) \Big)\\
P^j_{end} = \underset{j}{\mathtt{softmax}} \Big( F_{end} \big( \mathbf{h}_t^j, \mathbf{h}_t^{\tilde{j}}, \mathbf{h}_c \big) \Big), \:\: s.t. \:\: \tilde{j} = \underset{j}{\mathtt{argmax}} P^j_{start}
\end{gather}
where $F_{op}$, $F_{start}$, and $F_{end}$ denote linear layers that take the concatenated vectors as input.

\subsubsection{Search} Given a neural agent $\pi_\theta(a|(G_c,G_t))$ capable of furthering a current geometry toward the target geometry, we can amplify its performance at test time using search. This allows us to explore multiple different reconstruction sequences at once, at the expense of extended interactions with the environment. By leveraging search, one gets the benefit of scaling: the larger budget we have to interact with the environment, the more likely we are going to succeed in recovering a working reconstruction sequence. The effectiveness of search is measured against a \emph{search budget}, which in our case, is the number of \emph{environment steps} executed in the \FGym{}. We consider the following standard search procedures from the neurally guided search literature:
\begin{itemize}
    \item \textbf{random rollouts}: This search procedure uses the learned policy to sample a sequence of steps in the environment. Every rollout consists of $N$ iterations; at each iteration an action is chosen according to $\pi_\theta$. This action is executed in the environment by taking an environment step and the updated current geometry is presented back to the policy to sample the next action. $N$ is capped to a fixed rollout length of $\max(\frac{f_p}{2}, 2)$, where $f_p$ is the number of planar faces in $G_t$. If the agent fails to recover the target geometry in the current roll-out, we restart with a new roll-out and repeat the process.
    
    \item \textbf{beam search}: We rollout in parallel the top-k (where k is the beam width) candidate construction sequences for $N$ iterations. Each sequence is ranked by the generation probability under $\pi_\theta$, $P_\theta(a_1 \dots a_r)$:
    
    \begin{equation*}
        P_\theta(a_1 \dots a_r) = \prod_{i = 1 \dots r} \pi_\theta(a_i | G_i, Gt)
    \end{equation*}

    At each iteration, we consider all possible extensions to the top-k candidates by one action under $\pi_\theta$, and re-rank the extended candidate sequences under $P_\theta$, keeping the top-k extended candidates. Then, for each of the $k$ extended sequences, we execute a step in the environment to obtain the updated current geometries. Each run of the beam search results in $kN$ environment steps. If the current $k$ sequences reaches the rollout length without recovering the target geometry, the beam search restarts with the beam width doubled, allowing it to search a wider range of candidates.
    
    \item \textbf{best first search}: This search procedure explores the search space by maintaining a priority queue of candidate sequences, where the priority is ordered by $P_\theta$. At each iteration, we dequeue the top candidate sequence and extend it by one action under $\pi_\theta$, and these extended sequences are added back to the queue. An environment step is taken in a lazy fashion when the top candidate sequence is dequeued, and not when the extended sequences are added back to the queue. This process continues until the dequeued top candidate recovers the target geometry. 
\end{itemize}

\section{Evaluation}

\begin{table}
    \small
    \caption{Reconstruction results for IoU and exact reconstruction at 20 and 100 environment steps using random rollouts with different agents trained on human designed data. The best result in each column is shown in bold. Lower values are better for conciseness.}
    \label{tab:benchmark}
    \centering
    \begin{tabular}{l|ccccc}
    \toprule
        \textbf{Agent}          & \multicolumn{2}{c}{\textbf{IoU}} & \multicolumn{2}{c}{\textbf{Exact Recon. \%}} & \textbf{Concise.}\\
        {} & 20 Steps &    100 Steps &        20 Steps &  100 Steps & {}\\
    \midrule
      gat &   \textbf{0.8742} &    \textbf{0.9128} &                 0.6191 &    0.6742 &      1.0206\\
      gcn &   0.8644 &    0.9042 &                 \textbf{0.6232} &    \textbf{0.6754} &      1.0168\\
      gin &   0.8346 &    0.8761 &                 0.5901 &    0.6301 &      1.0042\\
      mlp &   0.8274 &    0.8596 &                 0.5658 &    0.5965 &      \textbf{0.9763}\\
     rand &   0.6840 &    0.8386 &                 0.4157 &    0.5380 &      1.2824\\
    \bottomrule
    \end{tabular}
\end{table}

\begin{figure}
    \includegraphics[width=\columnwidth]{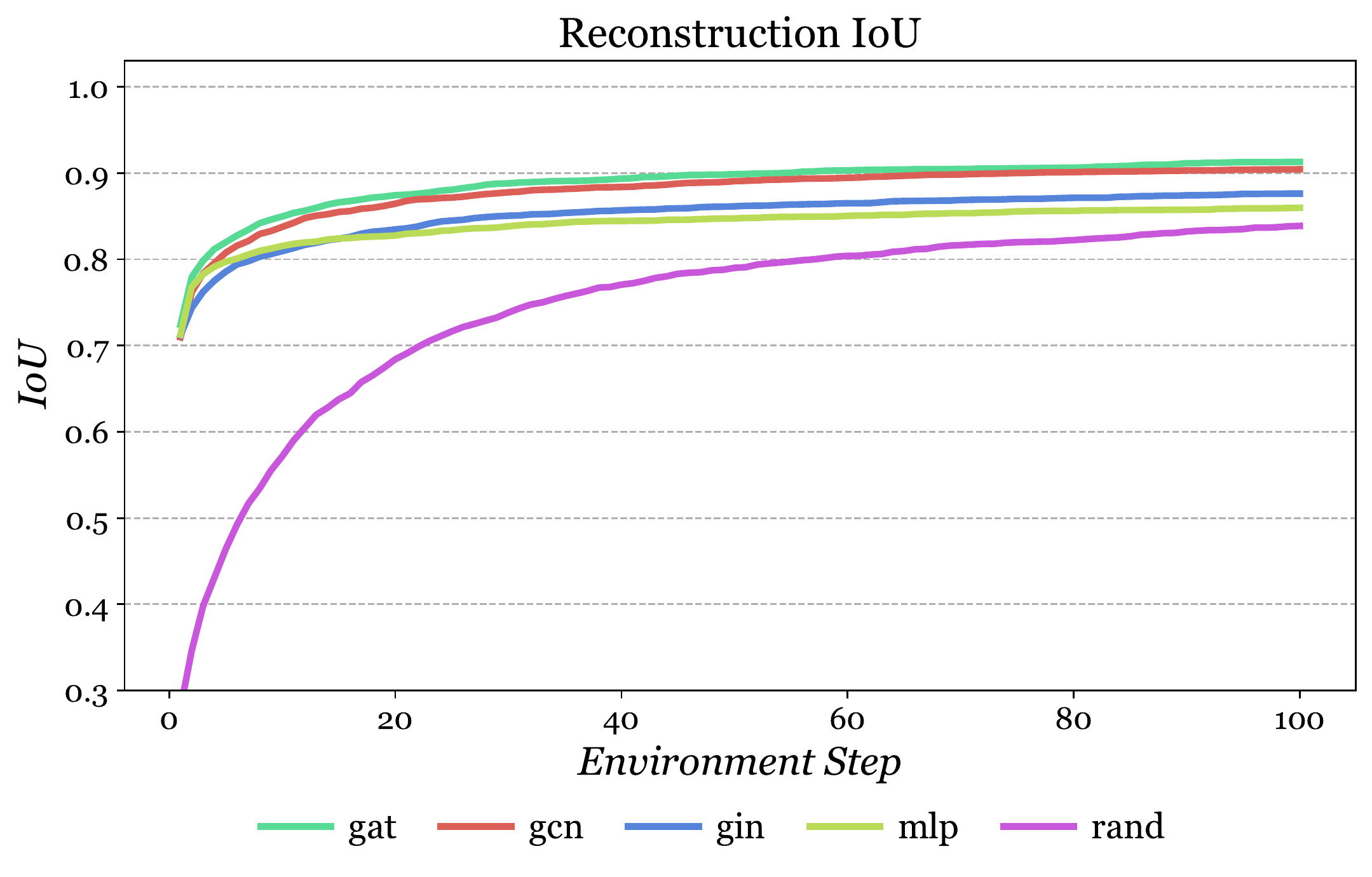}
    \caption{ Reconstruction IoU over 100 environment steps using random rollouts with different agents trained on human designed data.}
    \label{figure:reconstruction_results_iou}
\end{figure}

\begin{figure}
    \includegraphics[width=\columnwidth]{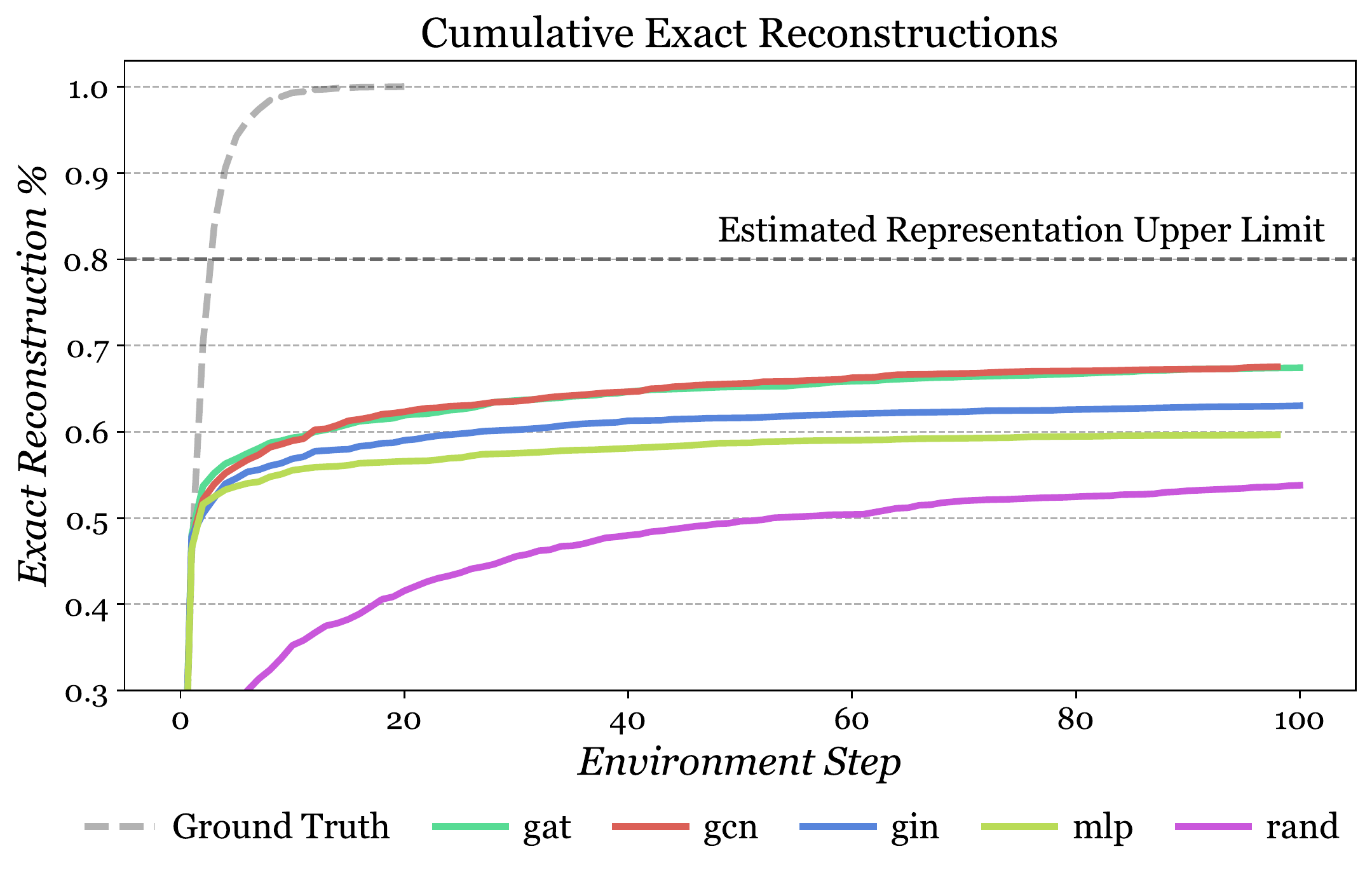}
    \caption{ Cumulative exact reconstructions over 100 environment steps using random rollouts with different agents trained on human designed data. The estimated upper limit of the face extrusion action representation is shown at $0.8$.}
    \label{figure:reconstruction_results_exact}
\end{figure}

We proposed a general strategy consisting of neurally guided search, powered by a neural-network trained via imitation on human designed, synthetic, and augmented data. To justify this strategy, we perform ablation studies, comparing our approach against a set of baselines on the \FRec{}. We seek to answer the following:
\begin{itemize}
    \item How do different neural representations, when used to represent the agent's policy $\pi_\theta$, perform on the CAD reconstruction task?
    \item How does training a neural policy under human designed, synthetic, and augmented data affect CAD reconstruction performance?
    \item How do different neurally guided search procedures from the literature perform on the CAD reconstruction task?
\end{itemize}

For evaluation, we track the best IoU the agent has discovered so far, and whether exact reconstruction is achieved as a function of environment steps. We cap the total search budget to $100$ steps to reflect a real world scenario. For experiments using human design data we train on the 59.2\% of the training set that can be directly converted to a face extrusion sequence. We evaluate on the full test set in all cases. We estimate that approximately 80\% of designs in our dataset can be reconstructed by finding alternative face extrusion sequences and note this when reporting exact reconstruction results.

\subsection{Comparing Different Neural Representations}
\label{section:baseline_comparison}
We evaluate five different kinds of neural network representations for $\pi_\theta$ to understand how different networks perform on the CAD reconstruction task. The \textbf{rand} agent uniformly samples from the available actions to serve as a naive baseline without any learning.
\textbf{mlp} is a simple agent using a MLP that does not take advantage of message passing via graph topology. \textbf{gcn}, \textbf{gin}, and \textbf{gat} are MPN agents that use a Graph Convolution Network \citep{kipf2016semi},  Graph Isomorphism Network \citep{xu2018powerful}, and Graph Attention Network \citep{velivckovic2017graph} respectively.
We use two MPN layers for all comparisons, with standard layer settings as described in Section~\ref{section:appendix_agent} of the appendix.

We report the reconstruction IoU and exact reconstructions using random rollout search for each agent as a function of the number of environment steps in Figure~\ref{figure:reconstruction_results_iou} and \ref{figure:reconstruction_results_exact} respectively. We detail the exact results at step 20 and 100 in Table~\ref{tab:benchmark}. Step 20 represents the point where it is possible to perform exact reconstructions for all designs in the test set. We also detail the conciseness of the recovered sequence for exact reconstructions.
We note that all neurally guided agents outperform the random agent baseline. The topology information available with a MPN is found to improve reconstruction performance. The gat and gcn agents show the best performance but fall well short of exact reconstruction on all designs in the test set, demonstrating that the CAD reconstruction task is non-trivial and an open problem for future research.

\subsection{Comparing Human and Synthetic Data Performance}
\label{section:augmentation_comparison}
\begin{figure}
    \includegraphics[width=\columnwidth]{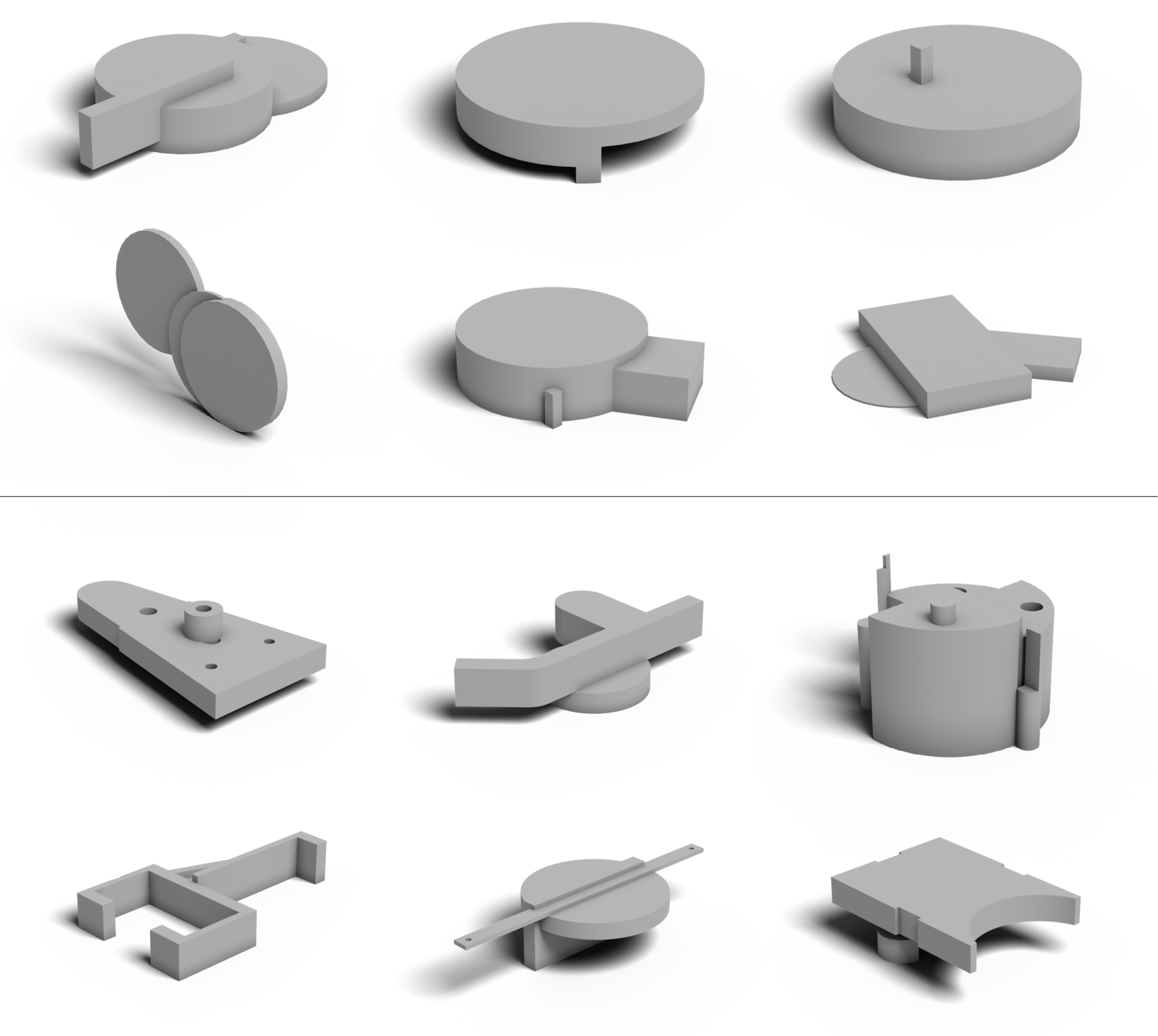}
    \caption{Top: example \textit{synthetic} data created by extruding circles and rectangles. Bottom: example \textit{semi-synthetic} data created by extruding human designed sketches.}
    \label{figure:data_augmentation}
\end{figure}

We evaluate four gcn agents trained on different data sources to understand how synthetic data performs compared to human design data.
\textbf{real} is trained on the standard human design training set. 
\textbf{syn} is trained on synthetic data from procedurally generated sketches of rectangles and circles extruded randomly (Figure~\ref{figure:data_augmentation}, top). Leveraging basic primitives is a common method to generate synthetic data for program synthesis~\citep{li2020sketch2CAD, sharma2017csgnet, ellis2019write}, that typically results in less sophisticated designs compared to human design data. 
\textbf{semi-syn} is trained on semi-synthetic designs that use existing sketches in the training set with two or more extrude operations to match the distribution of the number of faces in the training set (Figure~\ref{figure:data_augmentation}, bottom). This approach results in more complex designs than the pure synthetic designs. We deliberately use these two approaches for data generation to better compare human design data to synthetic data in different distributions.  
\textbf{aug} is trained on the human design training set mixed with additional semi-synthetic data.
We hold the training data quantity constant across agents, with the exception of the aug agent that contains a larger quantity from two sources. All agents are evaluated on the standard human design test set.

\begin{table}[!t]
    \small
    \caption{ Reconstruction results for IoU and exact reconstruction at 20 and 100 environment steps using random rollouts and gcn agents trained on human designed data (real), a mixture of human designed and semi-synthetic data (aug), semi-synthetic data (semi-syn), and synthetic data (syn). The best result in each column is shown in bold. Lower values are better for conciseness.}
    \label{tab:data_augmentation_table}
    \centering
    \begin{tabular}{c|ccccc}
    \toprule
       \textbf{Agent} & \multicolumn{2}{c}{\textbf{IoU}} & \multicolumn{2}{c}{\textbf{Exact Recon. \%}} & \textbf{Concise.} \\
             {} &    20 Steps &    100 Steps &        20 Steps & 100 Steps & {} \\
    \midrule
     real       &   0.8644 &    \textbf{0.9042} &                 0.6232 &    \textbf{0.6754} &      1.0168\\
      aug       &   \textbf{0.8707} &    0.8928 &                 \textbf{0.6452} &    0.6701 &      \textbf{0.9706}\\
     semi-syn   &   0.8154 &    0.8473 &                 0.5780 &    0.6104 &      1.0070\\
      syn       &   0.6646 &    0.7211 &                 0.4383 &    0.4835 &      1.0519\\
    \bottomrule
    \end{tabular}
\end{table}

\begin{figure}
    \includegraphics[width=\columnwidth]{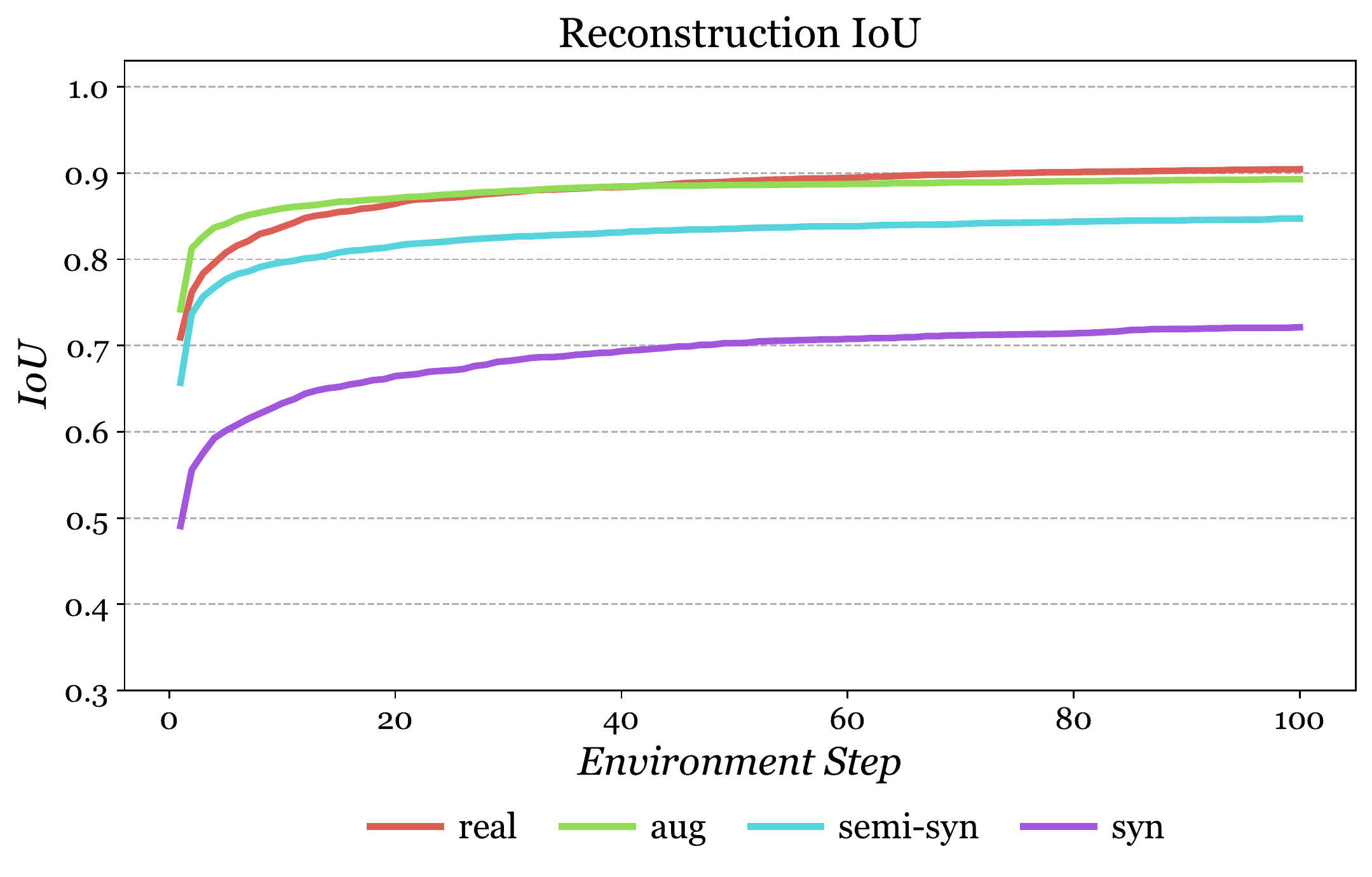}
    \caption{ Reconstruction IoU over 100 environment steps using random rollouts and gcn agents trained on human designed data (real), a mixture of human designed and semi-synthetic data (aug), semi-synthetic data (semi-syn), and synthetic data (syn).}
    \label{figure:reconstruction_results_aug_iou}
\end{figure}

\begin{figure}
    \includegraphics[width=\columnwidth]{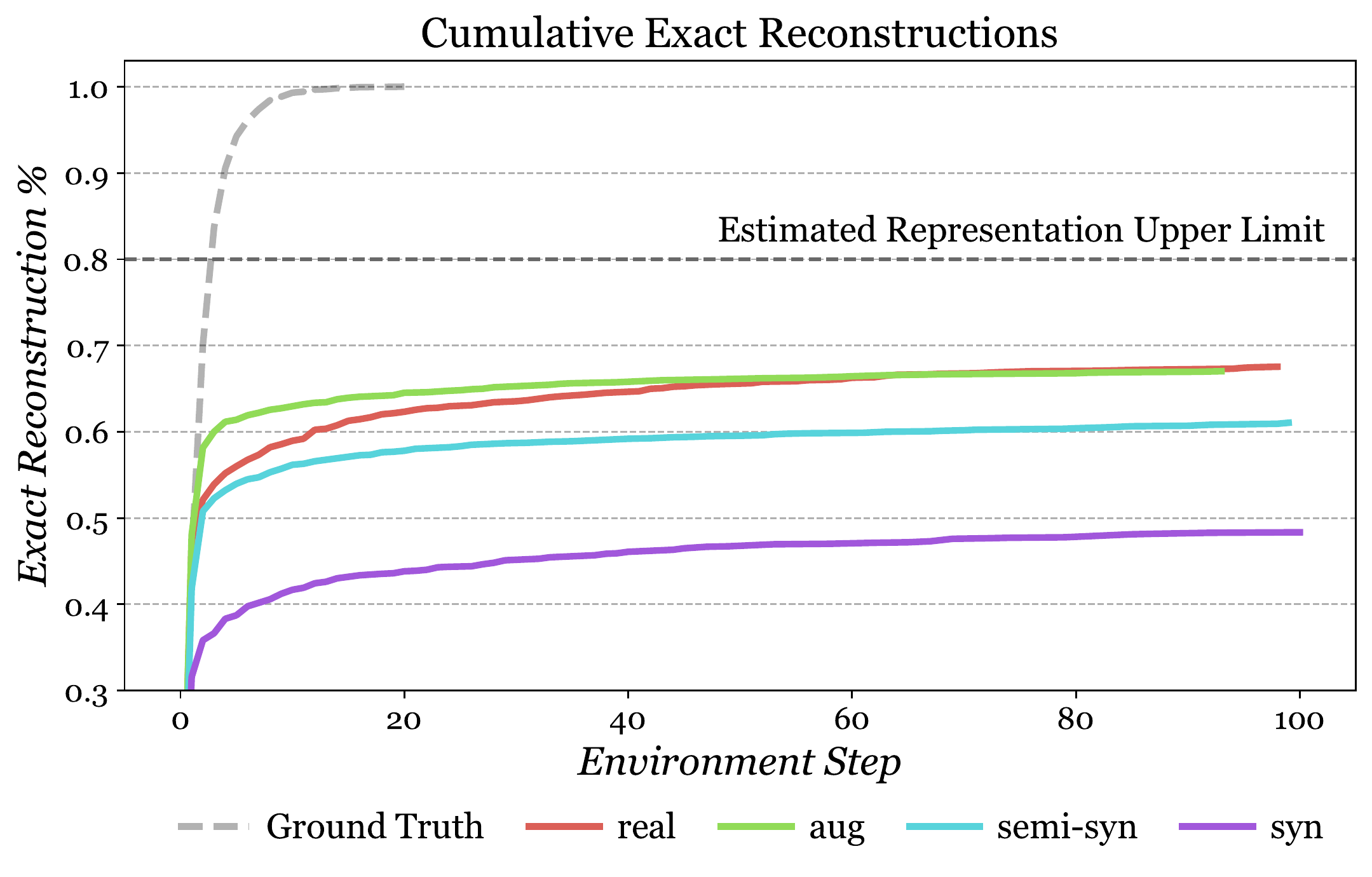}
    \caption{ Cumulative exact reconstructions over 100 environment steps using random rollouts and gcn agents trained on human designed data (real), a mixture of human designed and semi-synthetic data (aug), semi-synthetic data (semi-syn), and synthetic data (syn). The estimated upper limit of the face extrusion action representation is shown at $0.8$.}
    \label{figure:reconstruction_results_aug_exact}
\end{figure}

Figure~\ref{figure:reconstruction_results_aug_iou} and \ref{figure:reconstruction_results_aug_exact} show that training on human design data offers a significant advantage over synthetic and semi-synthetic data for reconstruction IoU and exact reconstructions respectively. 
For the aug agent reconstruction performance is aided early on by data augmentation. We attribute this early performance improvement to semi-synthetic designs with 1 or 2 extrusions appearing similar to human designs. Conversely, we observe that semi-synthetic designs with multiple randomly applied extrusions appear less and less similar to human design due to the random composition of extrusions. This difference in distribution between human and synthetic designs becomes more prevalent as search progressess. Table~\ref{tab:data_augmentation_table} provides exact results at environment step 20 and 100. 

\begin{table}[!t]
    \small
    \caption{ Reconstruction results for IoU and exact reconstruction at 20 and 100 environment steps using gcn agents with best first search (best), random rollout search (rand) and beam search (beam). The best result in each column is shown in bold. Lower values are better for conciseness.}
    \label{tab:benchmark_search_table}
    \centering
    \begin{tabular}{c|ccccc}
    \toprule
       \textbf{Agent} & \multicolumn{2}{c}{\textbf{IoU}} & \multicolumn{2}{c}{\textbf{Exact Recon. \%}} & \textbf{Concise.} \\
             {} &    20 Steps &    100 Steps &        20 Steps & 100 Steps & {} \\
    \midrule
     best &   \textbf{0.8831} &    \textbf{0.9186} &                 0.5971 &    0.6348 &      \textbf{0.9215} \\
     rand &   0.8644 &    0.9042 &                 \textbf{0.6232} &    \textbf{0.6754} &      1.0168 \\
     beam &   0.8640 &    0.8982 &                 0.5739 &    0.6122 &      0.9275 \\
    \bottomrule
    \end{tabular}
\end{table}

\begin{figure}
    \includegraphics[width=\columnwidth]{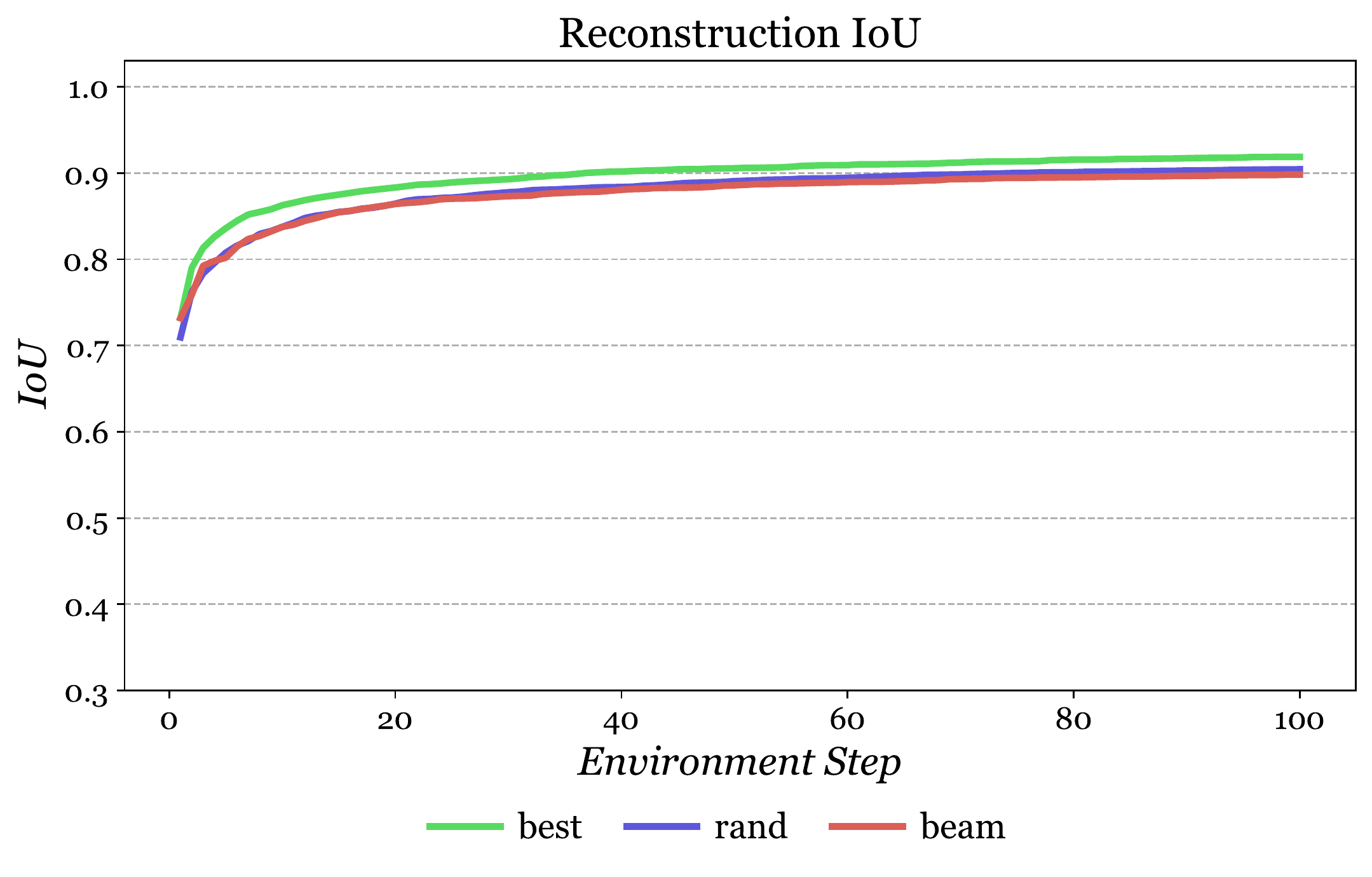}
    \caption{Reconstruction IoU over 100 environment steps using the gcn agent with best first search (best), random rollout search (rand) and beam search (beam).}
    \label{figure:reconstruction_results_search_iou}
\end{figure}

\begin{figure}
    \includegraphics[width=\columnwidth]{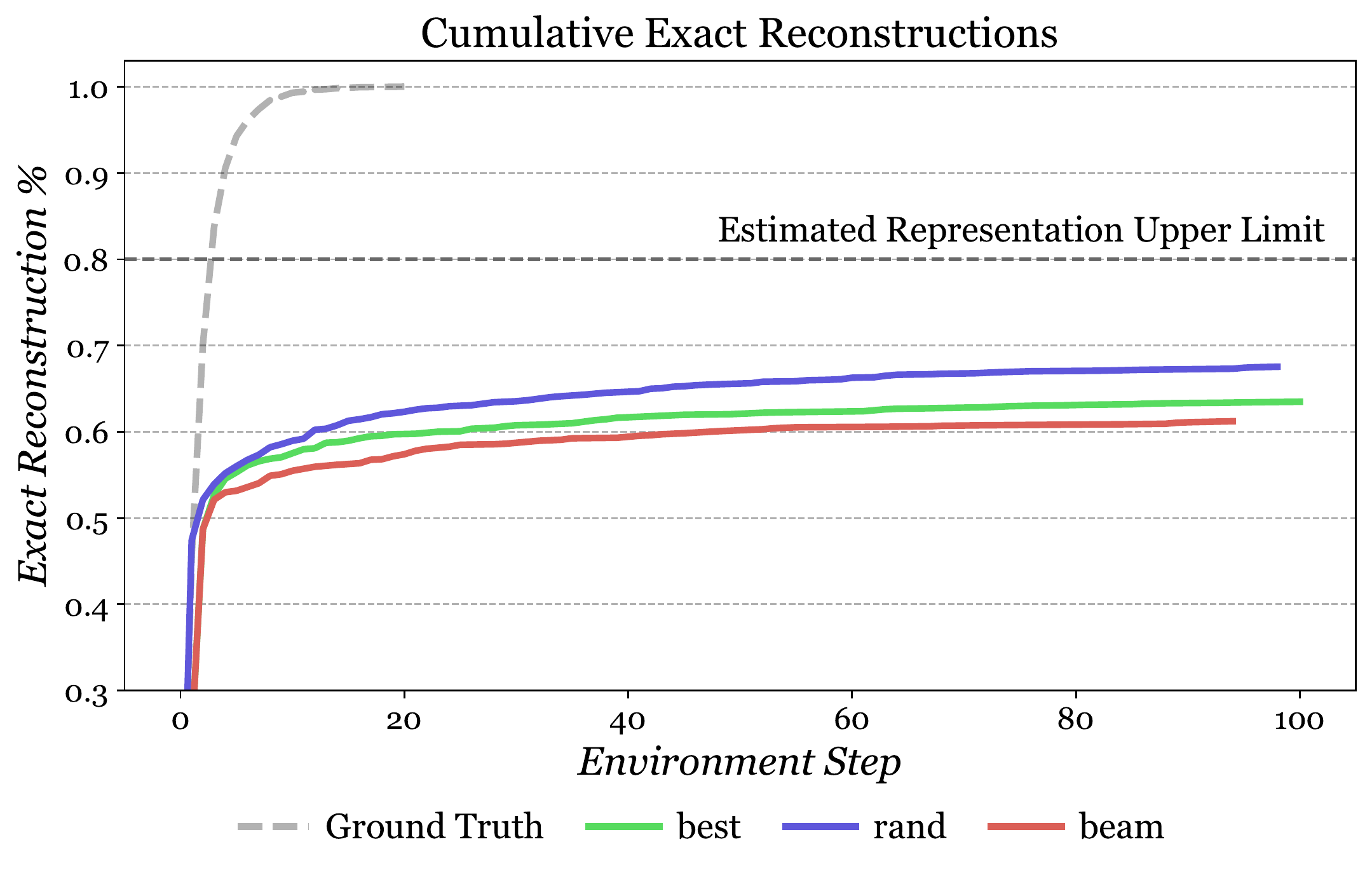}
    \caption{Cumulative exact reconstructions over 100 environment steps using the gcn agent with best first search (best), random rollout search (rand) and beam search (beam). The estimated upper limit of the face extrusion action representation is shown at $0.8$.}
    \label{figure:reconstruction_results_search_exact}
\end{figure}

\subsection{Qualitative Results} 
\begin{figure*}
    \includegraphics[width=\textwidth]{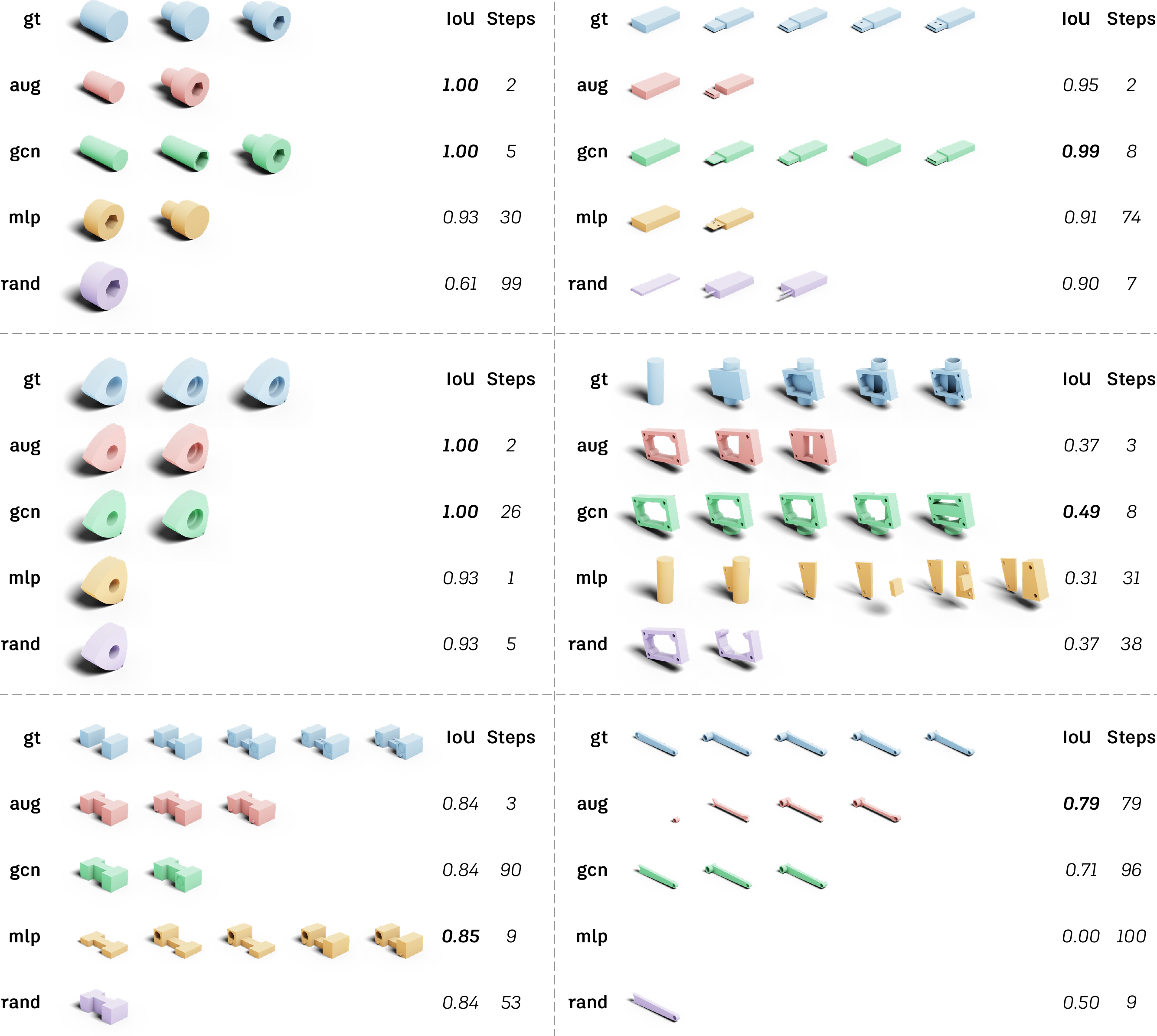}
    \caption{Qualitative construction sequence results comparing the ground truth (\textbf{gt}) to reconstructions using different agents with random rollout search.}
    \label{figure:reconstruction_results_qual}
\end{figure*}

Figure~\ref{figure:reconstruction_results_qual} shows a visualization of ground truth construction sequences compared with the reconstruction results from other agents using random search. The rollout with the highest IoU is shown with the IoU score and total environment steps taken. Steps that don't change the geometry or occur after the highest IoU are omitted from the visualization. 

\subsection{Comparing Search Procedures}
We compare the effects of three different search procedures from the neurally guided search literature. Here, \textbf{rand} is random rollout, \textbf{beam} is beam search, and \textbf{best} is best-first search. For each search algorithm we use the \textbf{gcn} agent described in Section~\ref{section:baseline_comparison} trained on the standard human design training set. 
Figure~\ref{figure:reconstruction_results_search_iou}, \ref{figure:reconstruction_results_search_exact}, and Table~\ref{tab:benchmark_search_table} show that all three search algorithms perform similarly for reconstruction IoU, while rand performs best for exact reconstruction. The performance of rand for exact reconstruction can be explained by the limited search budget of 100 environment steps: the rand algorithm is more likely to sample distinct sequences for a small number of samples, whereas beam will sample half its sequences identical to the previous rounds before the doubling, and best might not be sampled enough to explore a sequence long enough to contain the correct program. 

We expect beam and best to outperform rand as the number of search budget increases, similar to \citet{ellis2019write}. However, the limitation of the search budget is important, as each design in our test set takes between 5-35 seconds to reconstruct on average. The majority of evaluation time is spent inside the \FGym{} executing modeling operations and graph generation, both computationally expensive yet crucial operations that must be taken during reconstruction. 

\subsection{Discussion}
For practical application of CAD reconstruction it is necessary to have an exact reconstruction where all details of a design are reconstructed in a concise way. 
It is notable that incorrect reconstructions can score well with the IoU metric, but omit important design details. For example, the small holes in the USB connector in Figure~\ref{figure:reconstruction_results_qual}b are omitted from the gcn reconstruction. We suggest IoU should be a secondary metric, with future work focusing on improving exact reconstruction performance with concise construction sequences. Conciseness should always be considered alongside exact reconstruction performance as naive approaches that only reconstruct short sequences can achieve good conciseness scores.

\section{Conclusion and Future Directions}
In this paper we presented the \textit{Fusion 360 Gallery} reconstruction dataset and environment for learning CAD reconstruction from sequential 3D CAD data. We outlined a standard CAD reconstruction task, together with evaluation metrics, and presented results from a neurally guided search approach.

\subsection{Limitations}
Our dataset contains only designs created using \textit{sketch} and \textit{extrude} rather than the full array of CAD modeling operations. Short construction sequences make up a sizable portion of the data: 3267/8625 (38\%) of designs have only a single extrude operation. From the single extrude designs, some exhibit more complexity: 347 have $>$1 sketch profile resulting in $\geq$1 bodies from a single extrude operation, and 998 have $\geq$8 sketch curves. Other designs are washers, pegs, and plates, common in mechanical CAD assemblies. We avoid filtering simple designs to ensure the dataset is representative of user-designed CAD. Spline curves represent 4\% of curves in the dataset and are not currently supported by our high-level DSL, however they can be reconstructed via the \code{reconstruct\_curve()} command (Section~\ref{section:reconstruction_commands}).

The success of the \textbf{rand} agent demonstrates that short construction sequences can be solved by a naive approach. This is due to several factors: 1) our action representation that uses B-Rep faces, 2) our search procedure discarding invalid actions, and 3) designs in the dataset with a low number of planar faces and extrude steps. For example, a washer has four B-Rep faces (\textit{planar-top}, \textit{cylinder-inside}, \textit{cylinder-outside}, \textit{planar-bottom}), giving the random agent a 2/2 chance of success as either \textit{planar-top} $\rightarrow$ \textit{planar-bottom}, or vice versa, are considered correct and extrusions from non-planar faces are invalid. Although the random agent can achieve moderate success with simple designs, the problem quickly becomes challenging for more complex designs. All agents struggle to achieve exact reconstructions within the search budget for construction sequence lengths $\geq$4.

\subsection{Future Work}
Future extensions of this work include \emph{sample efficient} search strategies to ensure successful recovery of construction sequences with fewer interactions with the environment and leveraging constraints present in the dataset to guide CAD program synthesis.
More broadly we envision the dataset can aid the creation of 3D geometry using the same CAD modeling operations as human designers, exploiting the knowledge of domain experts on how shapes are defined and leveraging the strengths of industrial CAD modeling software.  By learning to translate point cloud, image, or mesh data into a sequence of high level modeling operations \cite{tian2018learning, Ellis2018},  watertight CAD models may be synthesized, providing an alternative approach to the reverse engineering problem \cite{Buonamici2017}. 
A remaining challenge is to develop representations that can be conditioned on the design geometry and topology created so far, leveraging the sequential nature of the data for self-attention \cite{Vaswani2017}.
Finally, beyond the simplified design space of sketch and extrude lies the full breadth of rich sequential CAD modeling operations.



\bibliography{main}


\begin{thebibliography}{66}


\ifx \showCODEN    \undefined \def \showCODEN     #1{\unskip}     \fi
\ifx \showDOI      \undefined \def \showDOI       #1{#1}\fi
\ifx \showISBNx    \undefined \def \showISBNx     #1{\unskip}     \fi
\ifx \showISBNxiii \undefined \def \showISBNxiii  #1{\unskip}     \fi
\ifx \showISSN     \undefined \def \showISSN      #1{\unskip}     \fi
\ifx \showLCCN     \undefined \def \showLCCN      #1{\unskip}     \fi
\ifx \shownote     \undefined \def \shownote      #1{#1}          \fi
\ifx \showarticletitle \undefined \def \showarticletitle #1{#1}   \fi
\ifx \showURL      \undefined \def \showURL       {\relax}        \fi
\providecommand\bibfield[2]{#2}
\providecommand\bibinfo[2]{#2}
\providecommand\natexlab[1]{#1}
\providecommand\showeprint[2][]{arXiv:#2}

\bibitem[\protect\citeauthoryear{Ansaldi, De~Floriani, and Falcidieno}{Ansaldi
  et~al\mbox{.}}{1985}]%
        {ansaldi1985geometric}
\bibfield{author}{\bibinfo{person}{Silvia Ansaldi}, \bibinfo{person}{Leila
  De~Floriani}, {and} \bibinfo{person}{Bianca Falcidieno}.}
  \bibinfo{year}{1985}\natexlab{}.
\newblock \showarticletitle{Geometric modeling of solid objects by using a face
  adjacency graph representation}.
\newblock \bibinfo{journal}{\emph{ACM SIGGRAPH Computer Graphics}}
  \bibinfo{volume}{19}, \bibinfo{number}{3} (\bibinfo{year}{1985}),
  \bibinfo{pages}{131--139}.
\newblock


\bibitem[\protect\citeauthoryear{Autodesk}{Autodesk}{2012}]%
        {inventorFR}
\bibfield{author}{\bibinfo{person}{Autodesk}.} \bibinfo{year}{2012}\natexlab{}.
\newblock \bibinfo{booktitle}{\emph{Inventor Feature Recognition}}.
\newblock
\urldef\tempurl%
\url{https://apps.autodesk.com/INVNTOR/en/Detail/Index?id=9172877436288348979}
\showURL{%
\tempurl}


\bibitem[\protect\citeauthoryear{Autodesk}{Autodesk}{2014}]%
        {fusion360API}
\bibfield{author}{\bibinfo{person}{Autodesk}.} \bibinfo{year}{2014}\natexlab{}.
\newblock \bibinfo{booktitle}{\emph{Fusion 360 API}}.
\newblock
\urldef\tempurl%
\url{http://help.autodesk.com/view/fusion360/ENU/?guid=GUID-7B5A90C8-E94C-48DA-B16B-430729B734DC}
\showURL{%
\tempurl}


\bibitem[\protect\citeauthoryear{Autodesk}{Autodesk}{2015}]%
        {autodeskOnlineGallery}
\bibfield{author}{\bibinfo{person}{Autodesk}.} \bibinfo{year}{2015}\natexlab{}.
\newblock \bibinfo{booktitle}{\emph{Autodesk Online Gallery}}.
\newblock
\urldef\tempurl%
\url{https://gallery.autodesk.com}
\showURL{%
\tempurl}


\bibitem[\protect\citeauthoryear{Benko, Kos, Benkő, Andor, Kós, Varady,
  Andor, and Martin}{Benko et~al\mbox{.}}{2002}]%
        {Benko02constrainedfitting}
\bibfield{author}{\bibinfo{person}{Pal Benko}, \bibinfo{person}{Geza Kos},
  \bibinfo{person}{Pál Benkő}, \bibinfo{person}{Laszlo Andor},
  \bibinfo{person}{Géza Kós}, \bibinfo{person}{Tamas Varady},
  \bibinfo{person}{László Andor}, {and} \bibinfo{person}{Ralph Martin}.}
  \bibinfo{year}{2002}\natexlab{}.
\newblock \bibinfo{title}{Constrained Fitting in Reverse Engineering}.
\newblock
\newblock


\bibitem[\protect\citeauthoryear{Buchele}{Buchele}{2000}]%
        {buchele2000three}
\bibfield{author}{\bibinfo{person}{Suzanne~Fox Buchele}.}
  \bibinfo{year}{2000}\natexlab{}.
\newblock \showarticletitle{Three-dimensional binary space partitioning tree
  and constructive solid geometry tree construction from algebraic boundary
  representations.}
\newblock  (\bibinfo{year}{2000}).
\newblock


\bibitem[\protect\citeauthoryear{Buchele and Crawford}{Buchele and
  Crawford}{2003}]%
        {buchele2003three}
\bibfield{author}{\bibinfo{person}{Suzanne~F Buchele} {and}
  \bibinfo{person}{Richard~H Crawford}.} \bibinfo{year}{2003}\natexlab{}.
\newblock \showarticletitle{Three-dimensional halfspace constructive solid
  geometry tree construction from implicit boundary representations}. In
  \bibinfo{booktitle}{\emph{Proceedings of the eighth ACM symposium on Solid
  modeling and applications}}. \bibinfo{pages}{135--144}.
\newblock


\bibitem[\protect\citeauthoryear{Buchele and Roles}{Buchele and Roles}{2001}]%
        {buchele2001binary}
\bibfield{author}{\bibinfo{person}{Suzanne~F Buchele} {and}
  \bibinfo{person}{Angela~C Roles}.} \bibinfo{year}{2001}\natexlab{}.
\newblock \showarticletitle{Binary space partitioning tree and constructive
  solid geometry representations for objects bounded by curved surfaces.}. In
  \bibinfo{booktitle}{\emph{CCCG}}. Citeseer, \bibinfo{pages}{49--52}.
\newblock


\bibitem[\protect\citeauthoryear{Buonamici, Carfagni, Furferi, Governi, Lapini,
  and Volpe}{Buonamici et~al\mbox{.}}{2018}]%
        {Buonamici2017}
\bibfield{author}{\bibinfo{person}{Francesco Buonamici},
  \bibinfo{person}{Monica Carfagni}, \bibinfo{person}{Rocco Furferi},
  \bibinfo{person}{Lapo Governi}, \bibinfo{person}{Alessandro Lapini}, {and}
  \bibinfo{person}{Yary Volpe}.} \bibinfo{year}{2018}\natexlab{}.
\newblock \showarticletitle{Reverse engineering modeling methods and tools: a
  survey}.
\newblock \bibinfo{journal}{\emph{Computer-Aided Design and Applications}}
  \bibinfo{volume}{15}, \bibinfo{number}{3} (\bibinfo{year}{2018}),
  \bibinfo{pages}{443--464}.
\newblock
\urldef\tempurl%
\url{https://doi.org/10.1080/16864360.2017.1397894}
\showDOI{\tempurl}
\showeprint{https://doi.org/10.1080/16864360.2017.1397894}


\bibitem[\protect\citeauthoryear{Chang, Funkhouser, Guibas, Hanrahan, Huang,
  Li, Savarese, Savva, Song, Su, et~al\mbox{.}}{Chang et~al\mbox{.}}{2015}]%
        {chang2015shapenet}
\bibfield{author}{\bibinfo{person}{Angel~X Chang}, \bibinfo{person}{Thomas
  Funkhouser}, \bibinfo{person}{Leonidas Guibas}, \bibinfo{person}{Pat
  Hanrahan}, \bibinfo{person}{Qixing Huang}, \bibinfo{person}{Zimo Li},
  \bibinfo{person}{Silvio Savarese}, \bibinfo{person}{Manolis Savva},
  \bibinfo{person}{Shuran Song}, \bibinfo{person}{Hao Su}, {et~al\mbox{.}}}
  \bibinfo{year}{2015}\natexlab{}.
\newblock \showarticletitle{Shapenet: An information-rich 3d model repository}.
\newblock \bibinfo{journal}{\emph{arXiv preprint arXiv:1512.03012}}
  (\bibinfo{year}{2015}).
\newblock


\bibitem[\protect\citeauthoryear{Chen, Tagliasacchi, and Zhang}{Chen
  et~al\mbox{.}}{2020}]%
        {chen2020bsp}
\bibfield{author}{\bibinfo{person}{Zhiqin Chen}, \bibinfo{person}{Andrea
  Tagliasacchi}, {and} \bibinfo{person}{Hao Zhang}.}
  \bibinfo{year}{2020}\natexlab{}.
\newblock \showarticletitle{Bsp-net: Generating compact meshes via binary space
  partitioning}. In \bibinfo{booktitle}{\emph{Proceedings of the IEEE/CVF
  Conference on Computer Vision and Pattern Recognition}}.
  \bibinfo{pages}{45--54}.
\newblock


\bibitem[\protect\citeauthoryear{Dassault}{Dassault}{2019}]%
        {featureWorks}
\bibfield{author}{\bibinfo{person}{Dassault}.} \bibinfo{year}{2019}\natexlab{}.
\newblock \bibinfo{booktitle}{\emph{Solidworks FeatureWorks}}.
\newblock
\urldef\tempurl%
\url{https://help.solidworks.com/2019/english/SolidWorks/fworks/c_Overview_of_FeatureWorks.htm}
\showURL{%
\tempurl}


\bibitem[\protect\citeauthoryear{Deng, Dong, Socher, Li, Li, and Fei-Fei}{Deng
  et~al\mbox{.}}{2009}]%
        {deng2009imagenet}
\bibfield{author}{\bibinfo{person}{Jia Deng}, \bibinfo{person}{Wei Dong},
  \bibinfo{person}{Richard Socher}, \bibinfo{person}{Li-Jia Li},
  \bibinfo{person}{Kai Li}, {and} \bibinfo{person}{Li Fei-Fei}.}
  \bibinfo{year}{2009}\natexlab{}.
\newblock \showarticletitle{Imagenet: A large-scale hierarchical image
  database}. In \bibinfo{booktitle}{\emph{2009 IEEE conference on computer
  vision and pattern recognition}}. Ieee, \bibinfo{pages}{248--255}.
\newblock


\bibitem[\protect\citeauthoryear{Devlin, Uesato, Bhupatiraju, Singh, Mohamed,
  and Kohli}{Devlin et~al\mbox{.}}{2017}]%
        {devlin2017robustfill}
\bibfield{author}{\bibinfo{person}{Jacob Devlin}, \bibinfo{person}{Jonathan
  Uesato}, \bibinfo{person}{Surya Bhupatiraju}, \bibinfo{person}{Rishabh
  Singh}, \bibinfo{person}{Abdel-rahman Mohamed}, {and}
  \bibinfo{person}{Pushmeet Kohli}.} \bibinfo{year}{2017}\natexlab{}.
\newblock \showarticletitle{Robustfill: Neural program learning under noisy
  i/o}.
\newblock \bibinfo{journal}{\emph{arXiv preprint arXiv:1703.07469}}
  (\bibinfo{year}{2017}).
\newblock


\bibitem[\protect\citeauthoryear{Du, Inala, Pu, Spielberg, Schulz, Rus,
  Solar-Lezama, and Matusik}{Du et~al\mbox{.}}{2018}]%
        {du2018inversecsg}
\bibfield{author}{\bibinfo{person}{Tao Du}, \bibinfo{person}{Jeevana~Priya
  Inala}, \bibinfo{person}{Yewen Pu}, \bibinfo{person}{Andrew Spielberg},
  \bibinfo{person}{Adriana Schulz}, \bibinfo{person}{Daniela Rus},
  \bibinfo{person}{Armando Solar-Lezama}, {and} \bibinfo{person}{Wojciech
  Matusik}.} \bibinfo{year}{2018}\natexlab{}.
\newblock \showarticletitle{Inversecsg: Automatic conversion of 3d models to
  csg trees}.
\newblock \bibinfo{journal}{\emph{ACM Transactions on Graphics (TOG)}}
  \bibinfo{volume}{37}, \bibinfo{number}{6} (\bibinfo{year}{2018}),
  \bibinfo{pages}{1--16}.
\newblock


\bibitem[\protect\citeauthoryear{Eitz, Hays, and Alexa}{Eitz
  et~al\mbox{.}}{2012}]%
        {eitz2012humans}
\bibfield{author}{\bibinfo{person}{Mathias Eitz}, \bibinfo{person}{James Hays},
  {and} \bibinfo{person}{Marc Alexa}.} \bibinfo{year}{2012}\natexlab{}.
\newblock \showarticletitle{How do humans sketch objects?}
\newblock \bibinfo{journal}{\emph{ACM Transactions on graphics (TOG)}}
  \bibinfo{volume}{31}, \bibinfo{number}{4} (\bibinfo{year}{2012}),
  \bibinfo{pages}{1--10}.
\newblock


\bibitem[\protect\citeauthoryear{Ellis, Nye, Pu, Sosa, Tenenbaum, and
  Solar-Lezama}{Ellis et~al\mbox{.}}{2019}]%
        {ellis2019write}
\bibfield{author}{\bibinfo{person}{Kevin Ellis}, \bibinfo{person}{Maxwell Nye},
  \bibinfo{person}{Yewen Pu}, \bibinfo{person}{Felix Sosa},
  \bibinfo{person}{Josh Tenenbaum}, {and} \bibinfo{person}{Armando
  Solar-Lezama}.} \bibinfo{year}{2019}\natexlab{}.
\newblock \showarticletitle{Write, execute, assess: Program synthesis with a
  repl}. In \bibinfo{booktitle}{\emph{Advances in Neural Information Processing
  Systems}}. \bibinfo{pages}{9169--9178}.
\newblock


\bibitem[\protect\citeauthoryear{Ellis, Ritchie, Solar-Lezama, and
  Tenenbaum}{Ellis et~al\mbox{.}}{2018}]%
        {Ellis2018}
\bibfield{author}{\bibinfo{person}{Kevin Ellis}, \bibinfo{person}{Daniel
  Ritchie}, \bibinfo{person}{Armando Solar-Lezama}, {and} \bibinfo{person}{Josh
  Tenenbaum}.} \bibinfo{year}{2018}\natexlab{}.
\newblock \showarticletitle{Learning to Infer Graphics Programs from Hand-Drawn
  Images}. In \bibinfo{booktitle}{\emph{Advances in Neural Information
  Processing Systems}}, \bibfield{editor}{\bibinfo{person}{S.~Bengio},
  \bibinfo{person}{H.~Wallach}, \bibinfo{person}{H.~Larochelle},
  \bibinfo{person}{K.~Grauman}, \bibinfo{person}{N.~Cesa-Bianchi}, {and}
  \bibinfo{person}{R.~Garnett}} (Eds.), Vol.~\bibinfo{volume}{31}.
  \bibinfo{publisher}{Curran Associates, Inc.}, \bibinfo{pages}{6059--6068}.
\newblock
\urldef\tempurl%
\url{https://proceedings.neurips.cc/paper/2018/file/6788076842014c83cedadbe6b0ba0314-Paper.pdf}
\showURL{%
\tempurl}


\bibitem[\protect\citeauthoryear{Fayolle and Pasko}{Fayolle and Pasko}{2016}]%
        {fayolle2016evolutionary}
\bibfield{author}{\bibinfo{person}{Pierre-Alain Fayolle} {and}
  \bibinfo{person}{Alexander Pasko}.} \bibinfo{year}{2016}\natexlab{}.
\newblock \showarticletitle{An evolutionary approach to the extraction of
  object construction trees from 3D point clouds}.
\newblock \bibinfo{journal}{\emph{Computer-Aided Design}}  \bibinfo{volume}{74}
  (\bibinfo{year}{2016}), \bibinfo{pages}{1--17}.
\newblock


\bibitem[\protect\citeauthoryear{Friedrich, Fayolle, Gabor, and
  Linnhoff-Popien}{Friedrich et~al\mbox{.}}{2019}]%
        {friedrich2019optimizing}
\bibfield{author}{\bibinfo{person}{Markus Friedrich},
  \bibinfo{person}{Pierre-Alain Fayolle}, \bibinfo{person}{Thomas Gabor}, {and}
  \bibinfo{person}{Claudia Linnhoff-Popien}.} \bibinfo{year}{2019}\natexlab{}.
\newblock \showarticletitle{Optimizing evolutionary CSG tree extraction}. In
  \bibinfo{booktitle}{\emph{Proceedings of the Genetic and Evolutionary
  Computation Conference}}. \bibinfo{pages}{1183--1191}.
\newblock


\bibitem[\protect\citeauthoryear{Gao, Yang, Wu, Yuan, Fu, Lai, and Zhang}{Gao
  et~al\mbox{.}}{2019}]%
        {gao2019sdm}
\bibfield{author}{\bibinfo{person}{Lin Gao}, \bibinfo{person}{Jie Yang},
  \bibinfo{person}{Tong Wu}, \bibinfo{person}{Yu-Jie Yuan},
  \bibinfo{person}{Hongbo Fu}, \bibinfo{person}{Yu-Kun Lai}, {and}
  \bibinfo{person}{Hao Zhang}.} \bibinfo{year}{2019}\natexlab{}.
\newblock \showarticletitle{SDM-NET: Deep generative network for structured
  deformable mesh}.
\newblock \bibinfo{journal}{\emph{ACM Transactions on Graphics (TOG)}}
  \bibinfo{volume}{38}, \bibinfo{number}{6} (\bibinfo{year}{2019}),
  \bibinfo{pages}{1--15}.
\newblock


\bibitem[\protect\citeauthoryear{Gilmer, Schoenholz, Riley, Vinyals, and
  Dahl}{Gilmer et~al\mbox{.}}{2017}]%
        {gilmer2017neural}
\bibfield{author}{\bibinfo{person}{Justin Gilmer}, \bibinfo{person}{Samuel~S
  Schoenholz}, \bibinfo{person}{Patrick~F Riley}, \bibinfo{person}{Oriol
  Vinyals}, {and} \bibinfo{person}{George~E Dahl}.}
  \bibinfo{year}{2017}\natexlab{}.
\newblock \showarticletitle{Neural message passing for quantum chemistry}. In
  \bibinfo{booktitle}{\emph{International Conference on Machine Learning}}.
  PMLR, \bibinfo{pages}{1263--1272}.
\newblock


\bibitem[\protect\citeauthoryear{Gryaditskaya, Sypesteyn, Hoftijzer, Pont,
  Durand, and Bousseau}{Gryaditskaya et~al\mbox{.}}{2019}]%
        {gryaditskaya2019opensketch}
\bibfield{author}{\bibinfo{person}{Yulia Gryaditskaya}, \bibinfo{person}{Mark
  Sypesteyn}, \bibinfo{person}{Jan~Willem Hoftijzer}, \bibinfo{person}{Sylvia
  Pont}, \bibinfo{person}{Fredo Durand}, {and} \bibinfo{person}{Adrien
  Bousseau}.} \bibinfo{year}{2019}\natexlab{}.
\newblock \showarticletitle{Opensketch: A richly-annotated dataset of product
  design sketches}.
\newblock \bibinfo{journal}{\emph{ACM Transactions on Graphics (TOG)}}
  \bibinfo{volume}{38}, \bibinfo{number}{6} (\bibinfo{year}{2019}),
  \bibinfo{pages}{232}.
\newblock


\bibitem[\protect\citeauthoryear{Hamza and Saitou}{Hamza and Saitou}{2004}]%
        {hamza2004optimization}
\bibfield{author}{\bibinfo{person}{Karim Hamza} {and} \bibinfo{person}{Kazuhiro
  Saitou}.} \bibinfo{year}{2004}\natexlab{}.
\newblock \showarticletitle{Optimization of constructive solid geometry via a
  tree-based multi-objective genetic algorithm}. In
  \bibinfo{booktitle}{\emph{Genetic and Evolutionary Computation Conference}}.
  Springer, \bibinfo{pages}{981--992}.
\newblock


\bibitem[\protect\citeauthoryear{Jayaraman, Sanghi, Lambourne, Davies, Shayani,
  and Morris}{Jayaraman et~al\mbox{.}}{2020}]%
        {jayaraman2020uvnet}
\bibfield{author}{\bibinfo{person}{Pradeep~Kumar Jayaraman},
  \bibinfo{person}{Aditya Sanghi}, \bibinfo{person}{Joseph Lambourne},
  \bibinfo{person}{Thomas Davies}, \bibinfo{person}{Hooman Shayani}, {and}
  \bibinfo{person}{Nigel Morris}.} \bibinfo{year}{2020}\natexlab{}.
\newblock \showarticletitle{UV-Net: Learning from Curve-Networks and Solids}.
\newblock \bibinfo{journal}{\emph{arXiv preprint arXiv:2006.10211}}
  (\bibinfo{year}{2020}).
\newblock


\bibitem[\protect\citeauthoryear{Jones, Barton, Xu, Wang, Jiang, Guerrero,
  Mitra, and Ritchie}{Jones et~al\mbox{.}}{2020}]%
        {jones2020shapeAssembly}
\bibfield{author}{\bibinfo{person}{R.~Kenny Jones}, \bibinfo{person}{Theresa
  Barton}, \bibinfo{person}{Xianghao Xu}, \bibinfo{person}{Kai Wang},
  \bibinfo{person}{Ellen Jiang}, \bibinfo{person}{Paul Guerrero},
  \bibinfo{person}{Niloy Mitra}, {and} \bibinfo{person}{Daniel Ritchie}.}
  \bibinfo{year}{2020}\natexlab{}.
\newblock \showarticletitle{ShapeAssembly: Learning to Generate Programs for 3D
  Shape Structure Synthesis}.
\newblock \bibinfo{journal}{\emph{ACM Transactions on Graphics (TOG), Siggraph
  Asia 2020}} \bibinfo{volume}{39}, \bibinfo{number}{6} (\bibinfo{year}{2020}),
  \bibinfo{pages}{Article 234}.
\newblock


\bibitem[\protect\citeauthoryear{Kalyan, Mohta, Polozov, Batra, Jain, and
  Gulwani}{Kalyan et~al\mbox{.}}{2018}]%
        {kalyan2018neural}
\bibfield{author}{\bibinfo{person}{Ashwin Kalyan}, \bibinfo{person}{Abhishek
  Mohta}, \bibinfo{person}{Oleksandr Polozov}, \bibinfo{person}{Dhruv Batra},
  \bibinfo{person}{Prateek Jain}, {and} \bibinfo{person}{Sumit Gulwani}.}
  \bibinfo{year}{2018}\natexlab{}.
\newblock \showarticletitle{Neural-guided deductive search for real-time
  program synthesis from examples}.
\newblock \bibinfo{journal}{\emph{arXiv preprint arXiv:1804.01186}}
  (\bibinfo{year}{2018}).
\newblock


\bibitem[\protect\citeauthoryear{Kania, Zi{\k{e}}ba, and Kajdanowicz}{Kania
  et~al\mbox{.}}{2020}]%
        {kania2020ucsg}
\bibfield{author}{\bibinfo{person}{Kacper Kania}, \bibinfo{person}{Maciej
  Zi{\k{e}}ba}, {and} \bibinfo{person}{Tomasz Kajdanowicz}.}
  \bibinfo{year}{2020}\natexlab{}.
\newblock \showarticletitle{UCSG-Net--Unsupervised Discovering of Constructive
  Solid Geometry Tree}.
\newblock \bibinfo{journal}{\emph{arXiv preprint arXiv:2006.09102}}
  (\bibinfo{year}{2020}).
\newblock


\bibitem[\protect\citeauthoryear{Kim, Chi, Hu, Huang, and Ramani}{Kim
  et~al\mbox{.}}{2020}]%
        {sangpil2020large}
\bibfield{author}{\bibinfo{person}{Sangpil Kim}, \bibinfo{person}{Hyung-gun
  Chi}, \bibinfo{person}{Xiao Hu}, \bibinfo{person}{Qixing Huang}, {and}
  \bibinfo{person}{Karthik Ramani}.} \bibinfo{year}{2020}\natexlab{}.
\newblock \showarticletitle{A Large-scale Annotated Mechanical Components
  Benchmark for Classification and Retrieval Tasks with Deep Neural Networks}.
  In \bibinfo{booktitle}{\emph{Proceedings of 16th European Conference on
  Computer Vision (ECCV)}}.
\newblock


\bibitem[\protect\citeauthoryear{Kipf, Fetaya, Wang, Welling, and Zemel}{Kipf
  et~al\mbox{.}}{2018}]%
        {kipf2018neural}
\bibfield{author}{\bibinfo{person}{Thomas Kipf}, \bibinfo{person}{Ethan
  Fetaya}, \bibinfo{person}{Kuan-Chieh Wang}, \bibinfo{person}{Max Welling},
  {and} \bibinfo{person}{Richard Zemel}.} \bibinfo{year}{2018}\natexlab{}.
\newblock \showarticletitle{Neural relational inference for interacting
  systems}. In \bibinfo{booktitle}{\emph{International Conference on Machine
  Learning}}. PMLR, \bibinfo{pages}{2688--2697}.
\newblock


\bibitem[\protect\citeauthoryear{Kipf and Welling}{Kipf and Welling}{2016}]%
        {kipf2016semi}
\bibfield{author}{\bibinfo{person}{Thomas~N Kipf} {and} \bibinfo{person}{Max
  Welling}.} \bibinfo{year}{2016}\natexlab{}.
\newblock \showarticletitle{Semi-supervised classification with graph
  convolutional networks}.
\newblock \bibinfo{journal}{\emph{arXiv preprint arXiv:1609.02907}}
  (\bibinfo{year}{2016}).
\newblock


\bibitem[\protect\citeauthoryear{Koch, Matveev, Jiang, Williams, Artemov,
  Burnaev, Alexa, Zorin, and Panozzo}{Koch et~al\mbox{.}}{2019}]%
        {koch2019abc}
\bibfield{author}{\bibinfo{person}{Sebastian Koch}, \bibinfo{person}{Albert
  Matveev}, \bibinfo{person}{Zhongshi Jiang}, \bibinfo{person}{Francis
  Williams}, \bibinfo{person}{Alexey Artemov}, \bibinfo{person}{Evgeny
  Burnaev}, \bibinfo{person}{Marc Alexa}, \bibinfo{person}{Denis Zorin}, {and}
  \bibinfo{person}{Daniele Panozzo}.} \bibinfo{year}{2019}\natexlab{}.
\newblock \showarticletitle{ABC: A big CAD model dataset for geometric deep
  learning}. In \bibinfo{booktitle}{\emph{Proceedings of the IEEE Conference on
  Computer Vision and Pattern Recognition}}. \bibinfo{pages}{9601--9611}.
\newblock


\bibitem[\protect\citeauthoryear{Li, Pan, Bousseau, and Mitra}{Li
  et~al\mbox{.}}{2020b}]%
        {li2020sketch2CAD}
\bibfield{author}{\bibinfo{person}{Changjian Li}, \bibinfo{person}{Hao Pan},
  \bibinfo{person}{Adrien Bousseau}, {and} \bibinfo{person}{Niloy~J. Mitra}.}
  \bibinfo{year}{2020}\natexlab{b}.
\newblock \showarticletitle{Sketch2CAD: Sequential CAD Modeling by Sketching in
  Context}.
\newblock \bibinfo{journal}{\emph{ACM Trans. Graph. (Proceedings of SIGGRAPH
  Asia 2020)}} \bibinfo{volume}{39}, \bibinfo{number}{6}
  (\bibinfo{year}{2020}), \bibinfo{pages}{164:1--164:14}.
\newblock
\urldef\tempurl%
\url{https://doi.org/10.1145/3414685.3417807}
\showDOI{\tempurl}


\bibitem[\protect\citeauthoryear{Li, Niu, and Xu}{Li et~al\mbox{.}}{2020a}]%
        {li2020learning}
\bibfield{author}{\bibinfo{person}{Jun Li}, \bibinfo{person}{Chengjie Niu},
  {and} \bibinfo{person}{Kai Xu}.} \bibinfo{year}{2020}\natexlab{a}.
\newblock \showarticletitle{Learning part generation and assembly for
  structure-aware shape synthesis}. In \bibinfo{booktitle}{\emph{Proceedings of
  the AAAI Conference on Artificial Intelligence}}, Vol.~\bibinfo{volume}{34}.
  \bibinfo{pages}{11362--11369}.
\newblock


\bibitem[\protect\citeauthoryear{Liao, Li, Song, Wang, Hamilton, Duvenaud,
  Urtasun, and Zemel}{Liao et~al\mbox{.}}{2019}]%
        {liao2019efficient}
\bibfield{author}{\bibinfo{person}{Renjie Liao}, \bibinfo{person}{Yujia Li},
  \bibinfo{person}{Yang Song}, \bibinfo{person}{Shenlong Wang},
  \bibinfo{person}{Will Hamilton}, \bibinfo{person}{David~K Duvenaud},
  \bibinfo{person}{Raquel Urtasun}, {and} \bibinfo{person}{Richard Zemel}.}
  \bibinfo{year}{2019}\natexlab{}.
\newblock \showarticletitle{Efficient graph generation with graph recurrent
  attention networks}. In \bibinfo{booktitle}{\emph{Advances in Neural
  Information Processing Systems}}. \bibinfo{pages}{4255--4265}.
\newblock


\bibitem[\protect\citeauthoryear{Lin, Fan, Wang, and Nie{\ss}ner}{Lin
  et~al\mbox{.}}{2020}]%
        {lin2020modeling}
\bibfield{author}{\bibinfo{person}{Cheng Lin}, \bibinfo{person}{Tingxiang Fan},
  \bibinfo{person}{Wenping Wang}, {and} \bibinfo{person}{Matthias
  Nie{\ss}ner}.} \bibinfo{year}{2020}\natexlab{}.
\newblock \showarticletitle{Modeling 3D Shapes by Reinforcement Learning}.
\newblock \bibinfo{journal}{\emph{ECCV}} (\bibinfo{year}{2020}).
\newblock


\bibitem[\protect\citeauthoryear{Mo, Guerrero, Yi, Su, Wonka, Mitra, and
  Guibas}{Mo et~al\mbox{.}}{2019a}]%
        {mo2019structurenet}
\bibfield{author}{\bibinfo{person}{Kaichun Mo}, \bibinfo{person}{Paul
  Guerrero}, \bibinfo{person}{Li Yi}, \bibinfo{person}{Hao Su},
  \bibinfo{person}{Peter Wonka}, \bibinfo{person}{Niloy~J. Mitra}, {and}
  \bibinfo{person}{Leonidas~J. Guibas}.} \bibinfo{year}{2019}\natexlab{a}.
\newblock \showarticletitle{StructureNet: Hierarchical Graph Networks for 3D
  Shape Generation}.
\newblock \bibinfo{journal}{\emph{ACM Trans. Graph.}} \bibinfo{volume}{38},
  \bibinfo{number}{6}, Article \bibinfo{articleno}{242} (\bibinfo{date}{Nov.}
  \bibinfo{year}{2019}), \bibinfo{numpages}{19}~pages.
\newblock
\showISSN{0730-0301}
\urldef\tempurl%
\url{https://doi.org/10.1145/3355089.3356527}
\showDOI{\tempurl}


\bibitem[\protect\citeauthoryear{Mo, Zhu, Chang, Yi, Tripathi, Guibas, and
  Su}{Mo et~al\mbox{.}}{2019b}]%
        {mo2019partnet}
\bibfield{author}{\bibinfo{person}{Kaichun Mo}, \bibinfo{person}{Shilin Zhu},
  \bibinfo{person}{Angel~X Chang}, \bibinfo{person}{Li Yi},
  \bibinfo{person}{Subarna Tripathi}, \bibinfo{person}{Leonidas~J Guibas},
  {and} \bibinfo{person}{Hao Su}.} \bibinfo{year}{2019}\natexlab{b}.
\newblock \showarticletitle{Partnet: A large-scale benchmark for fine-grained
  and hierarchical part-level 3d object understanding}. In
  \bibinfo{booktitle}{\emph{Proceedings of the IEEE Conference on Computer
  Vision and Pattern Recognition}}. \bibinfo{pages}{909--918}.
\newblock


\bibitem[\protect\citeauthoryear{Nandi, Caspi, Grossman, and Tatlock}{Nandi
  et~al\mbox{.}}{2017}]%
        {nandi2017programming}
\bibfield{author}{\bibinfo{person}{Chandrakana Nandi}, \bibinfo{person}{Anat
  Caspi}, \bibinfo{person}{Dan Grossman}, {and} \bibinfo{person}{Zachary
  Tatlock}.} \bibinfo{year}{2017}\natexlab{}.
\newblock \showarticletitle{Programming language tools and techniques for 3D
  printing}. In \bibinfo{booktitle}{\emph{2nd Summit on Advances in Programming
  Languages (SNAPL 2017)}}. Schloss Dagstuhl-Leibniz-Zentrum fuer Informatik.
\newblock


\bibitem[\protect\citeauthoryear{Nandi, Wilcox, Panchekha, Blau, Grossman, and
  Tatlock}{Nandi et~al\mbox{.}}{2018}]%
        {nandi2018functional}
\bibfield{author}{\bibinfo{person}{Chandrakana Nandi}, \bibinfo{person}{James~R
  Wilcox}, \bibinfo{person}{Pavel Panchekha}, \bibinfo{person}{Taylor Blau},
  \bibinfo{person}{Dan Grossman}, {and} \bibinfo{person}{Zachary Tatlock}.}
  \bibinfo{year}{2018}\natexlab{}.
\newblock \showarticletitle{Functional programming for compiling and
  decompiling computer-aided design}.
\newblock \bibinfo{journal}{\emph{Proceedings of the ACM on Programming
  Languages}} \bibinfo{volume}{2}, \bibinfo{number}{ICFP}
  (\bibinfo{year}{2018}), \bibinfo{pages}{1--31}.
\newblock


\bibitem[\protect\citeauthoryear{Nandi, Willsey, Anderson, Wilcox, Darulova,
  Grossman, and Tatlock}{Nandi et~al\mbox{.}}{2020}]%
        {nandi2020synthesizing}
\bibfield{author}{\bibinfo{person}{Chandrakana Nandi}, \bibinfo{person}{Max
  Willsey}, \bibinfo{person}{Adam Anderson}, \bibinfo{person}{James~R. Wilcox},
  \bibinfo{person}{Eva Darulova}, \bibinfo{person}{Dan Grossman}, {and}
  \bibinfo{person}{Zachary Tatlock}.} \bibinfo{year}{2020}\natexlab{}.
\newblock \showarticletitle{Synthesizing Structured CAD Models with Equality
  Saturation and Inverse Transformations}. In
  \bibinfo{booktitle}{\emph{Proceedings of the 41st ACM SIGPLAN Conference on
  Programming Language Design and Implementation}}. \bibinfo{pages}{31–44}.
\newblock


\bibitem[\protect\citeauthoryear{Nash, Ganin, Eslami, and Battaglia}{Nash
  et~al\mbox{.}}{2020}]%
        {nash2020polygen}
\bibfield{author}{\bibinfo{person}{Charlie Nash}, \bibinfo{person}{Yaroslav
  Ganin}, \bibinfo{person}{SM~Ali Eslami}, {and} \bibinfo{person}{Peter
  Battaglia}.} \bibinfo{year}{2020}\natexlab{}.
\newblock \showarticletitle{Polygen: An autoregressive generative model of 3d
  meshes}. In \bibinfo{booktitle}{\emph{International Conference on Machine
  Learning}}. PMLR, \bibinfo{pages}{7220--7229}.
\newblock


\bibitem[\protect\citeauthoryear{Sangkloy, Burnell, Ham, and Hays}{Sangkloy
  et~al\mbox{.}}{2016}]%
        {sangkloy2016sketchy}
\bibfield{author}{\bibinfo{person}{Patsorn Sangkloy}, \bibinfo{person}{Nathan
  Burnell}, \bibinfo{person}{Cusuh Ham}, {and} \bibinfo{person}{James Hays}.}
  \bibinfo{year}{2016}\natexlab{}.
\newblock \showarticletitle{The sketchy database: learning to retrieve badly
  drawn bunnies}.
\newblock \bibinfo{journal}{\emph{ACM Transactions on Graphics (TOG)}}
  \bibinfo{volume}{35}, \bibinfo{number}{4} (\bibinfo{year}{2016}),
  \bibinfo{pages}{1--12}.
\newblock


\bibitem[\protect\citeauthoryear{Schor, Katzir, Zhang, and Cohen-Or}{Schor
  et~al\mbox{.}}{2019}]%
        {schor2019componet}
\bibfield{author}{\bibinfo{person}{Nadav Schor}, \bibinfo{person}{Oren Katzir},
  \bibinfo{person}{Hao Zhang}, {and} \bibinfo{person}{Daniel Cohen-Or}.}
  \bibinfo{year}{2019}\natexlab{}.
\newblock \showarticletitle{CompoNet: Learning to generate the unseen by part
  synthesis and composition}. In \bibinfo{booktitle}{\emph{Proceedings of the
  IEEE/CVF International Conference on Computer Vision}}.
  \bibinfo{pages}{8759--8768}.
\newblock


\bibitem[\protect\citeauthoryear{Schulz, Shamir, Levin, Sitthi-Amorn, and
  Matusik}{Schulz et~al\mbox{.}}{2014}]%
        {schulz2014FabByExample}
\bibfield{author}{\bibinfo{person}{Adriana Schulz}, \bibinfo{person}{Ariel
  Shamir}, \bibinfo{person}{David I.~W. Levin}, \bibinfo{person}{Pitchaya
  Sitthi-Amorn}, {and} \bibinfo{person}{Wojciech Matusik}.}
  \bibinfo{year}{2014}\natexlab{}.
\newblock \showarticletitle{Design and Fabrication by Example}.
\newblock \bibinfo{journal}{\emph{ACM Transactions on Graphics (Proceedings
  SIGGRAPH 2014)}} \bibinfo{volume}{33}, \bibinfo{number}{4}
  (\bibinfo{year}{2014}).
\newblock


\bibitem[\protect\citeauthoryear{Seff, Ovadia, Zhou, and Adams}{Seff
  et~al\mbox{.}}{2020}]%
        {SketchGraphs}
\bibfield{author}{\bibinfo{person}{Ari Seff}, \bibinfo{person}{Yaniv Ovadia},
  \bibinfo{person}{Wenda Zhou}, {and} \bibinfo{person}{Ryan~P. Adams}.}
  \bibinfo{year}{2020}\natexlab{}.
\newblock \showarticletitle{SketchGraphs: A Large-Scale Dataset for Modeling
  Relational Geometry in Computer-Aided Design}.
\newblock In \bibinfo{booktitle}{\emph{ICML 2020 Workshop on Object-Oriented
  Learning}}.
\newblock


\bibitem[\protect\citeauthoryear{Shah, Anderson, Kim, and Joshi}{Shah
  et~al\mbox{.}}{2001}]%
        {shah2001discourse}
\bibfield{author}{\bibinfo{person}{Jami~J Shah}, \bibinfo{person}{David
  Anderson}, \bibinfo{person}{Yong~Se Kim}, {and} \bibinfo{person}{Sanjay
  Joshi}.} \bibinfo{year}{2001}\natexlab{}.
\newblock \showarticletitle{A discourse on geometric feature recognition from
  CAD models}.
\newblock \bibinfo{journal}{\emph{J. Comput. Inf. Sci. Eng.}}
  \bibinfo{volume}{1}, \bibinfo{number}{1} (\bibinfo{year}{2001}),
  \bibinfo{pages}{41--51}.
\newblock


\bibitem[\protect\citeauthoryear{Shapiro and Vossler}{Shapiro and
  Vossler}{1993}]%
        {shapiro1993separation}
\bibfield{author}{\bibinfo{person}{Vadim Shapiro} {and}
  \bibinfo{person}{Donald~L Vossler}.} \bibinfo{year}{1993}\natexlab{}.
\newblock \showarticletitle{Separation for boundary to CSG conversion}.
\newblock \bibinfo{journal}{\emph{ACM Transactions on Graphics (TOG)}}
  \bibinfo{volume}{12}, \bibinfo{number}{1} (\bibinfo{year}{1993}),
  \bibinfo{pages}{35--55}.
\newblock


\bibitem[\protect\citeauthoryear{Sharma, Goyal, Liu, Kalogerakis, and
  Maji}{Sharma et~al\mbox{.}}{2017}]%
        {sharma2017csgnet}
\bibfield{author}{\bibinfo{person}{Gopal Sharma}, \bibinfo{person}{Rishabh
  Goyal}, \bibinfo{person}{Difan Liu}, \bibinfo{person}{Evangelos Kalogerakis},
  {and} \bibinfo{person}{Subhransu Maji}.} \bibinfo{year}{2017}\natexlab{}.
\newblock \showarticletitle{CSGNet: Neural Shape Parser for Constructive Solid
  Geometry. CoRR abs/1712.08290 (2017)}.
\newblock \bibinfo{journal}{\emph{arXiv preprint arXiv:1712.08290}}
  (\bibinfo{year}{2017}).
\newblock


\bibitem[\protect\citeauthoryear{Starly}{Starly}{2020}]%
        {fabwave}
\bibfield{author}{\bibinfo{person}{Binil Starly}.}
  \bibinfo{year}{2020}\natexlab{}.
\newblock \bibinfo{booktitle}{\emph{FabWave - 3D Part Repository}}.
\newblock
\urldef\tempurl%
\url{https://www.dimelab.org/fabwave}
\showURL{%
\tempurl}


\bibitem[\protect\citeauthoryear{Stava, Pirk, Kratt, Chen, Mundefinedch,
  Deussen, and Benes}{Stava et~al\mbox{.}}{2014}]%
        {stava2014trees}
\bibfield{author}{\bibinfo{person}{O. Stava}, \bibinfo{person}{S. Pirk},
  \bibinfo{person}{J. Kratt}, \bibinfo{person}{B. Chen}, \bibinfo{person}{R.
  Mundefinedch}, \bibinfo{person}{O. Deussen}, {and} \bibinfo{person}{B.
  Benes}.} \bibinfo{year}{2014}\natexlab{}.
\newblock \showarticletitle{Inverse Procedural Modelling of Trees}.
\newblock \bibinfo{journal}{\emph{Comput. Graph. Forum}} \bibinfo{volume}{33},
  \bibinfo{number}{6} (\bibinfo{date}{Sept.} \bibinfo{year}{2014}),
  \bibinfo{pages}{118–131}.
\newblock
\showISSN{0167-7055}
\urldef\tempurl%
\url{https://doi.org/10.1111/cgf.12282}
\showDOI{\tempurl}


\bibitem[\protect\citeauthoryear{Sung, Su, Kim, Chaudhuri, and Guibas}{Sung
  et~al\mbox{.}}{2017}]%
        {sung2017complementme}
\bibfield{author}{\bibinfo{person}{Minhyuk Sung}, \bibinfo{person}{Hao Su},
  \bibinfo{person}{Vladimir~G Kim}, \bibinfo{person}{Siddhartha Chaudhuri},
  {and} \bibinfo{person}{Leonidas Guibas}.} \bibinfo{year}{2017}\natexlab{}.
\newblock \showarticletitle{ComplementMe: Weakly-supervised component
  suggestions for 3D modeling}.
\newblock \bibinfo{journal}{\emph{ACM Transactions on Graphics (TOG)}}
  \bibinfo{volume}{36}, \bibinfo{number}{6} (\bibinfo{year}{2017}),
  \bibinfo{pages}{1--12}.
\newblock


\bibitem[\protect\citeauthoryear{Talton, Lou, Lesser, Duke, M\v{e}ch, and
  Koltun}{Talton et~al\mbox{.}}{2011}]%
        {talton2011metropolis}
\bibfield{author}{\bibinfo{person}{Jerry~O. Talton}, \bibinfo{person}{Yu Lou},
  \bibinfo{person}{Steve Lesser}, \bibinfo{person}{Jared Duke},
  \bibinfo{person}{Radom\'{\i}r M\v{e}ch}, {and} \bibinfo{person}{Vladlen
  Koltun}.} \bibinfo{year}{2011}\natexlab{}.
\newblock \showarticletitle{Metropolis Procedural Modeling}.
\newblock \bibinfo{journal}{\emph{ACM Trans. Graph.}} \bibinfo{volume}{30},
  \bibinfo{number}{2}, Article \bibinfo{articleno}{11} (\bibinfo{date}{April}
  \bibinfo{year}{2011}), \bibinfo{numpages}{14}~pages.
\newblock
\showISSN{0730-0301}
\urldef\tempurl%
\url{https://doi.org/10.1145/1944846.1944851}
\showDOI{\tempurl}


\bibitem[\protect\citeauthoryear{Tang, Agrawal, and Faenza}{Tang
  et~al\mbox{.}}{2019}]%
        {tang2019reinforcement}
\bibfield{author}{\bibinfo{person}{Yunhao Tang}, \bibinfo{person}{Shipra
  Agrawal}, {and} \bibinfo{person}{Yuri Faenza}.}
  \bibinfo{year}{2019}\natexlab{}.
\newblock \showarticletitle{Reinforcement learning for integer programming:
  Learning to cut}.
\newblock \bibinfo{journal}{\emph{arXiv preprint arXiv:1906.04859}}
  (\bibinfo{year}{2019}).
\newblock


\bibitem[\protect\citeauthoryear{Tian, Luo, Sun, Ellis, Freeman, Tenenbaum, and
  Wu}{Tian et~al\mbox{.}}{2019}]%
        {tian2018learning}
\bibfield{author}{\bibinfo{person}{Yonglong Tian}, \bibinfo{person}{Andrew
  Luo}, \bibinfo{person}{Xingyuan Sun}, \bibinfo{person}{Kevin Ellis},
  \bibinfo{person}{William~T. Freeman}, \bibinfo{person}{Joshua~B. Tenenbaum},
  {and} \bibinfo{person}{Jiajun Wu}.} \bibinfo{year}{2019}\natexlab{}.
\newblock \showarticletitle{Learning to Infer and Execute 3D Shape Programs}.
  In \bibinfo{booktitle}{\emph{International Conference on Learning
  Representations}}.
\newblock


\bibitem[\protect\citeauthoryear{Vanegas, Garcia-Dorado, Aliaga, Benes, and
  Waddell}{Vanegas et~al\mbox{.}}{2012}]%
        {vanegas2012urban}
\bibfield{author}{\bibinfo{person}{Carlos~A. Vanegas}, \bibinfo{person}{Ignacio
  Garcia-Dorado}, \bibinfo{person}{Daniel~G. Aliaga}, \bibinfo{person}{Bedrich
  Benes}, {and} \bibinfo{person}{Paul Waddell}.}
  \bibinfo{year}{2012}\natexlab{}.
\newblock \showarticletitle{Inverse Design of Urban Procedural Models}.
\newblock \bibinfo{journal}{\emph{ACM Trans. Graph.}} \bibinfo{volume}{31},
  \bibinfo{number}{6}, Article \bibinfo{articleno}{168} (\bibinfo{date}{Nov.}
  \bibinfo{year}{2012}), \bibinfo{numpages}{11}~pages.
\newblock
\showISSN{0730-0301}
\urldef\tempurl%
\url{https://doi.org/10.1145/2366145.2366187}
\showDOI{\tempurl}


\bibitem[\protect\citeauthoryear{Vaswani, Shazeer, Parmar, Uszkoreit, Jones,
  Gomez, Kaiser, and Polosukhin}{Vaswani et~al\mbox{.}}{2017}]%
        {Vaswani2017}
\bibfield{author}{\bibinfo{person}{Ashish Vaswani}, \bibinfo{person}{Noam
  Shazeer}, \bibinfo{person}{Niki Parmar}, \bibinfo{person}{Jakob Uszkoreit},
  \bibinfo{person}{Llion Jones}, \bibinfo{person}{Aidan~N Gomez},
  \bibinfo{person}{\L~ukasz Kaiser}, {and} \bibinfo{person}{Illia Polosukhin}.}
  \bibinfo{year}{2017}\natexlab{}.
\newblock \showarticletitle{Attention is All you Need}. In
  \bibinfo{booktitle}{\emph{Advances in Neural Information Processing
  Systems}}, \bibfield{editor}{\bibinfo{person}{I.~Guyon},
  \bibinfo{person}{U.~V. Luxburg}, \bibinfo{person}{S.~Bengio},
  \bibinfo{person}{H.~Wallach}, \bibinfo{person}{R.~Fergus},
  \bibinfo{person}{S.~Vishwanathan}, {and} \bibinfo{person}{R.~Garnett}}
  (Eds.), Vol.~\bibinfo{volume}{30}. \bibinfo{publisher}{Curran Associates,
  Inc.}, \bibinfo{pages}{5998--6008}.
\newblock
\urldef\tempurl%
\url{https://proceedings.neurips.cc/paper/2017/file/3f5ee243547dee91fbd053c1c4a845aa-Paper.pdf}
\showURL{%
\tempurl}


\bibitem[\protect\citeauthoryear{Veli{\v{c}}kovi{\'c}, Cucurull, Casanova,
  Romero, Lio, and Bengio}{Veli{\v{c}}kovi{\'c} et~al\mbox{.}}{2017}]%
        {velivckovic2017graph}
\bibfield{author}{\bibinfo{person}{Petar Veli{\v{c}}kovi{\'c}},
  \bibinfo{person}{Guillem Cucurull}, \bibinfo{person}{Arantxa Casanova},
  \bibinfo{person}{Adriana Romero}, \bibinfo{person}{Pietro Lio}, {and}
  \bibinfo{person}{Yoshua Bengio}.} \bibinfo{year}{2017}\natexlab{}.
\newblock \showarticletitle{Graph attention networks}.
\newblock \bibinfo{journal}{\emph{arXiv preprint arXiv:1710.10903}}
  (\bibinfo{year}{2017}).
\newblock


\bibitem[\protect\citeauthoryear{Weiler}{Weiler}{1986}]%
        {weiler1986}
\bibfield{author}{\bibinfo{person}{K.J. Weiler}.}
  \bibinfo{year}{1986}\natexlab{}.
\newblock \bibinfo{booktitle}{\emph{Topological structures for geometric
  modeling}}.
\newblock \bibinfo{publisher}{University Microfilms}.
\newblock


\bibitem[\protect\citeauthoryear{Weiss}{Weiss}{2009}]%
        {weiss2009geometry}
\bibfield{author}{\bibinfo{person}{Daniel Weiss}.}
  \bibinfo{year}{2009}\natexlab{}.
\newblock \emph{\bibinfo{title}{Geometry-based structural optimization on CAD
  specification trees}}.
\newblock \bibinfo{thesistype}{Ph.D. Dissertation}. \bibinfo{school}{ETH
  Zurich}.
\newblock


\bibitem[\protect\citeauthoryear{Wu, Zhuang, Xu, Zhang, and Chen}{Wu
  et~al\mbox{.}}{2020}]%
        {wu2020pq}
\bibfield{author}{\bibinfo{person}{Rundi Wu}, \bibinfo{person}{Yixin Zhuang},
  \bibinfo{person}{Kai Xu}, \bibinfo{person}{Hao Zhang}, {and}
  \bibinfo{person}{Baoquan Chen}.} \bibinfo{year}{2020}\natexlab{}.
\newblock \showarticletitle{Pq-net: A generative part seq2seq network for 3d
  shapes}. In \bibinfo{booktitle}{\emph{Proceedings of the IEEE/CVF Conference
  on Computer Vision and Pattern Recognition}}. \bibinfo{pages}{829--838}.
\newblock


\bibitem[\protect\citeauthoryear{Wu, Song, Khosla, Yu, Zhang, Tang, and
  Xiao}{Wu et~al\mbox{.}}{2015}]%
        {wu20153d}
\bibfield{author}{\bibinfo{person}{Zhirong Wu}, \bibinfo{person}{Shuran Song},
  \bibinfo{person}{Aditya Khosla}, \bibinfo{person}{Fisher Yu},
  \bibinfo{person}{Linguang Zhang}, \bibinfo{person}{Xiaoou Tang}, {and}
  \bibinfo{person}{Jianxiong Xiao}.} \bibinfo{year}{2015}\natexlab{}.
\newblock \showarticletitle{3d shapenets: A deep representation for volumetric
  shapes}. In \bibinfo{booktitle}{\emph{Proceedings of the IEEE conference on
  computer vision and pattern recognition}}. \bibinfo{pages}{1912--1920}.
\newblock


\bibitem[\protect\citeauthoryear{Xu, Hu, Leskovec, and Jegelka}{Xu
  et~al\mbox{.}}{2018}]%
        {xu2018powerful}
\bibfield{author}{\bibinfo{person}{Keyulu Xu}, \bibinfo{person}{Weihua Hu},
  \bibinfo{person}{Jure Leskovec}, {and} \bibinfo{person}{Stefanie Jegelka}.}
  \bibinfo{year}{2018}\natexlab{}.
\newblock \showarticletitle{How powerful are graph neural networks?}
\newblock \bibinfo{journal}{\emph{arXiv preprint arXiv:1810.00826}}
  (\bibinfo{year}{2018}).
\newblock


\bibitem[\protect\citeauthoryear{Zhang, Jaiswal, and Rai}{Zhang
  et~al\mbox{.}}{2018}]%
        {zhang2018featurenet}
\bibfield{author}{\bibinfo{person}{Zhibo Zhang}, \bibinfo{person}{Prakhar
  Jaiswal}, {and} \bibinfo{person}{Rahul Rai}.}
  \bibinfo{year}{2018}\natexlab{}.
\newblock \showarticletitle{FeatureNet: machining feature recognition based on
  3D convolution neural network}.
\newblock \bibinfo{journal}{\emph{Computer-Aided Design}}
  \bibinfo{volume}{101} (\bibinfo{year}{2018}), \bibinfo{pages}{12--22}.
\newblock


\bibitem[\protect\citeauthoryear{Zhou and Jacobson}{Zhou and Jacobson}{2016}]%
        {Thingi10K}
\bibfield{author}{\bibinfo{person}{Qingnan Zhou} {and} \bibinfo{person}{Alec
  Jacobson}.} \bibinfo{year}{2016}\natexlab{}.
\newblock \showarticletitle{Thingi10K: A Dataset of 10,000 3D-Printing Models}.
\newblock \bibinfo{journal}{\emph{arXiv preprint arXiv:1605.04797}}
  (\bibinfo{year}{2016}).
\newblock


\bibitem[\protect\citeauthoryear{Zou, Yumer, Yang, Ceylan, and Hoiem}{Zou
  et~al\mbox{.}}{2017}]%
        {zou20173d}
\bibfield{author}{\bibinfo{person}{Chuhang Zou}, \bibinfo{person}{Ersin Yumer},
  \bibinfo{person}{Jimei Yang}, \bibinfo{person}{Duygu Ceylan}, {and}
  \bibinfo{person}{Derek Hoiem}.} \bibinfo{year}{2017}\natexlab{}.
\newblock \showarticletitle{3d-prnn: Generating shape primitives with recurrent
  neural networks}. In \bibinfo{booktitle}{\emph{Proceedings of the IEEE
  International Conference on Computer Vision}}. \bibinfo{pages}{900--909}.
\newblock


\end{thebibliography}
\bibliographystyle{ACM-Reference-Format}

\appendix
\clearpage
\section{Appendix}

\subsection{Fusion 360 Gallery Reconstruction Dataset}
\label{section:appendix_dataset}
In this section we provide additional details on the \FRec{}.

\subsubsection{Data Processing}
\begin{figure}[b]
    \includegraphics[width=\columnwidth]{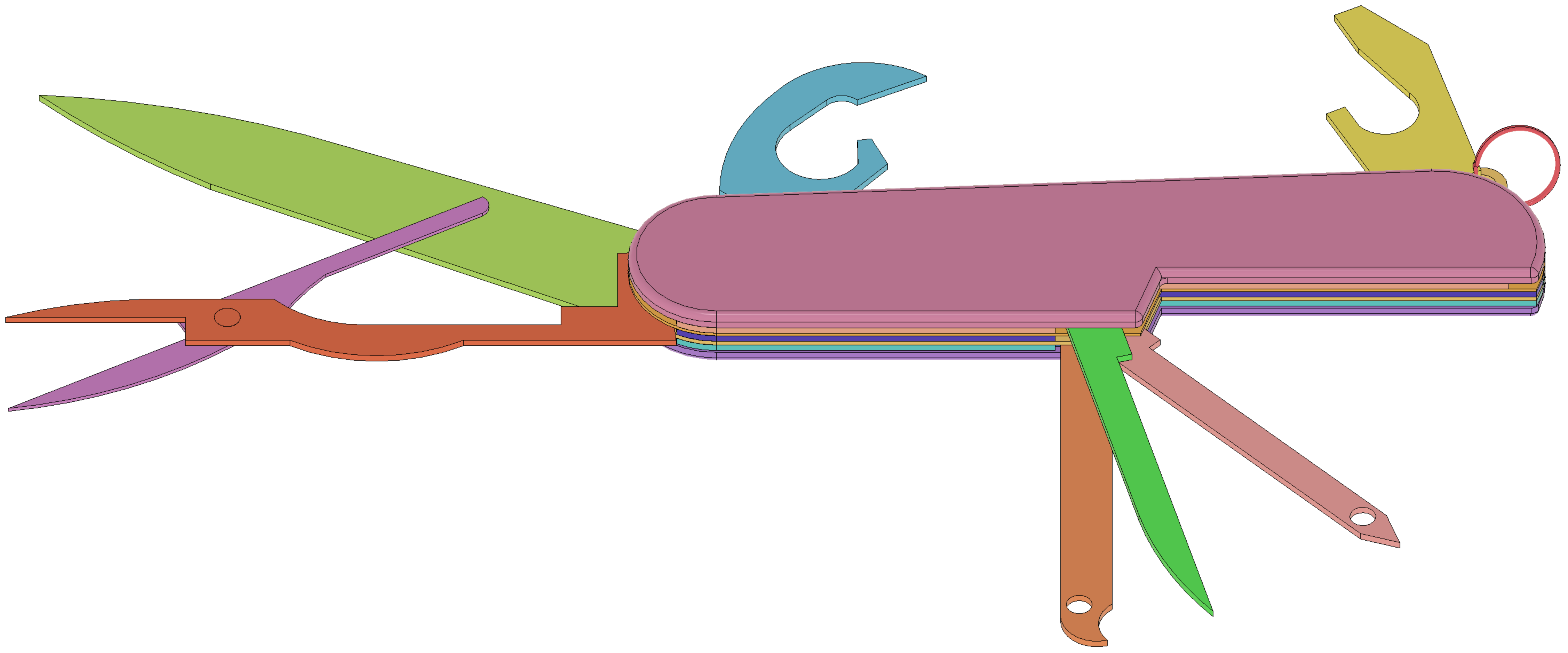}
    \caption{An example multi-component assembly that is broken up into separate designs (highlighted with color), each with an independent construction sequence. }
    \label{figure:assembly}
\end{figure}

To process the data we use the Fusion 360 Python API to parse the native Fusion 360 .f3d files. Figure~\ref{figure:assembly} shows an example assembly that is split up to produce multiple designs with independent construction sequences. The rounded edges are removed by suppressing fillets in the parametric CAD file. During processing color and material information is also removed.

\begin{figure*}
    \includegraphics[width=0.9\textwidth]{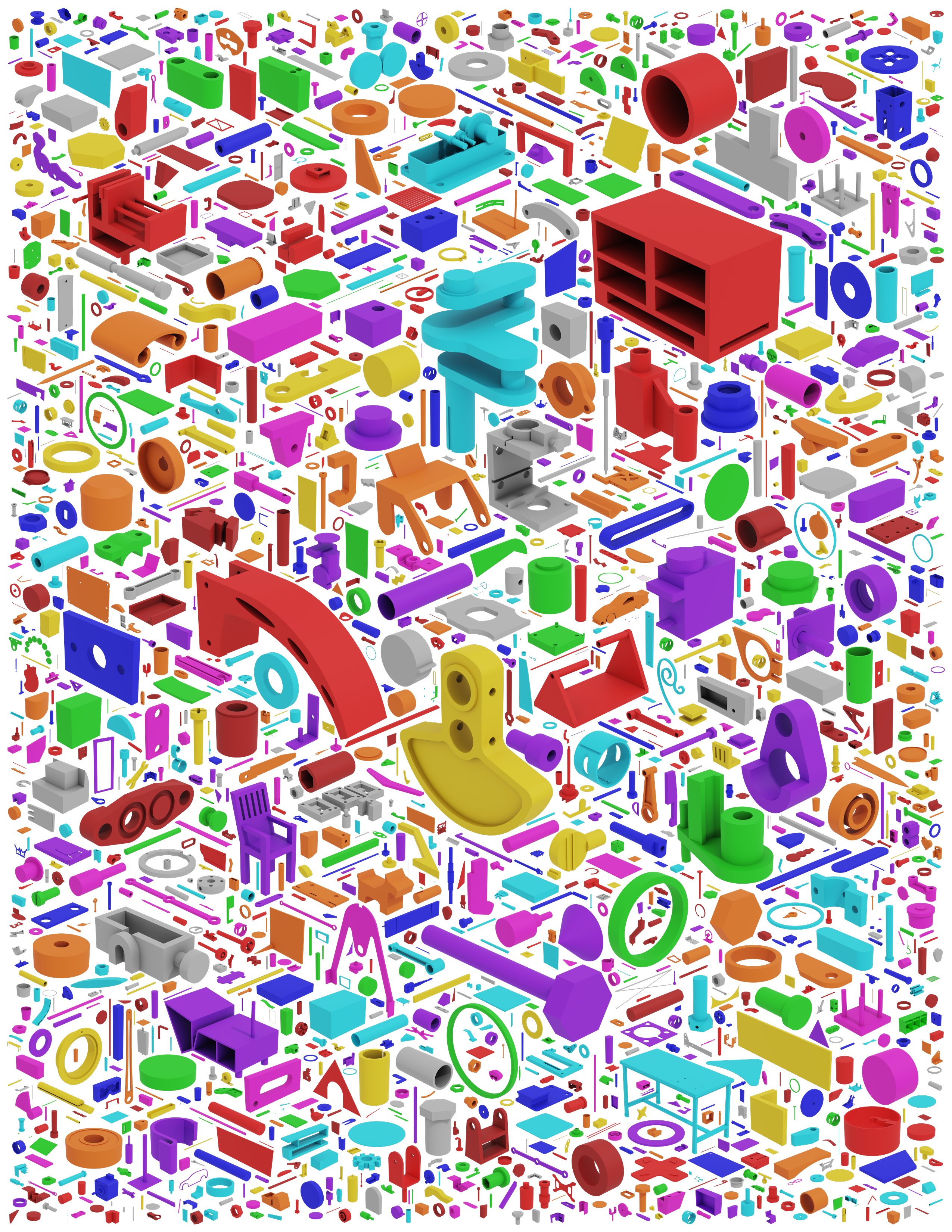}
    \caption{ A random sampling of designs from the \textit{Fusion 360 Gallery} reconstruction dataset. }
    \label{figure:dataset_mosaic_appendix}
\end{figure*}

After each construction sequence has been extracted we perform reconstruction and compare the reconstructed design to the original to ensure data validity. Failure cases and any duplicate designs, are not included in the dataset. We consider a design a duplicate when there is an exact match in all of the following: body count, face count, surface area to one decimal point, volume to one decimal point, and for each extrude in the construction sequence: extrude profile count, extrude body count, extrude face count, extrude side face count, extrude end face count, and extrude start face count. This process allows us to match designs that have been translated or rotated, while considering designs unique if they have matching geometry but different construction sequences. Duplicates account for approximately 5,000 designs. Figure~\ref{figure:dataset_mosaic_appendix} shows a random sampling of designs from the reconstruction dataset.

\subsubsection{Geometry Data Format}
\label{section:appendix_data_format}
As described in Section~\ref{section:dataset_geometry}, we provide geometry in several data formats that we provided additional details on in this section.

\paragraph{Boundary Representation} A B-Rep consists of faces, edges, loops, coedges and vertices \citep{weiler1986}. A face is a connected region of the model's surface. An edge defines the curve where two faces meet and a vertex defines the point where edges meet. Faces have an underlying parametric surface which is divided into visible and hidden regions by a series of boundary loops. A set of connected faces forms a body.

B-Rep data is provided as .smt files representing the ground truth geometry and .step as an alternate neutral B-Rep file format. The .smt file format is the native format used by Autodesk Shape Manager, the CAD kernel within Fusion 360, and has the advantage of minimizing conversion errors.

\paragraph{Mesh} Mesh data is provided in .obj format representing a triangulated version of the B-Rep. Each B-Rep face is triangulated separately and is therefore not manifold. 

Other representations, such as point clouds or voxels, can be generated using existing data conversion routines and are not included in the dataset. For convenience we include a thumbnail .png image file together with each geometry.

Files are provided in a single directory, with a naming convention as follows: \code{XXXXX\_YYYYYYYY\_ZZZZ[\_1234].ext}. Here \code{XXXXX} represents the project, \code{YYYYYYYY} the file, \code{ZZZZ} the component, and \code{\_1234} the extrude index. If \code{\_1234} is absent the file represents the final design.

\subsubsection{Design Complexity}
\begin{figure*}
    \includegraphics[width=\textwidth]{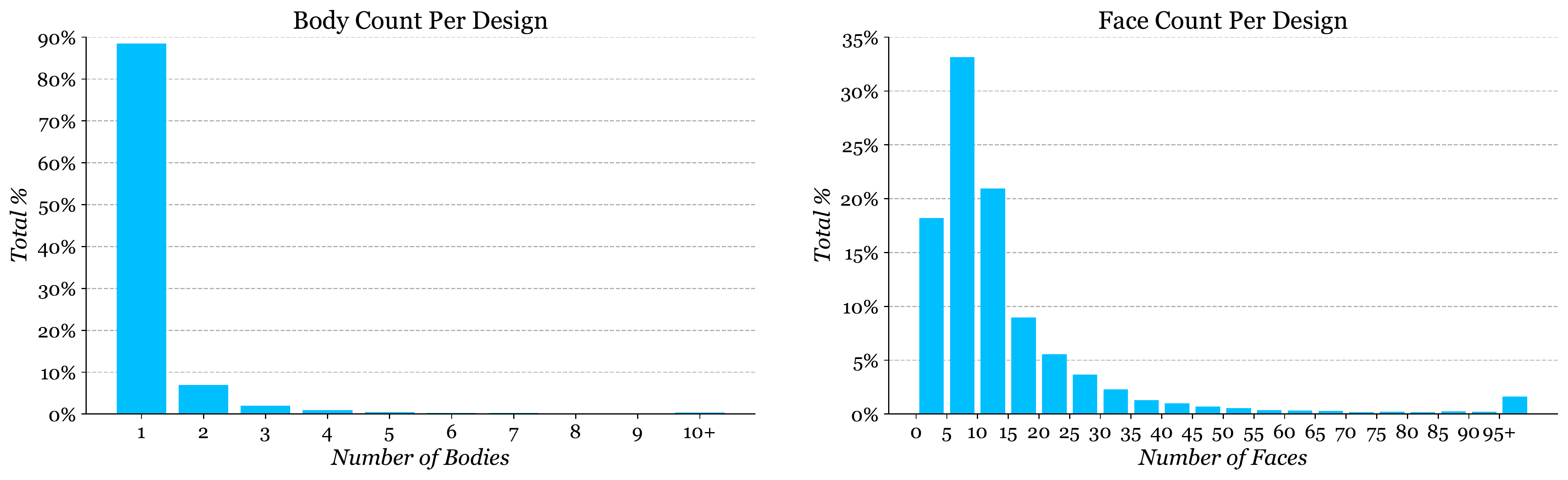}
    \caption{Left: The number of bodies per design shown as a distribution. Right: The number of B-Rep faces per design shown as a distribution. }
    \label{figure:data_stats_bodyface}
\end{figure*}

A key goal of the reconstruction dataset is to provide a suitably scoped baseline for learning-based approaches to CAD reconstruction. Restricting the modeling operations to \textit{sketch} and \textit{extrude} vastly narrows the design space and enables simpler shape grammars for reconstruction. Each design represents a component in Fusion 360 that can have multiple geometric bodies. Figure~\ref{figure:data_stats_bodyface} (left) illustrates that the vast majority of designs have a single body. The number of B-Rep faces in each design gives a good indication of the complexity of the dataset. Figure~\ref{figure:data_stats_bodyface} (right) shows the number of faces per design as a distribution, with the peak being between 5-10 faces per design. As we do not filter any of the designs based on complexity, this distribution reflects real designs where simple washers and flat plates are common components in mechanical assemblies. 

\begin{figure*}
    \includegraphics[width=\textwidth]{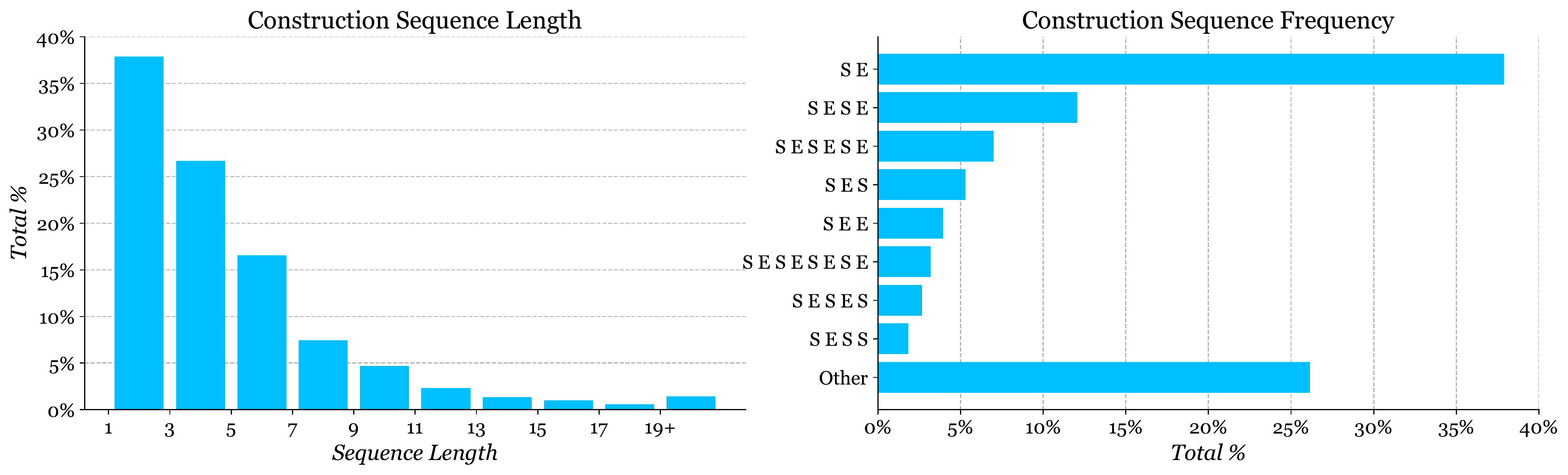}
    \caption{Left: The distribution of construction sequence length. Right: The distribution of common construction sequences. S indicates a \textit{Sketch} and E indicates an \textit{Extrude} operation. }
    \label{figure:data_stats_timeline}
\end{figure*}

\subsubsection{Construction Sequence}
The construction sequence is the series of \textit{sketch} and \textit{extrude} operations that are executed to produce the final geometry. We provide the construction sequence in a JSON format text file.  Each step in the construction sequence has associated parameters that are stored in that entity. For example, \textit{sketch} entities will store the curves that make up the sketch. Each construction sequence must have at least one \textit{sketch} and one \textit{extrude} step, for a minimum of two steps. The average number of steps is 4.74, the median 4, the mode 2, and the maximum 61. Figure~\ref{figure:data_stats_timeline} illustrates the distribution of construction sequence length and the most frequent construction sequence combinations.

With access to the full parametric history, it is possible to extract numerous relationships from the dataset that can be used for learning. Starting at a high level, we know the order of modeling operations in the construction sequence. The sketch geometry, B-Rep faces, and triangles derived from them, can be traced back to a position in the construction sequence. The type of geometry created by each modeling operation is also known. For example, sketches create trimmed profiles where the curves intersect to form closed loops; extrude operations produce B-Rep faces with information such as which faces were on the side or ends of an extrusion. In addition, the sequence of B-Rep models themselves contain valuable topology information that can be leveraged, such as the connectivity of B-Rep faces and edges. Finally geometric information like points and normal vectors can be sampled from the parametric surfaces. Feature diversity enables many different learning representations and architectures to be leveraged and compared.

\begin{figure}
    \begin{center}
        \includegraphics[width=1.0\columnwidth]{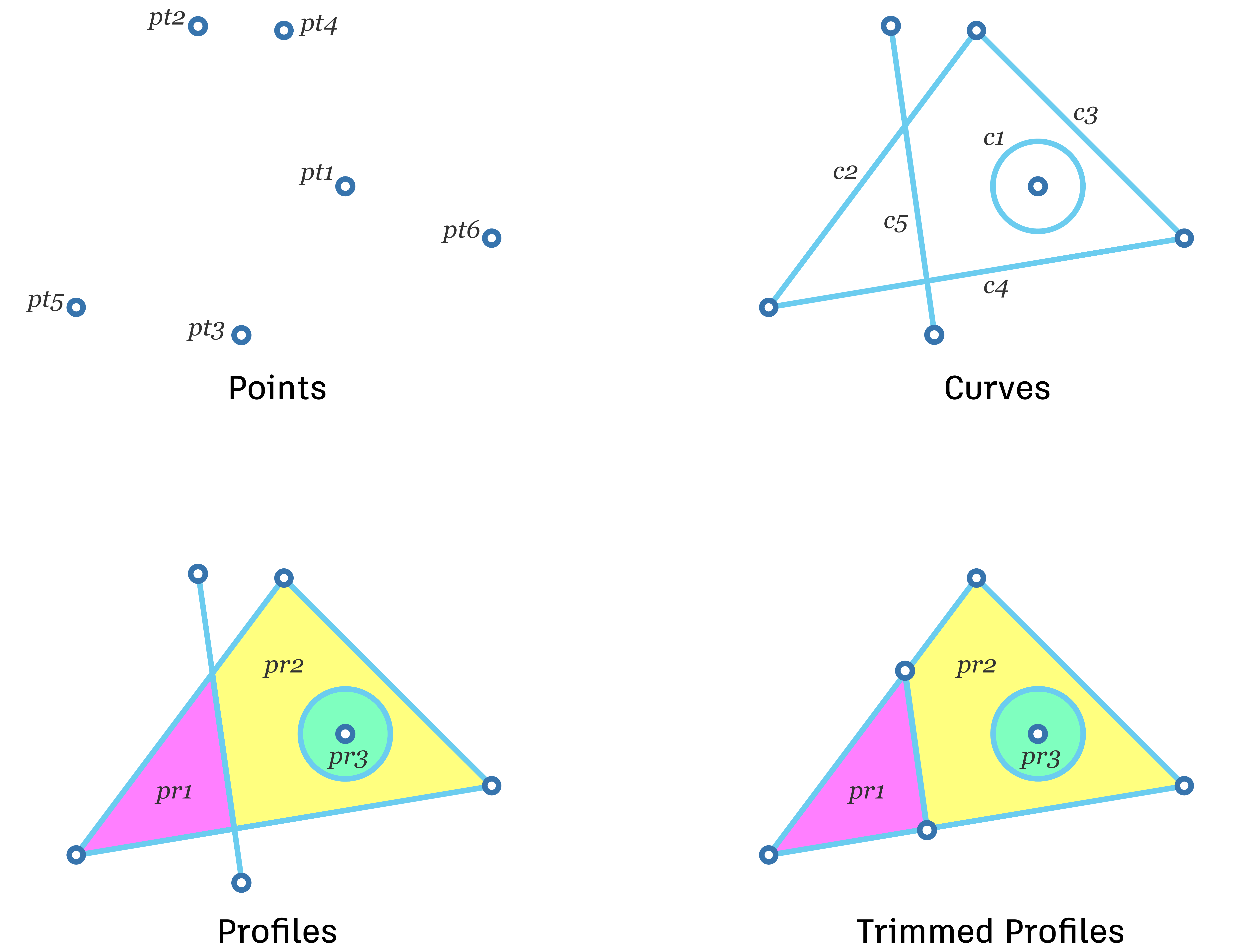}
        \caption{Sketch primitives.}
        \label{figure:dataset_sketch_entities}
    \end{center}
\end{figure}

\subsubsection{Sketch}
In this section we describe the sketch data in further detail and present statistics illustrating the data distribution. Figure~\ref{figure:dataset_sketch_entities} illustrates the geometric 2D primitives, described in section~\ref{section:sketch}, that make up a sketch. Sketches are represented as a series of points ($pt1$...$pt6$), that create curves ($c1$...$c5$), that in turn create profiles ($pr1$...$pr3$), illustrated with separate colors. Profiles can have inner loops to create holes, $c1$ is the inner loop of $pr2$ and the outer loop of $pr3$. Profiles also have a trimmed representation that contains only closed loops without open curves. The trimmed representation is shown in the lower right of Figure~\ref{figure:dataset_sketch_entities} where the $c5$ is trimmed and incorporated into $pr1$ and $pr2$.

\paragraph{Points} Each point is provided with a universally unique identifier (UUID) key and a \href{http://help.autodesk.com/cloudhelp/ENU/Fusion-360-API/files/Point3D.htm}{\code{Point3D}} data structure with $x$, $y$, and $z$. Sketch primitives are drawn in a local 2D coordinate system and later transformed into world coordinates. As such all sketch points have a $z$ value of 0.

\begin{figure*}
    \includegraphics[width=\textwidth]{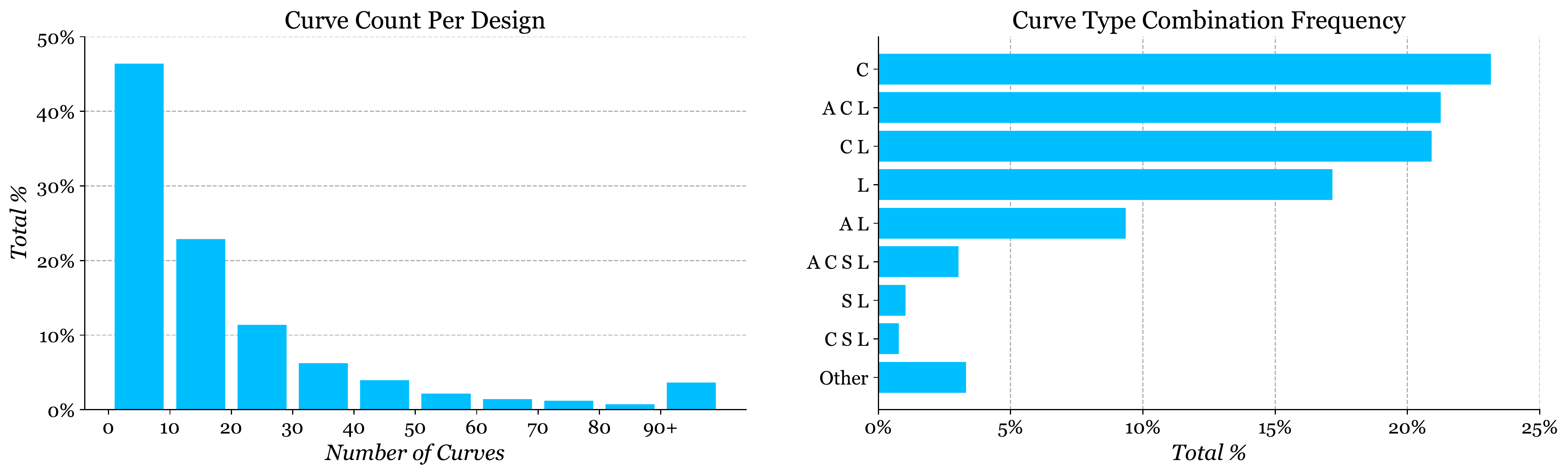}
    \caption{Left: The number of curves in each design, shown as a distribution. Right: Common curve combinations in each design, shown as a distribution. Each curve type is abbreviated as follows: C - \textit{SketchCircle}, A - \textit{SketchArc}, L - \textit{SketchLine}, S - \textit{SketchFittedSpline}. }
    \label{figure:data_stats_curves}
\end{figure*}

\begin{figure}
    \includegraphics[width=1.0\columnwidth]{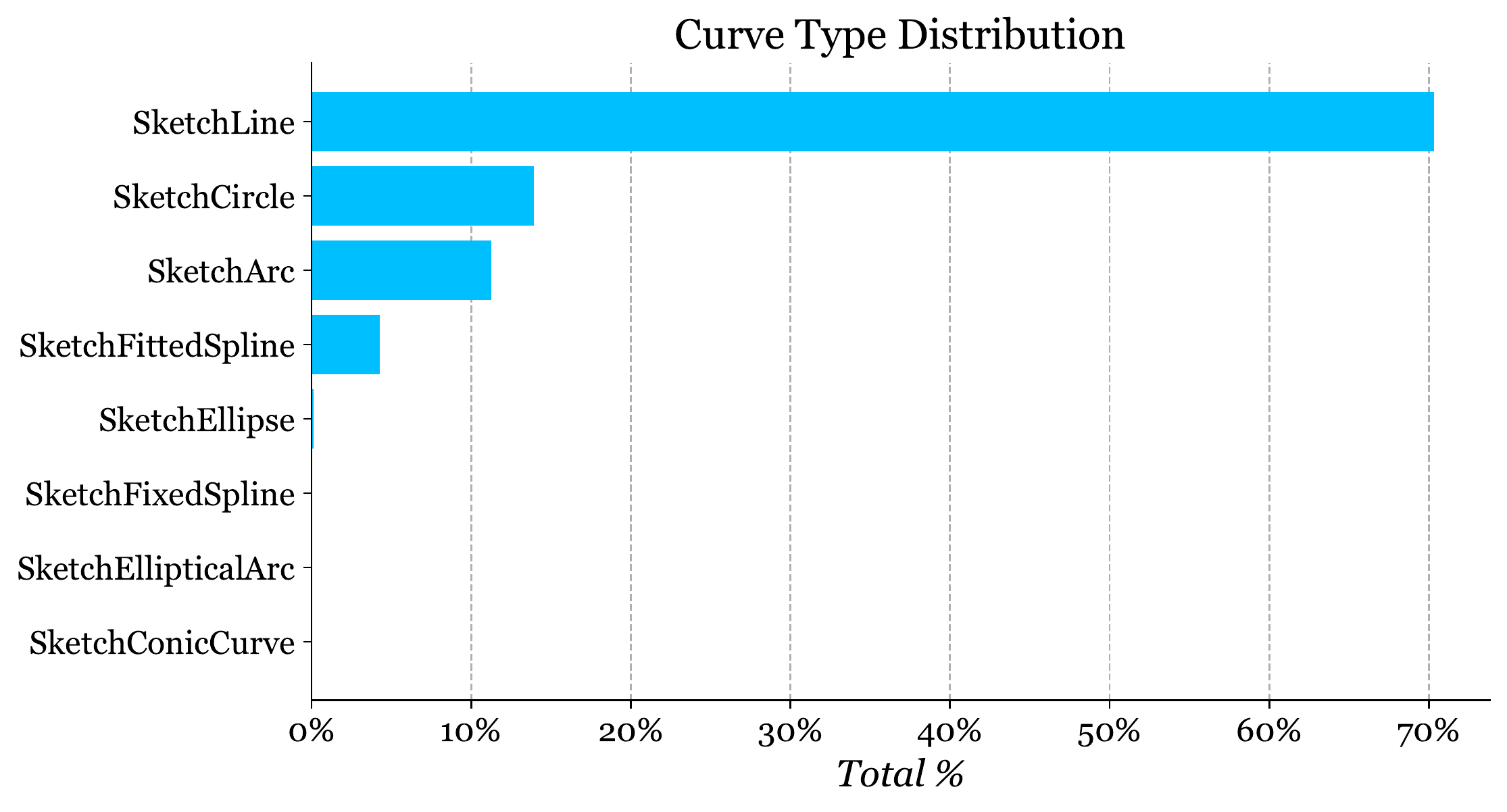}
    \caption{The distribution of curve types.}
    \label{figure:data_stats_curves_types}
\end{figure}

\paragraph{Curves} Each curve has a UUID key and a \href{https://help.autodesk.com/cloudhelp/ENU/Fusion-360-API/files/SketchCurve.htm}{\code{SketchCurve}} that can represent the curve types listed below. The parameters for each curve type can be referenced via the Fusion 360 API documentation linked below. 
\begin{itemize}
    \small
    \item \href{https://help.autodesk.com/cloudhelp/ENU/Fusion-360-API/files/SketchArc.htm}{\code{SketchArc}}
    \item \href{https://help.autodesk.com/cloudhelp/ENU/Fusion-360-API/files/SketchCircle.htm}{\code{SketchCircle}}
    \item \href{https://help.autodesk.com/cloudhelp/ENU/Fusion-360-API/files/SketchConicCurve.htm}{\code{SketchConicCurve}}
    \item \href{https://help.autodesk.com/cloudhelp/ENU/Fusion-360-API/files/SketchEllipse.htm}{\code{SketchEllipse}}
    \item \href{https://help.autodesk.com/cloudhelp/ENU/Fusion-360-API/files/SketchEllipticalArc.htm}{\code{SketchEllipticalArc}}
    \item \href{https://help.autodesk.com/cloudhelp/ENU/Fusion-360-API/files/SketchFittedSpline.htm}{\code{SketchFittedSpline}}
    \item \href{https://help.autodesk.com/cloudhelp/ENU/Fusion-360-API/files/SketchFixedSpline.htm}{\code{SketchFixedSpline}}
    \item \href{https://help.autodesk.com/cloudhelp/ENU/Fusion-360-API/files/SketchLine.htm}{\code{SketchLine}}
\end{itemize}

Figure~\ref{figure:data_stats_curves} illustrates the distribution of curve count per design and the frequency that different curve combinations are used together in a design. Figure~\ref{figure:data_stats_curves_types} shows the overall distribution of curve types in the dataset. It is notable that mechanical CAD sketches rely heavily on lines, circles, and arcs rather than spline curves.

\paragraph{Profiles} Profiles represent a collection of curves that join together to make a closed loop. In \textit{Fusion 360} profiles are automatically generated from arbitrary curves that don't necessarily connect at the end points. In Figure~\ref{figure:dataset_sketch_entities} two profiles ($pr1$ and $pr2$) are generated when the line crosses the triangle. We provide both the original curves (Figure~\ref{figure:dataset_sketch_entities}, top right) used to generate the profiles (Figure~\ref{figure:dataset_sketch_entities}, bottom left) and the trimmed profile information containing just the closed profile loop (Figure~\ref{figure:dataset_sketch_entities}, bottom right). Loops within profiles have a flag that can be set to specify holes.

\paragraph{Dimensions} 
User specified sketch dimensions are used to define set angles, diameters, distances etc. between sketch geometry to constraint the sketch as it is edited. Each dimension has a UUID key and a \href{https://help.autodesk.com/cloudhelp/ENU/Fusion-360-API/files/SketchDimension.htm}{\code{SketchDimension}} that can represent the dimension types listed below. Each dimension references one or more curves by UUID. The parameters for each dimension type can be referenced via the Fusion 360 API documentation linked below. 
\begin{itemize}
    \small
    \item \href{https://help.autodesk.com/cloudhelp/ENU/Fusion-360-API/files/SketchAngularDimension.htm}{\code{SketchAngularDimension}}
    \item \href{https://help.autodesk.com/cloudhelp/ENU/Fusion-360-API/files/SketchConcentricCircleDimension.htm}{\code{SketchConcentricCircleDimension}}
    \item \href{https://help.autodesk.com/cloudhelp/ENU/Fusion-360-API/files/SketchDiameterDimension.htm}{\code{SketchDiameterDimension}}
    \item \href{https://help.autodesk.com/cloudhelp/ENU/Fusion-360-API/files/SketchEllipseMajorRadiusDimension.htm}{\code{SketchEllipseMajorRadiusDimension}}
    \item \href{https://help.autodesk.com/cloudhelp/ENU/Fusion-360-API/files/SketchEllipseMinorRadiusDimension.htm}{\code{SketchEllipseMinorRadiusDimension}}
    \item \href{https://help.autodesk.com/cloudhelp/ENU/Fusion-360-API/files/SketchLinearDimension.htm}{\code{SketchLinearDimension}}
    \item \href{https://help.autodesk.com/cloudhelp/ENU/Fusion-360-API/files/SketchOffsetCurvesDimension.htm}{\code{SketchOffsetCurvesDimension}}
    \item \href{https://help.autodesk.com/cloudhelp/ENU/Fusion-360-API/files/SketchOffsetDimension.htm}{\code{SketchOffsetDimension}}
    \item \href{https://help.autodesk.com/cloudhelp/ENU/Fusion-360-API/files/SketchRadialDimension.htm}{\code{SketchRadialDimension}}
\end{itemize}

\begin{figure*}
    \includegraphics[width=\textwidth]{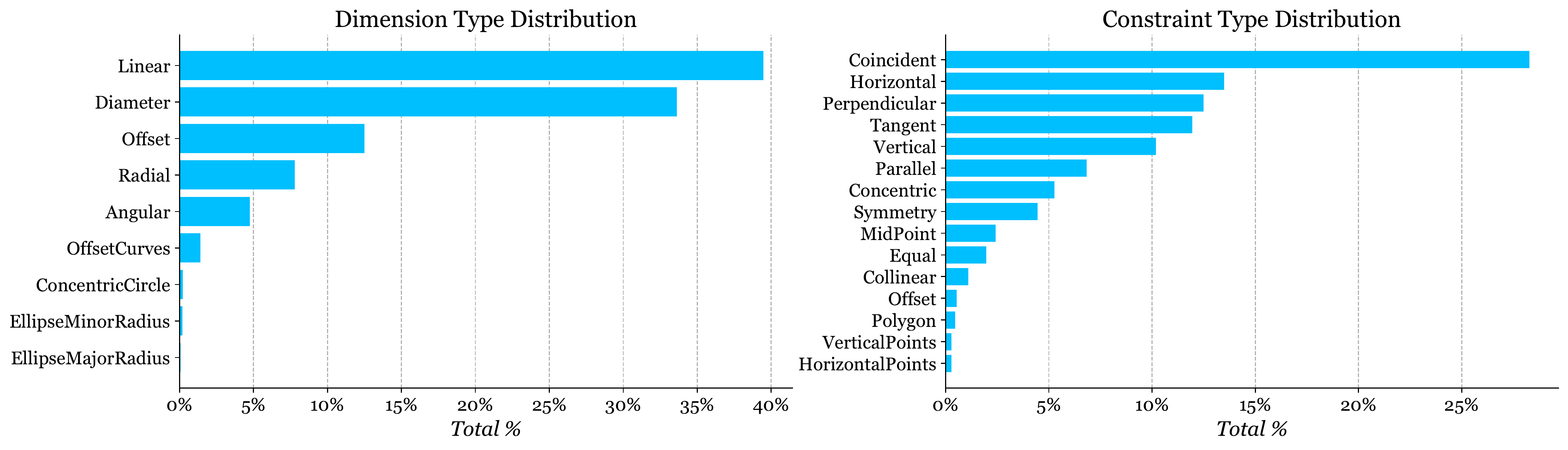}
    \caption{The distribution of constraint (left) and dimension (right) types.}
    \label{figure:data_stats_constraint_dimension}
\end{figure*}

\paragraph{Constraints} Constraints define geometric relationships between sketch geometry. For example, a symmetry constraint enables the user to have geometry mirrored, or a parallel constraint ensures two lines are always parallel. Each constraint has a UUID key and a \href{https://help.autodesk.com/cloudhelp/ENU/Fusion-360-API/files/GeometricConstraint.htm}{\code{GeometricConstraint}} that can represent the constraint types listed below. Each constraint references one or more curves by UUID. The parameters for each constraint type can be referenced via the Fusion 360 API documentation linked below. 
\begin{itemize}
    \small
    \item \href{https://help.autodesk.com/cloudhelp/ENU/Fusion-360-API/files/CircularPatternConstraint.htm}{\code{CircularPatternConstraint}}
    \item \href{https://help.autodesk.com/cloudhelp/ENU/Fusion-360-API/files/CoincidentConstraint.htm}{\code{CoincidentConstraint}}
    \item \href{https://help.autodesk.com/cloudhelp/ENU/Fusion-360-API/files/CollinearConstraint.htm}{\code{CollinearConstraint}}
    \item \href{https://help.autodesk.com/cloudhelp/ENU/Fusion-360-API/files/ConcentricConstraint.htm}{\code{ConcentricConstraint}}
    \item \href{https://help.autodesk.com/cloudhelp/ENU/Fusion-360-API/files/EqualConstraint.htm}{\code{EqualConstraint}}
    \item \href{https://help.autodesk.com/cloudhelp/ENU/Fusion-360-API/files/HorizontalConstraint.htm}{\code{HorizontalConstraint}}
    \item \href{https://help.autodesk.com/cloudhelp/ENU/Fusion-360-API/files/HorizontalPointsConstraint.htm}{\code{HorizontalPointsConstraint}}
    \item \href{https://help.autodesk.com/cloudhelp/ENU/Fusion-360-API/files/MidPointConstraint.htm}{\code{MidPointConstraint}}
    \item \href{https://help.autodesk.com/cloudhelp/ENU/Fusion-360-API/files/OffsetConstraint.htm}{\code{OffsetConstraint}}
    \item \href{https://help.autodesk.com/cloudhelp/ENU/Fusion-360-API/files/ParallelConstraint.htm}{\code{ParallelConstraint}}
    \item \href{https://help.autodesk.com/cloudhelp/ENU/Fusion-360-API/files/PerpendicularConstraint.htm}{\code{PerpendicularConstraint}}
    \item \href{https://help.autodesk.com/cloudhelp/ENU/Fusion-360-API/files/PolygonConstraint.htm}{\code{PolygonConstraint}}
    \item \href{https://help.autodesk.com/cloudhelp/ENU/Fusion-360-API/files/RectangularPatternConstraint.htm}{\code{RectangularPatternConstraint}}
    \item \href{https://help.autodesk.com/cloudhelp/ENU/Fusion-360-API/files/SmoothConstraint.htm}{\code{SmoothConstraint}}
    \item \href{https://help.autodesk.com/cloudhelp/ENU/Fusion-360-API/files/SymmetryConstraint.htm}{\code{SymmetryConstraint}}
    \item \href{https://help.autodesk.com/cloudhelp/ENU/Fusion-360-API/files/TangentConstraint.htm}{\code{TangentConstraint}}
    \item \href{https://help.autodesk.com/cloudhelp/ENU/Fusion-360-API/files/VerticalConstraint.htm}{\code{VerticalConstraint}}
    \item \href{https://help.autodesk.com/cloudhelp/ENU/Fusion-360-API/files/VerticalPointsConstraint.htm}{\code{VerticalPointsConstraint}}
\end{itemize}

Figure~\ref{figure:data_stats_constraint_dimension} illustrates the distribution of dimension and constraint types in the dataset.

\subsubsection{Extrude}
\label{section:appendix_dataset_extrude}
\begin{figure}
    \begin{center}
        \includegraphics[width=0.85\columnwidth]{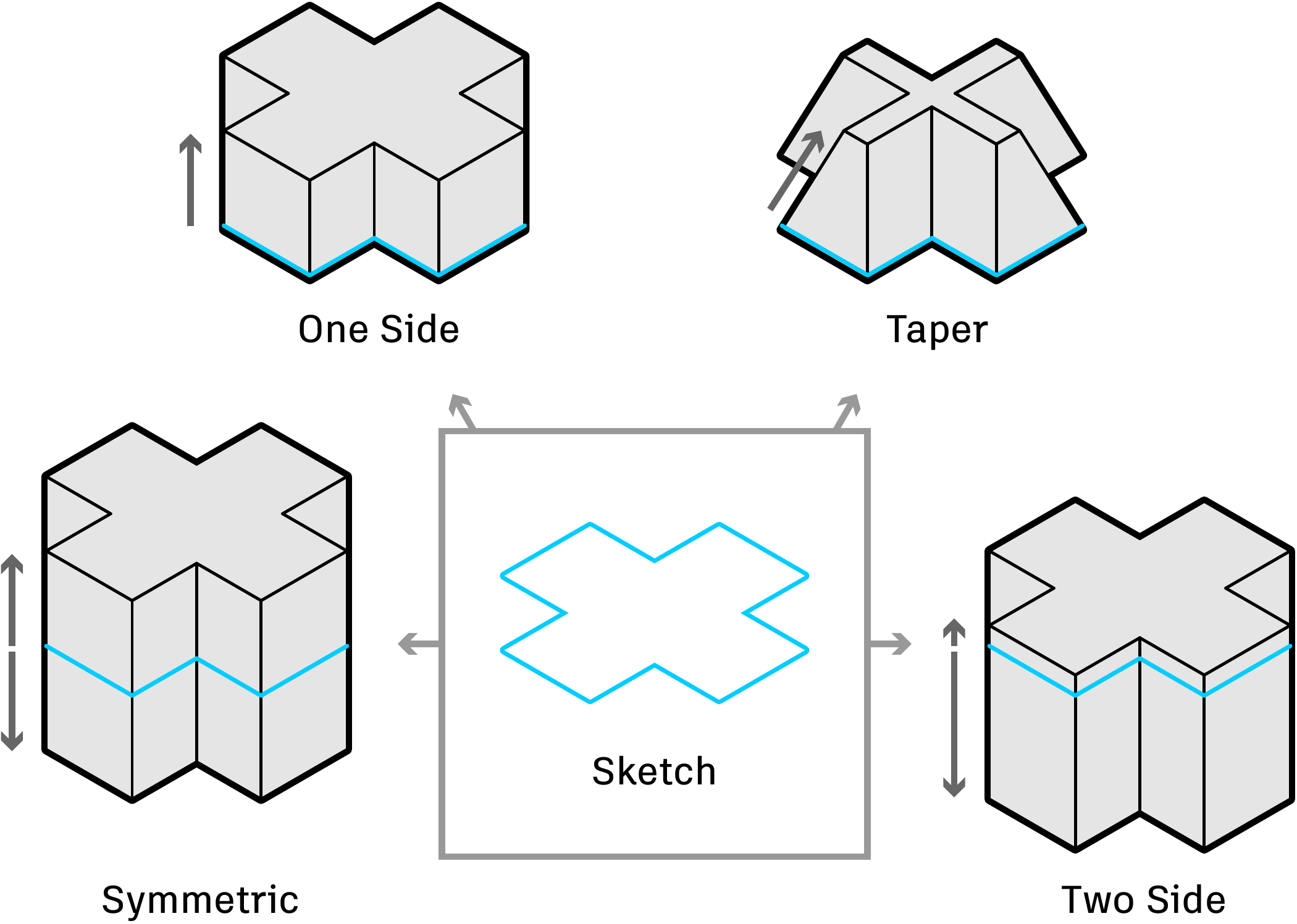}
        \caption{An extrude can be expressed in several different ways: perpendicular from a sketch for a set distance along one side, a symmetrical distance along both sides, or separate distances along two sides. Additionally the extrude can be tapered at an angle.}
        \label{figure:dataset_extrude_types}
    \end{center}
\end{figure}

In this section we describe the extrude data in further detail and present statistics illustrating the data distribution.
Extrude operations have a number of parameters that are set by the user while designing. 
Figure~\ref{figure:dataset_extrude_types} shows how a sketch (left) can be extruded a set distance on one side, symmetrically on two sides, with different distances on each side, as well as tapered. The first extrude operation of a construction sequence always creates a new body, with subsequent extrudes interacting with that body via Boolean operations. 

\begin{figure*}
    \begin{center}
        \includegraphics[width=1.0\textwidth]{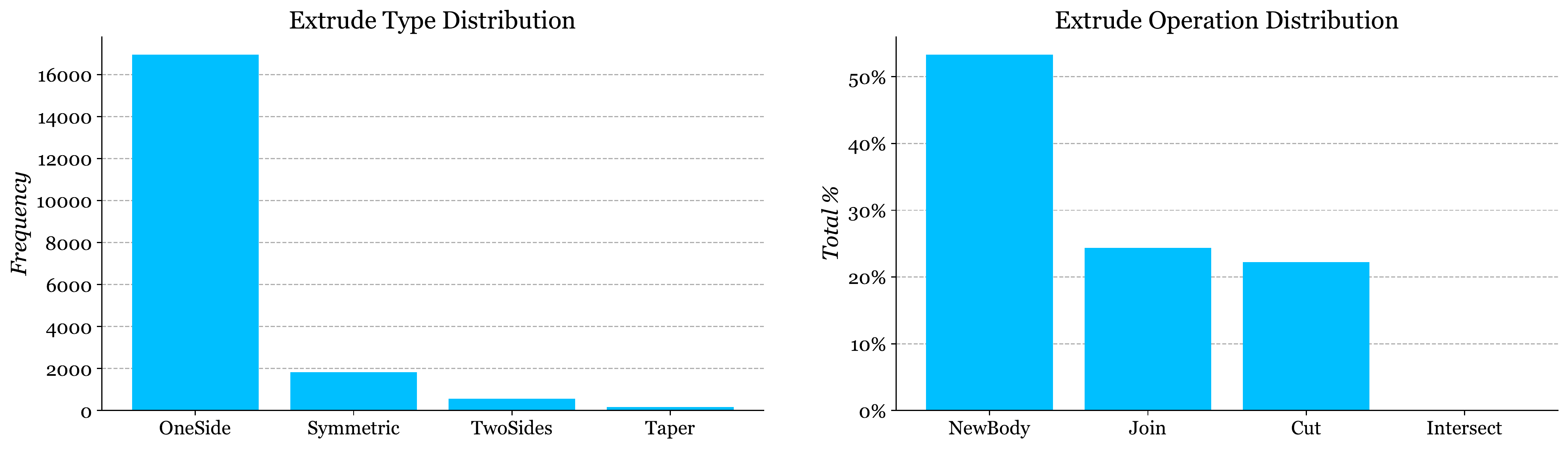}
        \caption{The distribution of extrude types (left) and operations (right).}
        \label{figure:data_stats_extrude}
    \end{center}
\end{figure*}

Figure~\ref{figure:data_stats_extrude} outlines the distribution of different extrude types and operations. Note that tapers can be applied in addition to any extrude type, so the overall frequency of each is shown rather than a relative percentage. 

\subsection{Fusion 360 Gym}
\label{section:gym}

In this section we provide additional information about the functionality available in the \FGym{}. 
The \FGym{} requires the Autodesk Fusion 360 desktop CAD application, available on both macOS and Windows for free to the academic community. Although Fusion 360 is a cloud connected desktop application, the \FGym{} does all processing locally. The \FGym{} consists of a \textit{server} that runs inside of Fusion 360 and receives commands from a \textit{client} running externally. 
Multiple instances of the \FGym{} \textit{server} can be run in parallel. The remainder of this section introduces the available commands from the \textit{client}.   

\subsubsection{Reconstruction Commands}
\label{section:reconstruction_commands}
Reconstruction commands can reconstruct the existing designs at different granularity levels from json files provided with the \FRec{}.

\begin{itemize}
    \small
    \item \code{reconstruct(file)}: reconstruct an entire design from the provided json file.
    \item \code{reconstruct\_sketch(sketch\_data, sketch\_plane, scale, \\translate, rotate)}: reconstruct a sketch from the provided sketch data. A \code{sketch\_plane} can be either: (1) a string value representing a construction plane: \code{XY}, \code{XZ}, or \code{YZ}; (2) a B-Rep planar face id; or (3) a point3d on a planar face of a B-Rep. 
    \item \code{reconstruct\_profile(sketch\_data, sketch\_name, profile\_id, scale, translate, rotate)}: reconstruct a single profile from the provide sketch data, a sketch name, and a profile id.
    \item \code{reconstruct\_curve(sketch\_data, sketch\_name, curve\_id, \\scale, translate, rotate)}: reconstruct a single curve from the provide sketch data, a sketch name, and a curve id.
    \item \code{reconstruct\_curves(sketch\_data, sketch\_name, scale, \\translate, rotate)}: reconstruct all curves from the provide sketch data and a sketch name.
\end{itemize}

\subsubsection{Target Reconstruction Commands}
Target reconstruction commands set the target design to be used with reconstruction.

\begin{itemize}
    \small
    \item \code{set\_target(file)}: set the target to be reconstructed with a .step or .smt file. The call returns a face adjacency graph representing the B-Rep geometry/topology and a \textit{bounding\_box} of the target that can be used for normalization.
    \item \code{revert\_to\_target()}: revert to the target design, removing all reconstruction geometry.
\end{itemize}

\subsubsection{Sketch Extrusion Commands}
Sketch extrusion commands allows users to incrementally create new designs by generating the underlying sketch primitives and extruding them by an arbitrary amount.
\begin{itemize}
    \small 
    \item \code{add\_sketch(sketch\_plane)}: add a sketch to the design. \\A \code{sketch\_plane} can be either: (1) a string value representing a construction plane: \code{XY}, \code{XZ}, or \code{YZ}; (2) a B-Rep planar face id; or (3) a point3d on a planar face of a B-Rep.
    \item \code{add\_point(sketch\_name, p, transform)}: add a point to create a new sequential line in the given sketch. \code{p} is either a point in the 2D sketch space or a point in the 3D world coordinate space if \code{transform="world"} is specified. 
    \item \code{add\_line(sketch\_name, p1, p2, transform)}: add a line to the given sketch. \code{p1} and \code{p2} are the same as defined in \code{add\_point()}.
    \item \code{add\_arc(sketch\_name, p1, p2, angle, transform)}: add an arc to the given sketch. \code{p1} is the start point of the arc and \code{p2} is the center point of the arc. Other properties of \code{p1} and \code{p2} are the same as defined in \code{add\_point()}. \code{angle} is the arc's angle, measured in degrees.
    \item \code{add\_circle(sketch\_name, p, radius, transform)}: add a circle to the given sketch. \code{p} is the center point of the circle. Other properties of \code{p} are the same as defined in \code{add\_point()}. \code{radius} is the radius of the circle. 
    \item \code{close\_profile(sketch\_name)}: close the current set of lines to create one or more profiles by joining the first point to the last point.
    \item \code{add\_extrude(sketch\_name, profile\_id, distance, operation)}: add an extrude to the design. Four operations are supported: 
    \begin{itemize}
        \small 
        \item \code{JoinFeatureOperation}
        \item \code{CutFeatureOperation}
        \item \code{IntersectFeatureOperation}
        \item \code{NewBodyFeatureOperation} 
    \end{itemize}
    It returns a data structure with:
    \begin{itemize}
        \small 
        \item \code{extrude}: B-Rep face information, including vertices, generated from the extrusion.
        \item \code{graph}: face adjacency graph of the current design in "PerFace" format.
        \item \code{bounding\_box}: bounding box of the current design that can be used for normalization.
        \item \code{iou}: intersection over union result if a target design has been set with \code{set\_target()}. 
    \end{itemize}
\end{itemize}

\subsubsection{Face Extrusion Commands}
Face extrusion commands enable a target design to be reconstructed using extrude operations from face to face. 

\begin{itemize}
    \small
    \item \code{add\_extrude\_by\_target\_face(start\_face, end\_face, \\operation)}: add an extrude between two faces of the target. Four operations are supported:
    \begin{itemize}
        \small 
        \item \code{JoinFeatureOperation}
        \item \code{CutFeatureOperation}
        \item \code{IntersectFeatureOperation}
        \item \code{NewBodyFeatureOperation} 
    \end{itemize}
    \item \code{add\_extrudes\_by\_target\_face(actions, revert)}: execute multiple extrude operations, between two faces of the target, in sequence.
\end{itemize}

\begin{figure*}
    \includegraphics[width=0.95\textwidth]{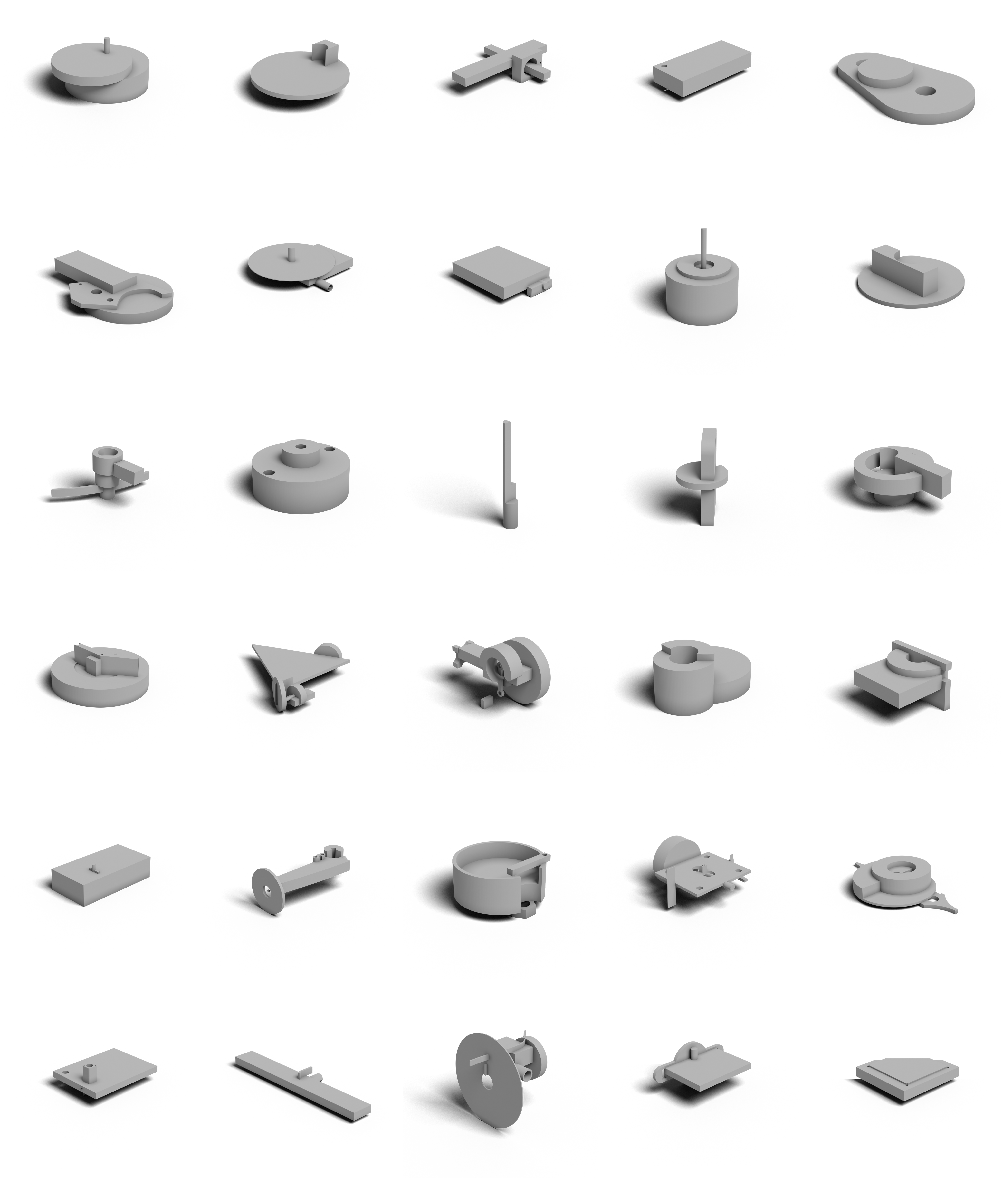}
    \caption{Example designs created using randomized reconstruction commands.}
    \label{figure:randomized_reconstruction}
\end{figure*}

\subsubsection{Randomized Reconstruction Commands}
Randomized reonstruction commands allow users to sample designs, sketches, and profiles from existing designs in the \textit{Fusion 360 Gallery} and support distribution matching of parameters, in support of generations of semi-synthetic data. Figure~\ref{figure:randomized_reconstruction} shows example designs created using randomized reconstruction commands.
\begin{itemize}
    \small
    \item \code{get\_distributions\_from\_dataset(data\_dir, filter, \\split\_file)}: get a list of distributions from the provided dataset. The command currently supports the following distributions: 
    \begin{itemize}
        \small
        \item \code{the starting sketch place}
        \item \code{the number of faces}
        \item \code{the number of extrusions}
        \item \code{the length of sequences}
        \item \code{the number of curves}
        \item \code{the number of bodies}
        \item \code{the sketch areas}
        \item \code{the profile areas}.
    \end{itemize}
    \item \code{get\_distribution\_from\_json(json\_file)}: return a list of distributions saved in the given json file.
    \item \code{distribution\_sampling(distributions, parameters)}: sample distribution matching parameters for one design from the distributions. 
    \item \code{sample\_design(data\_dir, filter, split\_file)}: randomly sample a json file from the given dataset. 
    \item \code{sample\_sketch(json\_file, sampling\_type, \\area\_distribution)}: sample one sketch from the provided design. Three sampling types are provided: 
    \begin{itemize}
        \small
        \item \code{random}, return a sketch randomly sampled from the provided design.
        \item \code{deterministic}, return the largest sketch in the design.
        \item \code{distributive}, return a sketch that its area is in the distribution of the provided dataset.
    \end{itemize}
    \item \code{sample\_profiles(sketch\_name, max\_number\_profiles, \\sampling\_type, area\_distribution)}: sample profiles from the provided sketch. Three sampling types are provided: 
    \begin{itemize}
        \small
        \item \code{random}, return profiles randomly sampled from the provided sketch.
        \item \code{deterministic}, return profiles that are larger than the average area of the profiles in the sketch.
        \item \code{distributive}, return profiles that the areas are in the distribution of the provided dataset.
    \end{itemize}
\end{itemize}

\subsubsection{Export Commands}
Export commands enable the existing designs to be exported in the following formats:
\begin{itemize}
    \small
    \item \code{mesh(file)}: retrieve a mesh in .obj or .stl format and write it to the local file provided.
    \item \code{brep(file)}: retrieve a brep in .step, .smt, or .f3d format and write it to a local file provided.
    \item \code{sketches(dir, format)}: retrieve each sketch in .png or .dxf format and write them to a local directory provided. 
    \item \code{screenshot(file, width, height)}: retrieve a screenshot of the current design as a png image and write it to a local file provided. 
    \item \code{graph(file, dir, format)}: retrieve a face adjacency graph in a given format and write it in a local directory provided.
\end{itemize}

\subsection{CAD Reconstruction}
\label{section:appendix_cad_reconstruction}
In this section we provide additional details of the experiments performed on the CAD reconstruction task described in Section~\ref{section:cad_reconstruction}.

\begin{figure}[t]
    \includegraphics[width=\columnwidth]{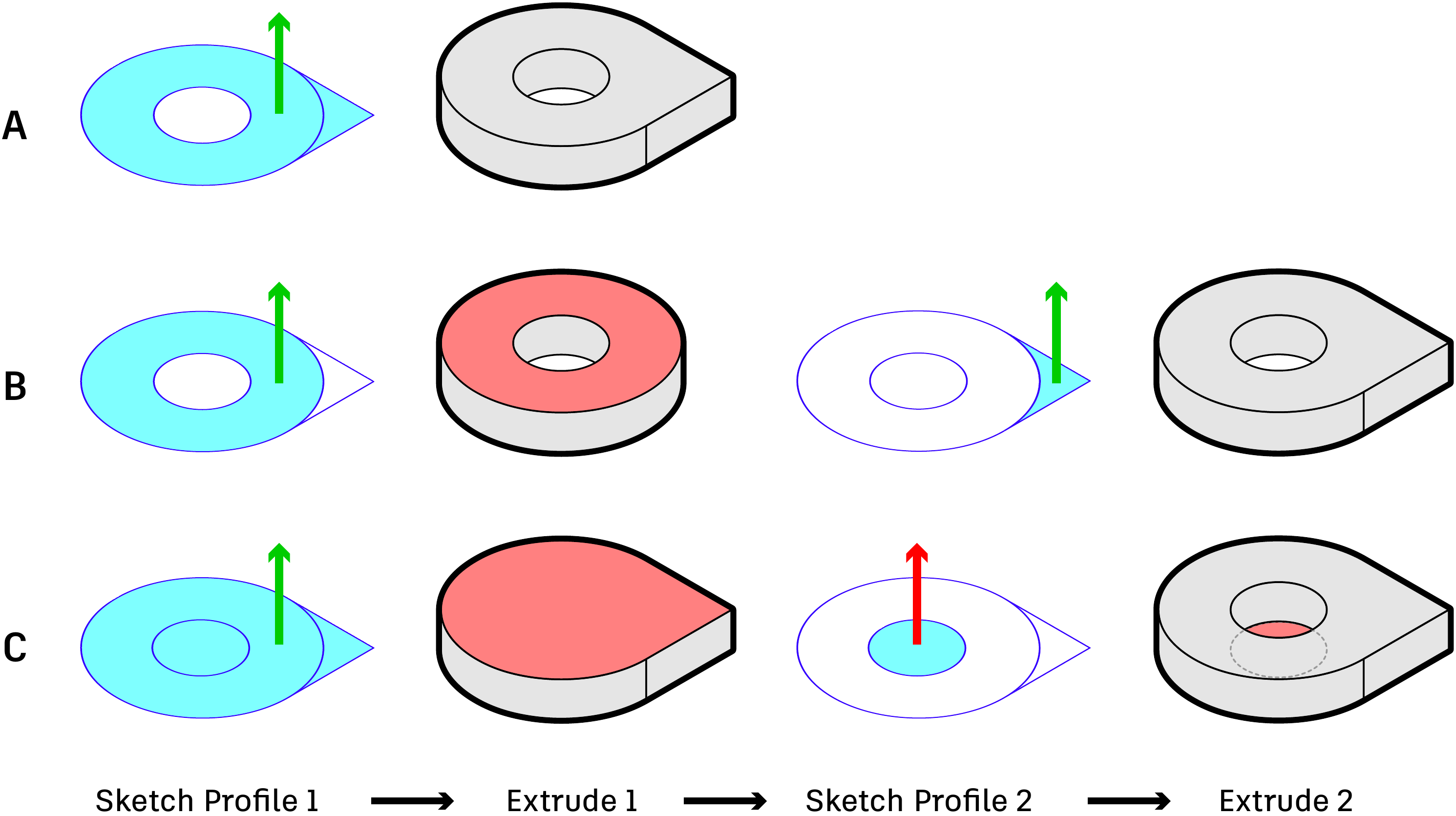}
    \caption{Different construction sequences (A-C) for the same geometry. During conversion to a face extrusion sequence, the necessary face information (highlighted in red) does not exist in the target, meaning B and C can not be converted. Green arrows indicate \textit{new body}/\textit{join} extrude operations, while red arrows indicate \textit{cut} extrude operations. }
    \label{figure:face_extrusion_limitations}
\end{figure}

\subsubsection{Data Preparation} 
The agents are trained on a subset of the reconstruction dataset that has been converted into a face extrusion sequence. Due to the simplified face extrusion representation, not all designs from the dataset can be converted to a face extrusion sequence. Figure~\ref{figure:face_extrusion_limitations} shows several common conversion limitations where necessary face information (highlighted in red) is not present in the target geometry. The intermediate top face in Figure~\ref{figure:face_extrusion_limitations} B disappears when merged with the top face of Extrude 2. In Figure~\ref{figure:face_extrusion_limitations} C a hole cut through the geometry means the intermediate top face of Extrude 1 is absent and there is no start or end face in the target geometry to perform the cut operation used in Extrude 2. Although it is possible to find alternate face extrusion sequences with heuristic rules, we instead try to maintain the user designed sequence with the exception of reversing the direction of the extrusion in some scenarios, e.g. the end face becomes the start face.

\subsubsection{Agent}
\label{section:appendix_agent}
All MPN agents employ a network architecture able to exploit the graph structure of the data, consisting of two layers passing messages along the edges of the graph. The vertex features in the face-adjacency graph are as follows:
\begin{itemize}
    \item \textit{Points}: A 10$\times$10 grid of 3D points sampled from the UV coordinate space of the B-Rep face and normalized to the bounding box of the target geometry.
    \item \textit{Normals}: A 10$\times$10 grid of 3D normal vectors sampled from the UV coordinate space of the B-Rep face.
    \item \textit{Trimming Mask}: A 10$\times$10 grid of binary values representing samples that are inside/outside the B-Rep face trimming boundary.
    \item \textit{Surface Type}: A one-hot encoded flag indicating the type of surface represented by the B-Rep face: \code{Cone}, \code{Cylinder}, \code{Elliptical}, \code{EllipticalCylinder}, \code{Nurbs}, \code{Plane}, \code{Sphere}, \\\code{Torus}.
\end{itemize}

Using the face extrusion sequence data, we train the agents in an offline manner without interacting with the \FGym{}. The \textbf{mlp} and \textbf{gcn} agents have a hidden dimension of 256 across all layers. The \textbf{gin} agent has two 256-dimensional linear layers within its graph convolution layer. The \textbf{gat} has 8 heads of 64 hidden dimensions each. The agents are trained with a dropout rate of 0.1 and a learning rate of 0.0001 for 100 epochs with the model saved at the lowest training loss. The learning rate is decreased by a factor of 0.1 upon plateau within 10 most recent epochs. Training is performed on an NVIDIA Tesla V100 with an Adam optimizer and takes approximately 6-8 hours.

\subsubsection{Search} 

In all search algorithms we mask out the following invalid actions so they are never taken:
\begin{itemize}
    \item Start faces surface types that are non-planar
    \item End faces surface types that are non-planar
    \item Operation types other than \textit{new body} when the current geometry is empty
\end{itemize}
Other invalid actions that require geometric checks, such as specifying a start face and end face that are co-planar, are returned as invalid from the \FGym{} and count against the search budget.

\begin{figure*}
    \includegraphics[width=\textwidth]{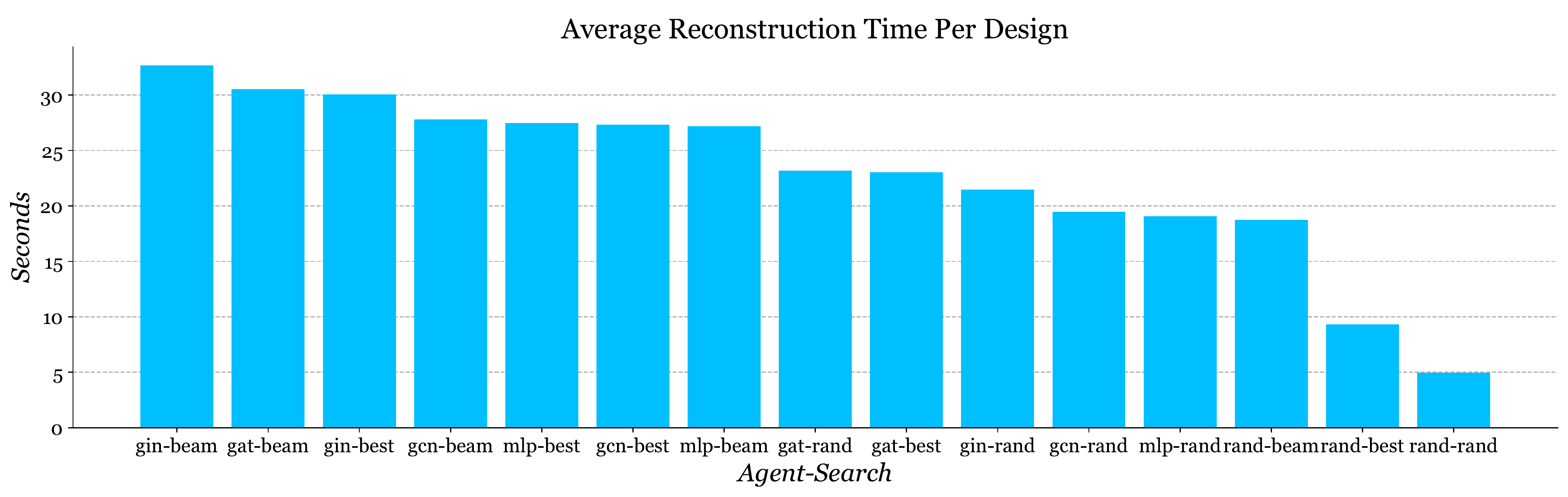}
    \caption{ Average reconstruction time per design for combinations of agents and search strategies. }
    \label{figure:reconstruction_time}
\end{figure*}

\subsubsection{Evaluation} 
We perform evaluation using the official test set containing 1725 designs. Evaluation is performed in an online manner using the \FGym{}. Figure~\ref{figure:reconstruction_time} shows the average reconstruction time for each design with combinations of agents and search strategies. We set a hard time limit of 10 minutes per design, after which we halt search, affecting between 0-14 designs depending on the agent and search strategy. Between 0-15 designs cause software crashes. 17 designs in the test set cannot be represented as graphs due to our data pipeline currently not supporting edges with more than two adjacent faces. In all failure cases we use the best seen IoU, or 0 if no IoU score is available, and consider the design to fail at exact reconstruction.

\begin{table*}
    \caption{Reconstruction results for multiple agent and search combinations trained on human designed data. IoU and exact reconstruction are shown at 20 and 100 search steps. The best result in each column is shown in bold. Lower values are better for conciseness. \textbf{\# Parameters} indicated the number of parameters used by each agent. }
    \label{tab:benchmark_appendix}
    \centering
    \begin{tabular}{cc|cccccc}
    \toprule
       \textbf{Agent} &  \textbf{Search} & \multicolumn{2}{c}{\textbf{IoU}} & \multicolumn{2}{c}{\textbf{Exact Reconstruction. \%}} & \textbf{Conciseness} & \textbf{\# Parameters.} \\
             &  {} &    20 Steps &    100 Steps &        20 Steps & 100 Steps & {}\\
    \midrule
          gcn &   rand &   0.8644 &    0.9042 &                 \textbf{0.6232} &    \textbf{0.6754} &      1.0168 & 3.02M\\
          gcn &   beam &   0.8640 &    0.8982 &                 0.5739 &    0.6122 &      0.9275 & 3.02M\\
          gcn &   best &   0.8831 &    \textbf{0.9186} &                 0.5971 &    0.6348 &      0.9215 & 3.02M\\
          mlp &   rand &   0.8274 &    0.8596 &                 0.5658 &    0.5965 &      0.9763 & 2.24M\\
          mlp &   beam &   0.8619 &    0.8995 &                 0.5525 &    0.5884 &      0.9271 & 2.24M\\
          mlp &   best &   0.8712 &    0.8991 &                 0.5675 &    0.5977 &      0.9305 & 2.24M\\
          gat &   rand &   0.8742 &    0.9128 &                 0.6191 &    0.6742 &      1.0206  & 3.03M\\
          gat &   beam &   0.8691 &    0.9016 &                 0.5791 &    0.6133 &      0.9261  & 3.03M\\
          gat &   best &   \textbf{0.8895} &    0.9139 &                 0.5994 &    0.6354 &      0.9290  & 3.03M\\
          gin &   rand &   0.8346 &    0.8761 &                 0.5901 &    0.6301 &      1.0042 & 3.62M\\
          gin &   beam &   0.8500 &    0.8913 &                 0.5594 &    0.5983 &      0.9299 & 3.62M\\
          gin &   best &   0.8693 &    0.9007 &                 0.5803 &    0.6122 &      0.9340 & 3.62M\\
         rand &   rand &   0.6840 &    0.8386 &                 0.4157 &    0.5380 &      1.2824 & -\\
         rand &   beam &   0.4785 &    0.6277 &                 0.2812 &    0.3896 &      0.9118 & -\\
         rand &   best &   0.6334 &    0.7994 &                 0.3693 &    0.4887 &      \textbf{0.8979} & -\\    
    \bottomrule
    \end{tabular}
\end{table*}

\subsubsection{Results}
Table~\ref{tab:benchmark_appendix} details the full set of results for all agents and search strategies in the extended baseline comparison experiment from Section~\ref{section:baseline_comparison}. We also include the number of parameters used by each agent.

\subsection{Tasks}
The ground-truth sequence of the dataset, along with the gym environment, can be used to automatically derive a range of labels for tasks other than CAD reconstruction, such as, program synthesis, sequence modeling, generative models, and geometric deep learning. Example tasks include:
\begin{itemize}
    \item \textit{Classification} of designs by construction sequence length.
    \item \textit{Segmentation} of B-Rep faces by extrude operation order or by start/side/end face of an extrude operation.
    \item \textit{Modeling operation order prediction} to recover the correct order of construction from raw geometry.
    \item \textit{Sketch synthesis} to recover the original sketch, including constraints and dimensions, from the 3D geometry. 
    \item \textit{Predicting next action} in the design sequence for `CAD autocomplete'.
    \item \textit{Generative models} that are aware of the design sequence and constraints.
\end{itemize}

\end{document}